\PassOptionsToPackage{table,dvipsnames}{xcolor} % pre-empt an option clash with egpubl
\documentclass{egpubl}
\usepackage{eg2027}
\ConferenceSubmission

\usepackage[T1]{fontenc}
\usepackage{dfadobe}
\usepackage{cite}          % comment out for biblatex/biber
\BibtexOrBiblatex
\electronicVersion
\PrintedOrElectronic
\ifpdf \usepackage[pdftex]{graphicx} \pdfcompresslevel=9
\else  \usepackage[dvips]{graphicx} \fi
\usepackage{egweblnk}
\makeatletter
\AtBeginDocument{%
  \let\arxiv@ps@titlepage\ps@titlepage
  \def\ps@titlepage{\arxiv@ps@titlepage\let\@oddhead\@empty\let\@evenhead\@empty}%
  \copyrightTextTitPag{}\copyrightTextRunPag{}% also drop the "submitted to EUROGRAPHICS 2027." footer
}
\let\arxiv@maketitle\@maketitle
\def\@maketitle{\vspace*{-1cm}\arxiv@maketitle}
\makeatother

\usepackage{amsmath}
\usepackage{amssymb}
\usepackage{amsthm}
\usepackage{booktabs}
\usepackage{multirow}
\usepackage{longtable}
\usepackage{subcaption}
\usepackage{algorithm}
\usepackage{algpseudocode}
\usepackage{xspace}
\usepackage{xcolor}

\newcommand{\sds}{SDS\xspace}
\newcommand{\vsd}{VSD\xspace}
\newcommand{\sdi}{SDI\xspace}
\newcommand{\mvsdi}{MV-SDI\xspace}
\newcommand{\K}{K}
\newcommand{\thetaparam}{\theta}
\newcommand{\camera}{\mathbf{c}}
\newcommand{\Render}{\mathcal{R}}
\newcommand{\noisedlatent}{\mathbf{z}_t}
\newcommand{\timestep}{t}
\newcommand{\prompt}{y}
\newcommand{\unet}{\epsilon_\phi}
\newcommand{\guidance}{\nabla_{\thetaparam} \mathcal{L}}
\DeclareMathOperator{\Var}{Var}
\DeclareMathOperator{\Cov}{Cov}
\DeclareMathOperator{\Corr}{Corr}
\DeclareMathOperator{\E}{\mathbb{E}}

\newcommand{\eg}{e.g.\@\xspace}

\begin{document}

% Number sections and subsections only; keep run-in \paragraph headings unnumbered
% (the EG class otherwise numbers \paragraph, e.g. "2.0.0.1").
\setcounter{secnumdepth}{2}

\title[Stratified Multi-View Aggregation for Score Distillation]%
      {Stratified Multi-View Aggregation for Score Distillation}

% Double-blind: submission ID instead of author names, affiliation left blank.
% \author[Paper~\paperID]{Paper~\paperID}
% Final authors (uncomment after acceptance, comment line above)
\author[M.~Lupa\c{s}cu et al.]
  {Marian Lupa\c{s}cu$^{1,2}$,
   Mihai-Sorin Stupariu$^1$,
   Wongi Park$^3$,
   Ionu\c{t} Mironic\u{a}$^2$
   \\
   $^1$University of Bucharest, Romania
   \quad $^2$Adobe Research
   \quad $^3$Ajou University, South Korea}

% Teaser
\teaser{
  \centering
  \vspace{-1.3cm}
  \includegraphics[width=0.97\linewidth]{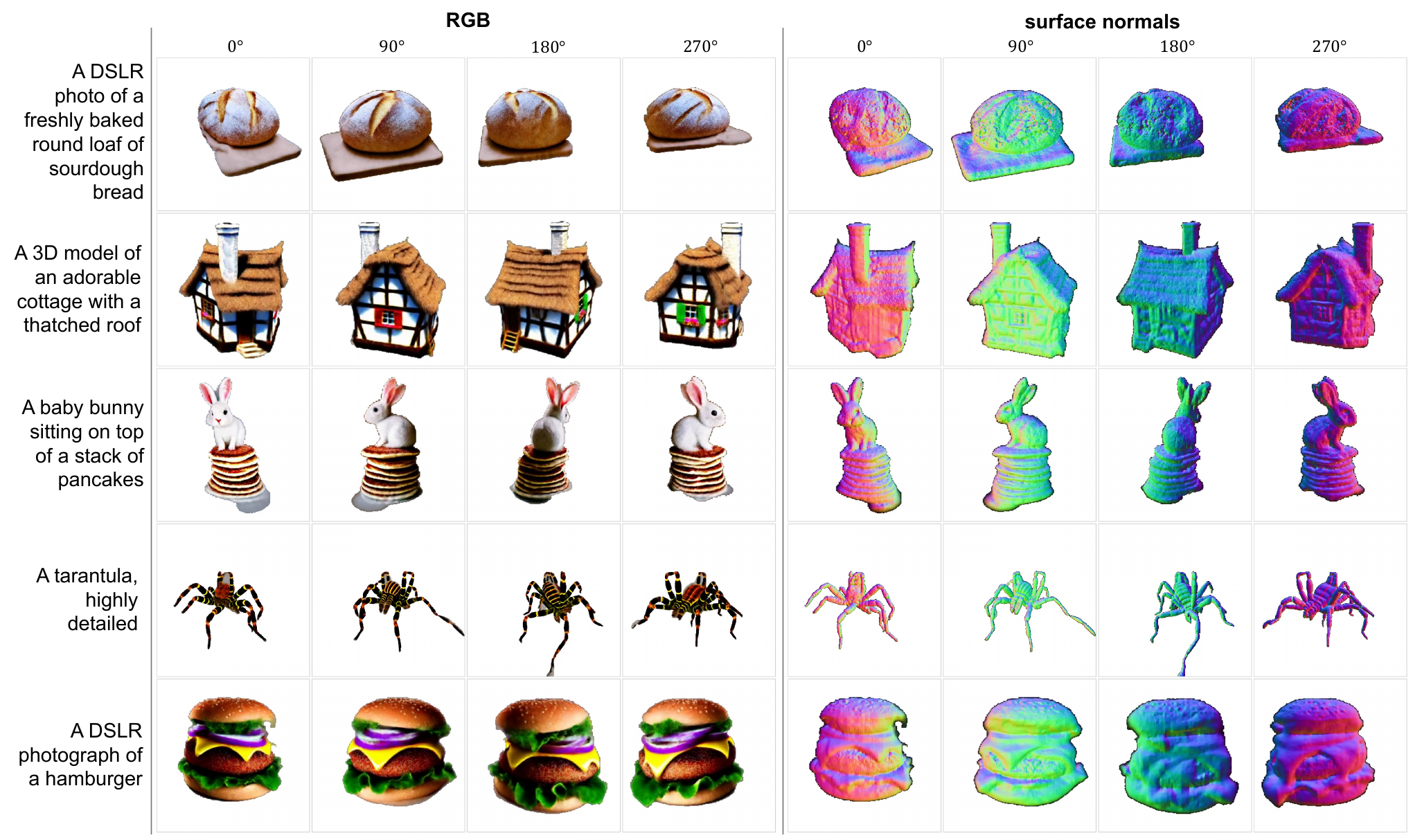}
  \vspace{-0.3cm}
  \caption{\textbf{Multi-View Aggregated Score Distillation (\mvsdi) yields sharp, view-consistent 3D assets in half the optimization steps.} \mvsdi aggregates score-distillation gradients from $K$ cameras drawn as antithetic (antipodal) pairs each step, replacing the single-camera estimate of \sdi~\cite{lukoianov2024sdi} at an identical UNet-call budget. Each $K{=}2$ result is shown as RGB and surface normals at $0^\circ/90^\circ/180^\circ/270^\circ$.}
  \label{fig:teaser}
}

\maketitle

%-------------------------------------------------------------------------
\begin{abstract}
Score distillation turns a pretrained 2D diffusion model into a 3D
generator, but the per-step gradient is estimated from a single random
view: this one-sample estimate has high variance (different views of
the same partial scene disagree) and is blind to global shape
consistency. Existing multi-view approaches address this by retraining
the diffusion prior on multi-view data; this improves consistency but
conflates the sampling contribution with the quality of the retrained
prior. We instead isolate the sampling axis, leaving the prior frozen.
We introduce Multi-View Aggregated Score Distillation (MV-SDI), a
training-free sampler that replaces the single-view per-step gradient
with an average over $K$ views at a fixed UNet-call budget. Averaging
$K$ views lowers the per-step gradient variance toward $1/K$ of its
single-view value. Drawing the $K$ views as antithetic antipodal pairs
adds no further variance reduction (measured antipodal correlation
$\rho \approx 0$) but stratifies angular coverage (every step covers
both hemispheres) removing the same-hemisphere clustering of
independent sampling. At a fixed $10{,}000$-UNet-call budget on the
43-prompt SDI benchmark, $K{=}2$ halves the optimization steps and
raises CLIP R-Precision from $74.8\%$ to $83.8\%$ and CLIP score from
$0.297$ to $0.312$ over the single-view SDI baseline, with consistent
gains on HPSv2 and ImageReward and a $0.0\%$ divergence rate. $K{=}4$
gives a fourfold step reduction at R-Precision $86.9\%$ and CLIP
$0.307$. The gains concentrate on hard prompts where single-view
distillation collapses, at a measured cost in CLIP-IQA. MV-SDI is
drop-in for gradient-based score-distillation pipelines, including
Score Distillation via Inversion and plain SDS, and requires no
retraining and no multi-view data. Code is available at: \href{https://github.com/marianlupascu/MV-SDI}{\texttt{MV-SDI repository}}.


\end{abstract}
%-------------------------------------------------------------------------

\section{Introduction}
\label{sec:intro}

Score Distillation Sampling (\sds)~\cite{poole2023dreamfusion} lifts
a frozen 2D diffusion prior into a 3D generator by supervising a
Neural Radiance Field (NeRF)~\cite{mildenhall2020nerf} from text
alone, without 3D data. Its variants share this recipe but change the
distillation loss: Variational Score Distillation
(\vsd)~\cite{wang2023prolificdreamer}, DDIM-reparametrized
distillation~\cite{song2021ddim}, and Score Distillation via Inversion
(\sdi)~\cite{lukoianov2024sdi}. Two obstacles limit their practical
use. Optimization is slow: producing a single asset requires
$10{,}000$ or more iterations, each invoking a UNet pass on Stable
Diffusion~\cite{rombach2022high}. Geometry is often view-inconsistent:
the 2D prior's frontal bias is never corrected because each step
supervises the model from a single camera. One view cannot reveal
that views disagree about global structure, leaving the \emph{Janus
problem}~\cite{poole2023dreamfusion} (objects rendered with duplicated
frontal faces instead of a consistent front-and-back) unchecked.

Slow convergence traces to the single-camera choice. The per-step
distillation gradient is a one-sample Monte Carlo estimate of an
expectation over views, and its variance, more than the capacity of
the prior, constrains the convergence rate~\cite{hammersley1964monte,
glynn2002some}. Recent work reduces this variance along the noise and
timestep axes of the estimator~\cite{wang2024steindreamer,
bettencourt2026carv}, but the camera axis remains untouched, and it is
the one that dominates when views disagree most. The same
single-camera choice exacerbates the Janus problem: views never
jointly penalize an inconsistent geometry. Existing multi-view
approaches~\cite{shi2023mvdream,shi2023imagedream} tackle this by
retraining the diffusion prior on multi-view data, which is costly
and conflates the sampling contribution with the quality of the
retrained prior. We instead ask: how much is recoverable by smarter
sampling alone, with the prior frozen?

We propose \emph{Multi-View Aggregated Score Distillation} (\mvsdi), a
training-free, memory-neutral sampler that replaces the single-view
per-step gradient with an average over $\K$ views at a fixed total
UNet-call budget, leaving the prior frozen. UNet evaluations dominate
the compute cost of score-distillation
pipelines~\cite{poole2023dreamfusion,lukoianov2024sdi}, so holding
this budget fixed means the $\K$ views per step translate to $\K\times$
fewer optimization steps, not $\K\times$ more compute. Averaging $\K$
estimates lowers the view-induced variance toward $1/\K$ of its
single-view value. We structure the $\K$ views as
\emph{antithetic antipodal pairs}, drawing each camera together with
its $180^\circ$-rotated twin to guarantee balanced hemispheric
coverage. To our knowledge, \mvsdi is the first to apply antithetic
sampling to the camera axis of score distillation. A negative
correlation between antipodes would push variance below $1/\K$; we
measure this correlation to be $\approx\!0$ in our setting
(Sec.~\ref{sec:experiments}), so the pairing helps through
\emph{stratified coverage} rather than further variance reduction.
Peak memory is unchanged: gradient accumulation across the $\K$ views
replaces the single-view update at the single-view footprint. Within
this framework we study antithetic structure along one, two, and
three orthogonal planes, with progressively larger elevation ranges
to probe where the 2D prior degrades.

Our contributions are: (i) \mvsdi, a training-free, memory-neutral
sampler that reframes the convergence and consistency problems of
score distillation as partly a sampling problem, with an antithetic
antipodal design that adds \emph{stratified coverage} on top of the
$1/\K$ variance reduction of plain aggregation; (ii) an evaluation on
the 43-prompt \sdi{} benchmark showing that \mvsdi improves every
alignment and preference metric at $2\text{--}4\times$ fewer
optimization steps, with the image-quality cost traceable to the 2D
prior rather than the sampler; and (iii) \emph{Consensus-Weighted
\mvsdi}, a self-supervised extension that learns a scalar weight per
view, down-weighting views that disagree with the multi-view
consensus.

\vspace{-0.2cm}
\section{Related Work}
\label{sec:related}

\begin{figure*}[t]
    \centering
    % \vspace{-0.5cm}
    \includegraphics[width=0.99\linewidth]{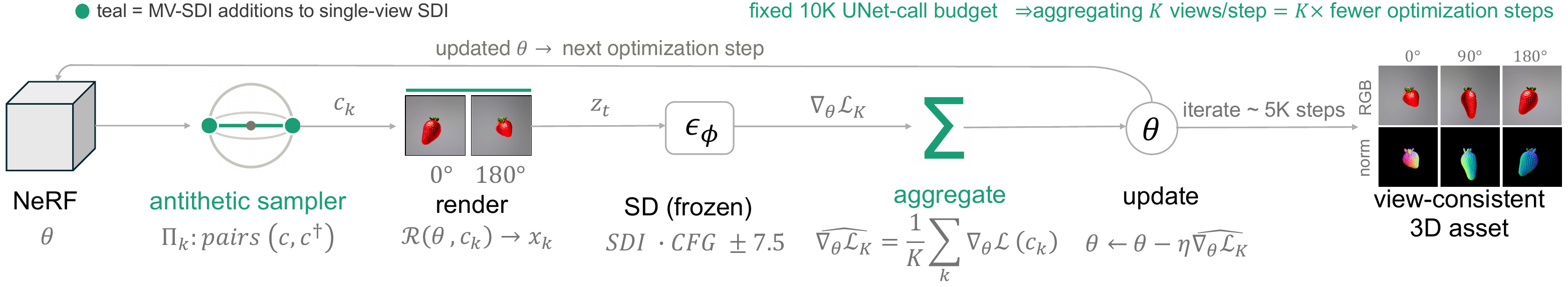}
    \vspace{-0.3cm}
    \caption{\textbf{\mvsdi in one optimization step.} The NeRF $\thetaparam$ is rendered from $\K$ cameras drawn in antithetic pairs $(\camera, \camera^\dagger)$ that share one noised timestep $(\timestep, \epsilon)$; each rendering passes through the frozen SD-2.1 prior under the \sdi loss, and the per-view gradients are averaged and applied by gradient accumulation, leaving peak memory at the single-view footprint. Teal marks the only two additions to single-view \sdi: the antithetic $\K$-view sampler $\Pi_\K$ and the $1/\K$ aggregation. Right: a converged asset (RGB and surface normals at $0^\circ/90^\circ$$/$$180^\circ$).}
    \label{fig:pipeline}
    \vspace{-0.4cm}
\end{figure*}

\paragraph{Score distillation for text-to-3D.} Score distillation began
with DreamFusion~\cite{poole2023dreamfusion}, which showed that a
frozen 2D diffusion prior~\cite{rombach2022high} can supervise a
NeRF~\cite{mildenhall2020nerf} directly, without any 3D data, by
matching the prior's noise prediction on randomly noised renderings.
Score Jacobian Chaining~\cite{wang2023sjc} derived the same lifting
independently, and Magic3D~\cite{lin2023magic3d} and
Fantasia3D~\cite{chen2023fantasia3d} soon scaled it to higher
resolution and disentangled geometry from appearance. The recipe was
in place, but the \sds gradient was noisy and biased toward
over-smoothed geometry, and much of the follow-up work is a search for
a better distillation loss. ProlificDreamer replaces score matching
with a per-scene LoRA critic (VSD)~\cite{wang2023prolificdreamer,
hu2022lora}. \sdi~\cite{lukoianov2024sdi}, which we build on, inverts
each rendering through DDIM~\cite{song2021ddim} to recover a
lower-variance target. NFSD~\cite{katzir2024nfsd} and
CSD~\cite{yu2024csd} isolate the classifier-free-guidance term and
drop the large guidance scale \sds needs;
LucidDreamer~\cite{liang2024luciddreamer} matches DDIM intervals;
Consistent3D~\cite{wu2024consistent3d} uses a deterministic ODE prior;
HiFA~\cite{zhu2023hifa} refines the guidance; and
ESD~\cite{wang2023taming} restores the variational entropy term to
curb mode collapse. A parallel thread reschedules the diffusion
timestep instead of the loss~\cite{huang2023dreamtime,
ma2024scaledreamer}. For all their differences, these methods share
one design choice: the per-step gradient is estimated from a single
random camera. That is the axis \mvsdi acts on. We build on \sdi as
the strongest of the family, but the recipe is loss-agnostic.

\vspace{-0.2cm}
\paragraph{Variance reduction in stochastic optimization.} The problem
\mvsdi addresses (a per-step gradient estimated from too few samples)
is a classical one in Monte Carlo optimization, and the toolkit
for it is long-standing~\cite{hammersley1964monte,owen2013monte}.
Antithetic sampling draws inputs in negatively correlated pairs so
that noise cancels in the average; control variates and common random
numbers exploit correlations between the estimand and a cheap
surrogate. These techniques underlie differentiable
rendering~\cite{zeltner2021monte}, reinforcement
learning~\cite{tucker2018mirage}, and, more recently, diffusion
training itself via antithetic timestep
sampling~\cite{kingma2021variational}. Score distillation has begun to
inherit these ideas, but along axes orthogonal to ours:
SteinDreamer~\cite{wang2024steindreamer} adds a Stein control variate
on the noise, CARV~\cite{bettencourt2026carv} importance-samples the
timestep and noise, and RewardSDS~\cite{chachy2025rewardsds} reweights
noise samples by a reward. All three act on the noise, timestep, or
reward term of the same estimator, leaving the camera term untouched.
\mvsdi acts on the camera term, so the three lines compose rather than
compete.

\vspace{-0.2cm}
\paragraph{Multi-view priors and the Janus problem.} A separate line
of work addresses view inconsistency at its root, by replacing the 2D
prior with a 3D-aware one. MVDream~\cite{shi2023mvdream} fine-tunes
Stable Diffusion to emit sets of consistent views in one pass;
ImageDream~\cite{shi2023imagedream} adds image conditioning; and a
broader family of pose-conditioned and multi-view diffusion models
targets novel-view or consistent multi-view
synthesis~\cite{liu2023zero123,shi2023zero123pp,liu2023syncdreamer,
long2024wonder3d,li2024era3d,tang2023mvdiffusion}. These priors are
powerful, but they demand multi-view training data and a retrained
backbone; more importantly for us, they entangle two changes at once:
a stronger prior and a multi-view sampler, so the sampling gain
cannot be isolated. Two recent efforts sidestep retraining. Consistent
Flow Distillation~\cite{yan2025cfd} keeps the prior frozen and makes
the injected \emph{noise} consistent across views; view-dependent
negative prompting~\cite{armandpour2023perpneg} tackles the Janus
problem with no new prior at all. \mvsdi belongs to this second camp.
It sits on top of any of the priors above and needs no retraining;
with a stock SD~2.1 prior, smarter sampling alone recovers much of
the quality and consistency usually credited to specialized 3D-aware
backbones.

\vspace{-0.2cm}
\paragraph{View selection and next-best-view.} A related question
surfaces in a different community under the name of next-best-view
planning: which viewpoints matter most? Given a real scene and a
sequential reconstruction budget, these methods pick the camera that
most reduces reconstruction uncertainty, either through learned
scanning agents~\cite{peralta2020nbv,chen2024gennbv} or through
ray-based uncertainty estimated on a neural radiance
field~\cite{lee2022uncertaintynbv}. The objective is information gain
under active perception. \mvsdi shares the vocabulary of view
selection but not the setting: there is no real scene to probe and no
sequential policy to learn. Cameras are drawn once, from a fixed
antithetic stratification of the rendering distribution, and the goal
is to reduce the \emph{variance of the distillation gradient} in a
generative optimisation loop. Both lines share the same theme: some
views are more useful than others; the objectives differ.

\vspace{-0.2cm}
\paragraph{3D representations and feed-forward generation.} Score
distillation is agnostic to the underlying 3D representation. We
optimize a NeRF, but a parallel line of work distills into 3D Gaussian
Splatting~\cite{kerbl20233dgs} for faster and sharper assets
(DreamGaussian~\cite{tang2024dreamgaussian},
GaussianDreamer~\cite{yi2024gaussiandreamer}). \mvsdi modifies only
the camera distribution of the per-step gradient, so it is
representation-agnostic in principle; we validate it on NeRF and
leave a Gaussian-splatting instantiation to future work. A more
radical alternative sidesteps per-scene optimisation altogether:
feed-forward models map text or a single image to 3D in one pass,
predicting a triplane NeRF (LRM~\cite{hong2024lrm},
Instant3D~\cite{li2024instant3d}) or Gaussians
(LGM~\cite{tang2024lgm}). These are fast at inference, but they
require large 3D training corpora and inherit the biases of that
distribution. \mvsdi sits at the other end of the trade-off: it
retains the appeal of optimisation-based methods, namely no 3D
supervision and an open-vocabulary 2D prior, while cutting much of
their per-asset cost.

\vspace{-0.7cm}

\vspace{-0.7cm}
\section{Approach}
\label{sec:approach}

Figure~\ref{fig:pipeline} summarizes the method: \mvsdi leaves the
single-view \sdi pipeline unchanged except for two components: it
draws $\K$ cameras per step in antithetic pairs and averages their
gradients under a fixed UNet-call budget. UNet evaluations dominate
the compute cost of score distillation~\cite{poole2023dreamfusion,
lukoianov2024sdi}, so holding this budget fixed means the $\K$ views
per step translate to $\K{\times}$ fewer optimization steps, not
$\K{\times}$ more compute.

\vspace{-0.2cm}
\subsection{Score Distillation via Reparametrized DDIM}
\label{sec:prelim}
Let $\thetaparam$ parametrize a 3D representation (we use Instant-NGP
NeRF~\cite{mueller2022instant} throughout) and
$\Render(\thetaparam, \camera)$ denote a differentiable rendering at
camera pose $\camera$. Let $\unet(\noisedlatent, \timestep, \prompt)$
be a frozen 2D diffusion UNet that predicts the noise added to a
latent $\noisedlatent$ at diffusion timestep $\timestep \in [0, 1]$,
conditioned on text prompt $\prompt$. \sds~\cite{poole2023dreamfusion}
updates $\thetaparam$ by minimizing:
\begin{equation}
    \mathcal{L}_{\sds}(\thetaparam) \;=\; \E_{\timestep, \camera, \epsilon}\big[ w(\timestep)\,\|\unet(\noisedlatent(\camera), \timestep, \prompt) - \epsilon\|_2^2\big],
    \label{eq:sds}
\end{equation}
where the noised latent
$\noisedlatent(\camera) = \alpha_{\timestep} \mathcal{E}(\Render(\thetaparam, \camera)) + \sigma_{\timestep} \epsilon$
follows the schedule $(\alpha_{\timestep}, \sigma_{\timestep})$,
$\epsilon \sim \mathcal{N}(\mathbf{0}, \mathbf{I})$ is Gaussian noise,
$\mathcal{E}$ is the SD encoder, and $w(\timestep)$ a weighting
function. The per-camera gradient is:
\begin{equation}
    \guidance_{\sds}(\camera) \;=\; w(\timestep)\,\big(\unet(\noisedlatent(\camera), \timestep, \prompt) - \epsilon\big)\,\frac{\partial \noisedlatent(\camera)}{\partial \thetaparam},
    \label{eq:sds-grad}
\end{equation}
where the dependence on $\camera$ enters through
$\Render(\thetaparam, \camera)$; we suppress the arguments
$(\timestep, \epsilon, \thetaparam)$ for brevity.
\sdi~\cite{lukoianov2024sdi} replaces the Gaussian noise $\epsilon$
with a \emph{reparametrized} target $\hat{\epsilon}_{\sdi}(\camera)$
obtained by DDIM-inverting the current rendering
$\Render(\thetaparam, \camera)$ and re-denoising it under the
prompt-conditioned UNet:
\begin{equation}
    \guidance_{\sdi}(\camera) \;=\; w(\timestep)\,\big(\unet(\noisedlatent(\camera), \timestep, \prompt) - \hat{\epsilon}_{\sdi}(\camera)\big)\,\frac{\partial \noisedlatent(\camera)}{\partial \thetaparam}.
    \label{eq:sdi-grad}
\end{equation}
In both cases the only stochasticity in a step is
$(\timestep, \camera, \epsilon)$, and the gradient is computed from
\emph{one} camera $\camera$. We instantiate \mvsdi on \sdi throughout,
since \sdi is the strongest single-view baseline in the family; the
aggregation of Section~\ref{sec:mvaggregation} applies unchanged to the
\sds gradient (Equation~\ref{eq:sds-grad}), which we confirm in
Section~\ref{sec:sds-composition}.

\vspace{-0.2cm}
\subsection{Multi-view aggregation}
\label{sec:mvaggregation}
The single-view estimator in Equation~\eqref{eq:sds-grad}--\eqref{eq:sdi-grad}
has variance $\sigma^2 := \Var_\camera[\guidance(\camera) \mid \timestep, \epsilon]$
(dropping the \sdi{} subscript since the analysis holds for both losses)
dominated by different views of the same incomplete 3D representation
produce very different gradients. We estimate $\guidance$ from $\K$
cameras per step, indexed by $k = 1, \dots, \K$,
\begin{equation}
    \widehat{\guidance}_{\K} \;=\; \frac{1}{\K} \sum_{k=1}^{\K} \guidance(\camera_k),
    \label{eq:mv}
\end{equation}
which for independently sampled cameras gives
$\Var[\widehat{\guidance}_{\K}] = \sigma^2 / \K$, the $1/\K$ variance
reduction.

\begin{figure}[t]
    \centering
    % \vspace{-0.2cm}
    \includegraphics[width=0.97\linewidth]{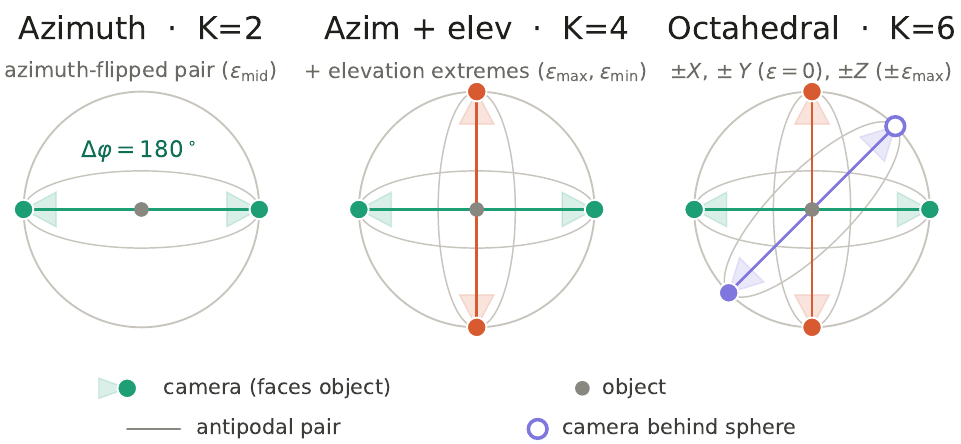}
    \vspace{-0.1cm}
    \caption{\textbf{Antithetic camera-sampling strategies on the view sphere.} Pairs along $1$ (azimuth), $2$ (Azim+elev), and $3$ (octahedral) orthogonal great circles. Each pair is rendered simultaneously and contributes to a single aggregated gradient.}
    \label{fig:sampling}
    \vspace{-0.7cm}
\end{figure}

\vspace{-0.2cm}
\paragraph{Antithetic camera pairs.} Beyond independent draws, we
sample cameras in \emph{antithetic pairs}
$(\camera, \camera^\dagger)$, where $\camera^\dagger$ rotates
$\camera$ by $180^\circ$ about a chosen axis; for $\K{=}2$ this is
one base view and its azimuth-flipped twin
($\phi \to \phi + 180^\circ$, same elevation and radius). The
motivation is negative correlation: if the gradient were a strongly
odd function of viewing direction, antipodal draws would push
variance below $\sigma^2/\K$~\cite{hammersley1964monte}. Whether the
SDI gradient is that odd is an empirical question, and we measure
the antipodal correlation to be $\approx\!0$
(Section~\ref{sec:mainresults}, Appendix~\ref{sec:appendix-math}), so
the pair \emph{attains} the $1/\K$ iid rate rather than beating it,
contrary to the classical antithetic expectation. Its value is as a
\emph{stratified} sampler: every step is guaranteed a view and its
antipode, removing unlucky same-hemisphere draws that independent
sampling produces. As Section~\ref{sec:mainresults} shows, it is this
coverage, not extra variance reduction, that eliminates the residual
divergences and steadies quality at unchanged mean alignment.

\vspace{-0.2cm}
\subsection{Multi-axis antithetic sampling}
\label{sec:multiaxis}
The azimuthal ($1$-plane) variant enforces consensus only about the front/back axis. To probe consensus on more axes, we extend antithetic structure to several orthogonal great circles of the camera sphere (Figure~\ref{fig:sampling}):
\begin{itemize}
    \item \textbf{Azimuth+Elevation} ($\K{=}4$, $2$ planes): one azimuth-flipped pair at mid-elevation plus one elevation-flipped pair at a second
    azimuth, \emph{i.e.}\ $(\phi_a, \eta_{\text{mid}})$,
    $(\phi_a{+}180^\circ, \eta_{\text{mid}})$ and
    $(\phi_b, \eta_{\max})$, $(\phi_b, \eta_{\min})$, where
    $\eta$ denotes camera elevation.
    \item \textbf{Octahedral} ($\K{=}6$, $3$ planes): three axis-aligned
    pairs $\pm X$, $\pm Y$, $\pm Z$ relative to a base azimuth (in an
    object-centered coordinate frame), with the vertical pair $\pm Z$
    at $\eta = \pm \eta_{\max}$ and the horizontal pairs $\pm X, \pm Y$
    at $\eta{=}0^\circ$.
\end{itemize}
Each strategy preserves the $\K$-view aggregation of Equation~\eqref{eq:mv}; only the joint distribution of the $\K$ cameras changes.

\vspace{-0.2cm}
\subsection{Memory-neutral implementation via grad. accumulation}
\label{sec:gradacc}
Rendering $\K$ views in a single pass scales NeRF memory linearly in
$\K$, which becomes the bottleneck as $\K$ grows. We instead use
\emph{gradient accumulation}: we render and back-propagate the $\K$
views sequentially, scaling each per-view loss by $1/\K$, and step the
optimizer only after all $\K$ views accumulate
(Algorithm~\ref{alg:mvsdi}). Peak memory then matches single-view
training, while the update equals multi-view averaging (measured in
Section~\ref{sec:setup}). Because we hold the \emph{total} UNet budget
fixed (e.g.\ $10{,}000$ calls), $\K$-view aggregation cuts the number
of optimization steps by $\K\times$, the speedup we report in
Section~\ref{sec:mainresults}. \mvsdi is a drop-in modification of the
training loop, compatible with any \sds, \vsd, or \sdi loss; we
instantiate it on \sdi and ablate $\K$ and the antithetic axes in
Section~\ref{sec:ablation}.

\vspace{-0.1cm}
\begin{algorithm}[h]
\caption{\mvsdi optimization}
\label{alg:mvsdi}
\begin{algorithmic}[1]
\Require NeRF $\thetaparam$, prompt $y$, camera sampler $\Pi_\K$, UNet-call budget $N$
\For{$\mathit{step} = 1, \dots, N/\K$} \Comment{$\K\times$ fewer steps at fixed budget}
    \State $\{\camera_k\}_{k=1}^\K \gets \Pi_\K()$ \Comment{sample $\K$ views}
    \State sample shared $\timestep, \epsilon$
    \For{$k = 1, \dots, \K$}
        \State $x_k \gets \Render(\thetaparam, \camera_k)$
        \State $\guidance_k \gets$ \sdi-gradient (Equation~\ref{eq:sdi-grad}) for $x_k, \timestep, \epsilon$
        \State $\thetaparam.\text{grad} \mathrel{+}= \frac{1}{\K} \guidance_k$ \Comment{accumulate}
    \EndFor
    \State \texttt{optimizer.step}()
\EndFor
\end{algorithmic}
\vspace{0.2em}
\small
\textit{Notation.} $\timestep \in [0, 1]$: diffusion timestep;
$\epsilon \sim \mathcal{N}(\mathbf{0}, \mathbf{I})$: Gaussian noise;
$x_k = \Render(\thetaparam, \camera_k)$: rendering at camera $\camera_k$;
$w(\timestep)$: weighting function from Equation~\eqref{eq:sds};
$\hat{\epsilon}_{\sdi}$: reparametrized target from Equation~\eqref{eq:sdi-grad}.
\end{algorithm}
\vspace{-0.5cm}

% \begin{algorithm}[t]
% \caption{\mvsdi optimization}
% \label{alg:mvsdi}
% \begin{algorithmic}[1]
% \Require NeRF $\thetaparam$, prompt $y$, camera sampler $\Pi_\K$ (one of: random, azimuth, mixed, octa), UNet-call budget $N$
% \For{$\mathit{step} = 1, \dots, N/\K$} \Comment{$\K\times$ fewer steps at fixed budget}
%     \State $\{\camera_k\}_{k=1}^\K \gets \Pi_\K()$ \Comment{sample $\K$ views}
%     \State sample shared timestep $\timestep$ and noise $\epsilon$
%     \For{$k = 1, \dots, \K$}
%         \State $x_k \gets \Render(\thetaparam, \camera_k)$
%         \State $\guidance_k \gets$ \sdi-gradient (Equation~\ref{eq:sdi-grad}) for $x_k, \timestep, \epsilon$
%         \State $\thetaparam.\text{grad} \mathrel{+}= \frac{1}{\K} \guidance_k$ \Comment{accumulate}
%     \EndFor
%     \State \texttt{optimizer.step}() \Comment{single update}
% \EndFor
% \end{algorithmic}
% \end{algorithm}
 
% \vspace{-0.4cm}
% \vspace{-0.2cm}
\section{Experiments}
\label{sec:experiments}

We describe the setup in Section~\ref{sec:setup}. Ten findings then
organize the analysis, in four groups: the aggregate gain and how
it scales in $\K$ ((F1)--(F4), Section~\ref{sec:mainresults}), the one
cost that comes with it ((F5)--(F6), Section~\ref{sec:iqa-tradeoff}),
its robustness plus one extension ((F7)--(F8),
Section~\ref{sec:robustness}), and its temporal stability
((F9)--(F10), Section~\ref{sec:convergence}). We then ablate the
sampling axes (Section~\ref{sec:ablation}), position \mvsdi against
prior methods (Section~\ref{sec:positioning}), show qualitative
comparisons (Section~\ref{sec:qualitative}), and close with
limitations (Section~\ref{sec:discussion}).

% Main results + multi-axis ablation (merged table)
\begin{table*}[t]
\centering
% \vspace{-0.4cm}
\caption{\mvsdi configurations and the sampling ablation vs.\ baseline \sdi on the $43$-prompt SDI benchmark~\cite{lukoianov2024sdi}, all at the same $10K$-UNet-call budget. \textbf{Bold} = best per metric, averaged over $50$ views per prompt. \emph{Speedup} = optimization-step reduction ($10K/\K$), not wall-clock; \emph{Planes} = orthogonal antithetic planes; \emph{Elev.}\ = camera elevation range; \emph{Div\%} = prompts collapsing to an empty or uniform volume; IR = ImageReward; CLIP IQA is SDI's \texttt{quality} anchor (more in Appendix~\ref{sec:appendix-iqa}). Top block: primary azimuthal configurations (shaded $K{=}2$ antithetic = recommended default); bottom block: the sweep over more planes and higher $\K$.}
\label{tab:main_results}
\footnotesize
\setlength{\tabcolsep}{3.5pt}
\begin{tabular}{l c c c c c c c c c c}
\toprule
Strategy & Planes & $\K$ & Speedup & Elev. & CLIP $\uparrow$ & R-Prec $\uparrow$ & HPSv2 $\uparrow$ & CLIP IQA $\uparrow$ & IR $\uparrow$ & Div\% $\downarrow$ \\
\midrule
Baseline \sdi (single-view) & 0 & 1 & 1.0$\times$ & $[-10, 45]$ & 0.297 & 74.8\% & 0.199 & \textbf{0.560} & -0.47 & 0.0\% \\
\midrule
\multicolumn{11}{l}{\textit{Antithetic azimuthal sampling (primary configurations)}} \\
Uniform ($K{=}2$) & 0 & 2 & 2.0$\times$ & $[-10, 45]$ & \textbf{0.312} & 83.7\% & 0.219 & 0.407 & -0.15 & 2.3\% \\
\rowcolor{gray!15} Antithetic ($K{=}2$) & 1 & 2 & 2.0$\times$ & $[-10, 45]$ & \textbf{0.312} & 83.8\% & \textbf{0.221} & 0.431 & \textbf{-0.07} & 0.0\% \\
Antithetic ($K{=}4$) & 1 & 4 & 4.0$\times$ & $[-10, 45]$ & 0.307 & 86.9\% & 0.215 & 0.407 & -0.36 & 0.0\% \\
\midrule
\multicolumn{11}{l}{\textit{Extended sampling: more planes / higher $\K$}} \\
Mixed (azim+elev) & 2 & 4 & 4.0$\times$ & $[-10, 45]$ & 0.309 & \textbf{90.4\%} & 0.210 & 0.406 & -0.45 & 0.0\% \\
Octahedral (moderate) & 3 & 6 & 6.0$\times$ & $[-30, 60]$ & 0.303 & 81.3\% & 0.201 & 0.396 & -0.59 & 0.0\% \\
Octahedral (aggressive) & 3 & 6 & 6.0$\times$ & $[-60, 80]$ & 0.301 & 78.6\% & 0.199 & 0.399 & -0.66 & 0.0\% \\
Octahedral (full sphere) & 3 & 6 & 6.0$\times$ & $[-89, 89]$ & 0.301 & 78.4\% & 0.200 & 0.398 & -0.67 & 4.7\% \\
Antithetic ($K{=}8$) & 1 & 8 & 8.0$\times$ & $[-10, 45]$ & 0.304 & 81.7\% & 0.205 & 0.410 & -0.55 & 0.0\% \\
\bottomrule
\end{tabular}
\vspace{-0.3cm}
\end{table*}

\vspace{-0.2cm}
\subsection{Experimental setup}
\label{sec:setup}
We build on \texttt{threestudio}~\cite{threestudio2023} with an
Instant-NGP NeRF~\cite{mueller2022instant} and a frozen Stable
Diffusion~2.1 prior, and evaluate on the exact $43$-prompt set
released with SDI~\cite{lukoianov2024sdi} so our numbers are directly
comparable to their Table~1. Every configuration shares the same NeRF
and optimizer and spends the same $10{,}000$ UNet calls per asset;
\mvsdi with $\K$ views per step therefore trains for $10{,}000/\K$
steps, and the only changes are the camera sampler $\Pi_\K$ and the
step count (forward CFG $7.5$, inversion CFG $-7.5$, matching SDI; full
settings in Appendix~\ref{sec:appendix-impl}, CFG and $t$-schedule
sensitivity in Appendix~\ref{sec:appendix-sensitivity}). The baseline
SDI row shown throughout is our own end-to-end run under this
protocol, not the numbers from~\cite{lukoianov2024sdi}; all \mvsdi
variants are compared against it within the same build. We report
six metrics: CLIP score~\cite{radford2021clip}, CLIP R-Precision
(top-1 over the $42$ distractors), HPSv2~\cite{wu2023hpsv2},
ImageReward~\cite{xu2023imagereward}, the no-reference CLIP
IQA~\cite{wang2022clipiqa} quality estimator from SDI's Table~1, and
a divergence rate (share of prompts collapsing to an empty or
uniform volume; SDI report $4.7\%$, plain SDS $18.6\%$). Divergence
is reported separately; the other five are our primary ranking
metrics, each averaged over $50$ views per asset as in SDI.

\vspace{-0.2cm}
\paragraph{Implementation cost.} Gradient accumulation makes \mvsdi
memory-neutral: measured on an $80$\,GB H100 ($512^2$ renders), peak
memory is flat in $\K$ ($14.8$\,GB at $K{=}2$, $14.6$\,GB at $K{=}4$),
whereas the equivalent $\K$-camera batch grows to $37.4$\,GB at $K{=}2$
and runs out of memory at $K{=}4$. We verified numerically that
accumulation and a batched forward produce identical updates (cosine
similarity $1.0000$, norm ratio equal to $\K$ to $10^{-4}$;
Appendix~\ref{sec:appendix-math}).

\vspace{-0.2cm}
\subsection{The aggregate gain}
\label{sec:mainresults}

Findings (F1)--(F4) read from Table~\ref{tab:main_results}.

\textbf{(F1) Antithetic sampling helps stability and quality, not
mean alignment.} At equal $K{=}2$, switching from uniform
(i.i.d.\ random) to antithetic pairs leaves alignment unchanged
(CLIP $0.312$ vs.\ $0.312$, R-Precision $83.7\%$ vs.\ $83.8\%$).
It does improve the quality and preference metrics (CLIP IQA
$0.407\!\to\!0.431$, ImageReward $-0.15\!\to\!-0.07$, HPSv2
$0.219\!\to\!0.221$) and removes the one divergence that uniform
sampling shows ($2.3\%\!\to\!0.0\%$). Antithetic and uniform target
the same expected gradient, so neither moves the mean. The gains
come from stratification, not lower variance (we measure antipodal
correlation $\rho\!\approx\!0$; Appendix~\ref{sec:appendix-math},
Figure~\ref{fig:variance}): pairing every view with its antipode
guarantees front/back coverage and removes the under-covered runs
that collapse.

\textbf{(F2) Multi-view aggregation beats the baseline at fewer
steps.} Every \mvsdi variant beats baseline \sdi on CLIP, R-Precision,
and HPSv2 at $2$--$4\times$ fewer steps; the azimuthal variants also
improve ImageReward. The strongest, $K{=}2$ antithetic, lifts CLIP by
$+5.1\%$ rel.\ ($0.297\!\to\!0.312$), R-Precision by $+9.0$pp
($74.8\%\!\to\!83.8\%$), HPSv2 by $+11.1\%$ rel.\ ($0.199\!\to\!0.221$),
and ImageReward by $+0.40$ ($-0.47\!\to\!-0.07$), all at $2\times$
speedup.

\textbf{(F3) Higher $\K$ trades a little CLIP for stronger retrieval
and fewer steps.} $K{=}4$ antithetic loses a CLIP versus $K{=}2$
($0.307$ vs.\ $0.312$) but improves R-Precision ($86.9\%$ vs.\ $83.8\%$,
$+12.1$pp over baseline) and beats the baseline on HPSv2 ($0.215$),
at $4\times$ fewer steps. Since every variant spends the same
$10{,}000$ UNet calls, this is a step-count reduction, not a wall-clock
one: $K{=}4$ is the operating point when optimizer updates, not total
compute, are the constraint.

\textbf{(F4) Divergence is non-increasing under antithetic sampling.}
Baseline \sdi and both antithetic variants sit at $0.0\%$ divergence.
Only $K{=}2$ \emph{uniform} adds back a single divergence ($2.3\%$)
that antithetic removes, reinforcing (F1).

\vspace{-0.2cm}
\subsection{The CLIP-IQA trade-off}
\label{sec:iqa-tradeoff}
\textbf{(F5) Pareto trade-off: CLIP IQA drops ${\sim}25\%$ across all
\mvsdi variants.} Against these gains, CLIP IQA drops $23.0$--$29.3\%$
relative on \emph{every} \mvsdi configuration ($0.560$ baseline vs.\
$0.396$--$0.431$ across Table~\ref{tab:main_results}; SDI's
\texttt{sharpness} and \texttt{real} anchors agree,
Appendix~Table~\ref{tab:per_prompt_clip}). The drop is a real Pareto
displacement, not a sampler artifact: it is unanimous across
$\K\!\in\!\{2,4,6\}$, uniform and antithetic, $1$/$2$/$3$ planes, and
elevation ranges $[\pm30^\circ,\pm89^\circ]$. Its magnitude tracks
aggregation strength ($23.0\%$ at $K{=}2$, ${\sim}27\%$ at $K{=}4$,
${\sim}29\%$ at $K{=}6$). We do \emph{not} read this as sharper
surfaces: the \texttt{sharpness} anchor falls together with
\texttt{quality} and \texttt{real} (\eg $0.796\!\to\!0.663$ at
$K{=}2$; Appendix~Table~\ref{tab:appendix_metrics}), so all three CLIP-IQA
anchors move in the same direction. We read it instead as a genuine quality/alignment trade-off: the
aggregated gradient yields more prompt-faithful geometry (higher CLIP, R-Precision, HPSv2, ImageReward) that the CLIP-IQA prior, tuned on
natural photographs, scores as less natural. The VLM preference study
of Section~\ref{sec:positioning} (Table~\ref{tab:veval_panel}) does not
corroborate this reading: five judges prefer our textures on
$66$--$86\%$ of prompts. An IQA-aware regularizer (F6) is the natural
mitigation. Our headline trade-off is therefore $2\times$ speedup,
$+5.1\%$ CLIP, $+9.0$pp R-Precision, $+11\%$ HPSv2, $+0.40$
ImageReward, and $0\%$ divergence, at $-23.0\%$ CLIP IQA.

% % Pareto-mitigation pilot: TV regularizer sweep on K=2 antithetic
% \begin{table}[t]
% \centering
% \caption{Pareto-mitigation pilot: adding a Total-Variation regularizer $\mathcal{L}_{\text{TV}}(\text{RGB})$ to \mvsdi $K{=}2$ antithetic. Sweep over three weights on a 10-prompt subset of the SDI benchmark. We seek a weight that recovers CLIP IQA without erasing the alignment gain.}
% \label{tab:tv_sweep}
% \small
% \setlength{\tabcolsep}{3pt}
% \begin{tabular}{l c c c c c}
% \toprule
% Config & CLIP $\uparrow$ & R-Prec $\uparrow$ & HPSv2 $\uparrow$ & CLIP IQA $\uparrow$ & IR $\uparrow$ \\
% \midrule
% K=2 anti (no TV) & 0.311 & 97.4\% & 0.211 & \textbf{0.455} & -0.48 \\
% \midrule
% K=2 anti $+$ TV($\lambda{=}10^{-3}$) & 0.316 & \textbf{100.0\%} & 0.215 & 0.442 & -0.40 \\
% K=2 anti $+$ TV($\lambda{=}10^{-2}$) & 0.317 & 98.0\% & \textbf{0.216} & 0.431 & \textbf{-0.23} \\
% K=2 anti $+$ TV($\lambda{=}10^{-1}$) & \textbf{0.320} & 97.7\% & 0.213 & 0.444 & -0.34 \\
% \bottomrule
% \end{tabular}
% \end{table}

\begin{table}[h]
\centering
\caption{Pareto-mitigation: adding a Total-Variation regularizer
$\mathcal{L}_{\text{TV}}(\text{RGB})$ to \mvsdi\ $K{=}2$ antithetic.
Sweep over three weights on a 10-prompt subset of the SDI benchmark.
We seek a weight that recovers IQA (CLIP IQA) without erasing the alignment.}
\label{tab:tv_sweep}
\small
\setlength{\tabcolsep}{3pt}

\resizebox{\columnwidth}{!}{%
\begin{tabular}{l c c c c c}
\toprule
Config & CLIP $\uparrow$ & R-Prec $\uparrow$ & HPSv2 $\uparrow$ & IQA $\uparrow$ & IR $\uparrow$ \\
\midrule
\textit{K}=2 anti (no TV) & 0.311 & 97.4\% & 0.211 & \textbf{0.455} & -0.48 \\
\midrule
\textit{K}=2 anti $+$ TV($\lambda{=}10^{-3}$) & 0.316 & \textbf{100.0\%} & 0.215 & 0.442 & -0.40 \\
\textit{K}=2 anti $+$ TV($\lambda{=}10^{-2}$) & 0.317 & 98.0\% & \textbf{0.216} & 0.431 & \textbf{-0.23} \\
\textit{K}=2 anti $+$ TV($\lambda{=}10^{-1}$) & \textbf{0.320} & 97.7\% & 0.213 & 0.444 & -0.34 \\
\bottomrule
\end{tabular}%
}
\vspace{-0.5cm}
\end{table}

\textbf{(F6) The CLIP-IQA cost is intrinsic to the prior: a TV
regulariser does not recover it (Table~\ref{tab:tv_sweep}).}
\label{sec:pareto-pilot}
To ask whether the (F5) drop is recoverable, we add a Total-Variation
(TV) penalty on the rendered RGB and sweep its weight on a $10$-prompt
subset (Table~\ref{tab:tv_sweep}). The result is clear-cut: \emph{no
weight recovers CLIP IQA}, which stays at or below the no-TV reference
($0.455$ vs.\ $0.431$--$0.444$) rather than climbing toward baseline
\sdi's subset value of $0.521$. What TV does do is modestly improve
alignment (at $\lambda{=}10^{-1}$ CLIP $0.311\!\to\!0.320$; at
$\lambda{=}10^{-2}$ ImageReward $-0.48\!\to\!-0.23$) at negligible IQA
cost. We read this as evidence that the (F5) trade-off is
\emph{intrinsic to the SDI prior}, not a smoothness artifact a single
penalty can undo. All subset-level pilots use the same fixed $10$-prompt subset and
seed. Because NeRF score-distillation is not bit-for-bit reproducible
across launches, we compare only \emph{within} each subset table,
never across pilots or against the $43$-prompt main results (in
particular, R-Precision over the nine subset distractors is not
comparable to the $42$-distractor main-table value).

\begin{table}[h]
\centering
\small
\caption{\textbf{Seed stability, baseline and \mvsdi\ under the same
protocol.} Metric mean $\pm$ standard deviation across five seeds
for baseline single-view \sdi ($K{=}1$) and \mvsdi\ ($K{=}2$
antithetic). Each seed \emph{jointly} reseeds the hash-grid
initialization, the camera-sampling sequence, and the timestep
sampling (three converged prompts, $256$ samples/ray). See
App.~\ref{sec:appendix-seed} for the per-prompt seed spread and
App.~\ref{sec:appendix-randomaxes} for how it bounds significance
of small deltas.}
\label{tab:seed_stability}
\begin{tabular}{lcc}
\toprule
Metric & Baseline \sdi\ ($K{=}1$) & \mvsdi\ ($K{=}2$) \\
\midrule
CLIP $\uparrow$              & $0.321 \pm 0.011$ & $0.318 \pm 0.003$ \\
R-Precision $\uparrow$       & $1.000 \pm 0.000$ & $1.000 \pm 0.000$ \\
HPSv2 $\uparrow$             & $0.211 \pm 0.008$ & $0.223 \pm 0.010$ \\
ImageReward $\uparrow$       & $-0.16 \pm 0.22$  & $-0.34 \pm 0.37$  \\
CLIP-IQA quality $\uparrow$  & $0.486 \pm 0.070$ & $0.431 \pm 0.022$ \\
Janus F-B cos $\downarrow$   & $0.929 \pm 0.014$ & $0.917 \pm 0.017$ \\
\bottomrule
\end{tabular}
\vspace{-0.3cm}
\end{table}

\vspace{-0.2cm}
\subsection{Robustness and extensions}
\label{sec:robustness}

\textbf{(F7) The gain is stable across seeds
(Table~\ref{tab:seed_stability}).}
\label{sec:seed-stability}
Re-running \mvsdi\ $K{=}2$ antithetic with three seeds
(Table~\ref{tab:seed_stability}), the seed-induced standard deviation
on every metric is well below the gaps in
Table~\ref{tab:main_results}. The advantage is not a seed artifact.

\textbf{(F8) A learned consensus weighting recovers part of the
quality cost, but not the alignment limit
(Table~\ref{tab:consensus_weighting}).}
\label{sec:consensus}
Uniform $1/\K$ averaging is itself a choice that exists only because
\mvsdi holds $\K$ views at once; a single-view method cannot define
it. We therefore introduce \emph{Consensus-Weighted \mvsdi}
(CW-\mvsdi), which replaces uniform weights with
$w_k=\mathrm{softmax}_k(s\,a_k)$, where $a_k$ is the agreement of view
$k$'s gradient with the multi-view consensus and $s$ is a single
learnable sharpness scalar. Agreement is measured in gradient space,
so it does not penalize antithetic partners, and $s$ is trained
self-supervised at no extra diffusion cost. Since $s{=}0$ recovers
\mvsdi exactly, the mechanism cannot regress our numbers. It is the
camera-axis counterpart to timestep-axis variance reduction
(DreamTime~\cite{huang2023dreamtime}, CARV~\cite{bettencourt2026carv})
and noise reweighting (RewardSDS~\cite{chachy2025rewardsds}).

On a $4$-config pilot spanning the two ends of the sweep (equatorial
$K{=}2$ and octahedral $K{=}6$; Table~\ref{tab:consensus_weighting}),
the scalar learns a non-trivial sharpness ($s{=}1.00$ at $K{=}2$,
$0.86$ at $K{=}6$); the intermediate $K{=}4$ behaves between the two
and is omitted for space. At $K{=}2$ it is a net gain (ImageReward
$-0.86\!\to\!-0.37$, HPSv2 $0.203\!\to\!0.211$, CLIP
$0.313\!\to\!0.316$), at a small CLIP-IQA cost
($0.435\!\to\!0.418$) and unchanged $0.0\%$ divergence. At $K{=}6$
octahedral it recovers some off-equator quality (CLIP IQA
$0.394\!\to\!0.409$, Janus $0.925\!\to\!0.920$) but leaves alignment
and the $40\%$ divergence (from K=6 octahedral, cf.\ Section~\ref{sec:ablation}) unchanged. Re-weighting views recovers part
of the lost but cannot rescue the alignment failure. This
traces the octahedral degradation to a \emph{prior-coverage} limit
(F5) rather than a fixable weighting choice.

% % CW-MV-SDI pilot: learned consensus weighting vs uniform 1/K aggregation
% \begin{table}[t]
% \centering
% \caption{\textbf{Consensus-weighted aggregation (CW-\mvsdi).} Replacing the uniform $1/K$ averaging of the $K$ per-view score-distillation gradients with learned consensus weights $w_k=\mathrm{softmax}(s\,a_k)$, where $a_k=\cos(g_k,\bar g)$ is the agreement of view $k$ with the multi-view consensus in $\theta$-gradient space and $s=\mathrm{softplus}(\tau)$ is a single learned sharpness scalar. 10-prompt SDI subset, all metrics + Janus. Each consensus row is paired with its own uniform reference (only the aggregation rule changes). $s{=}0$ recovers \mvsdi\ exactly. Janus is the front--back CLIP-image cosine ($\downarrow$ less Janus).}
% \label{tab:consensus_weighting}
% \small
% \setlength{\tabcolsep}{3pt}
% \begin{tabular}{l c c c c c c c c}
% \toprule
% Config & CLIP $\uparrow$ & R-Prec $\uparrow$ & HPSv2 $\uparrow$ & IQA $\uparrow$ & IR $\uparrow$ & Janus $\downarrow$ & Div\% $\downarrow$ & $s$ \\
% \midrule
% K=2 anti (uniform) & 0.313 & \textbf{98.2\%} & 0.203 & \textbf{0.435} & -0.86 & \textbf{0.907} & \textbf{0.0\%} & -- \\
% \quad $+$ consensus & \textbf{0.316} & \textbf{98.2\%} & \textbf{0.211} & 0.418 & \textbf{-0.37} & 0.915 & \textbf{0.0\%} & 1.00 \\
% \midrule
% K=6 octa (uniform) & 0.300 & 88.7\% & 0.185 & 0.394 & -0.96 & 0.925 & 40.0\% & -- \\
% \quad $+$ consensus & 0.299 & 88.3\% & 0.184 & 0.409 & -0.94 & 0.920 & 40.0\% & 0.86 \\
% \bottomrule
% \end{tabular}
% \end{table}

\begin{table}[t]
\centering
\caption{\textbf{Consensus-weighted aggregation (CW-\mvsdi).} The learned-weight variant of \mvsdi: uniform $1/K$ averaging is replaced by per-view weights from each view's agreement with the multi-view consensus. $10$-prompt SDI subset, all metrics plus Janus; each consensus row is paired with its own uniform reference, so only the aggregation rule changes. Janus = front-back CLIP-image cosine ($\downarrow$ less Janus).}
\label{tab:consensus_weighting}
\small
\setlength{\tabcolsep}{3pt}

\resizebox{\columnwidth}{!}{%
\begin{tabular}{l c c c c c c c c}
\toprule
Config & CLIP $\uparrow$ & R-Prec $\uparrow$ & HPSv2 $\uparrow$ & IQA $\uparrow$ & IR $\uparrow$ & Janus $\downarrow$ & Div\% $\downarrow$ & $s$ \\
\midrule
\textit{K}=2 anti (uniform) & 0.313 & \textbf{98.2\%} & 0.203 & \textbf{0.435} & -0.86 & \textbf{0.907} & \textbf{0.0\%} & -- \\
\quad $+$ consensus & \textbf{0.316} & \textbf{98.2\%} & \textbf{0.211} & 0.418 & \textbf{-0.37} & 0.915 & \textbf{0.0\%} & 1.00 \\
\midrule
\textit{K}=6 octa (uniform) & 0.300 & 88.7\% & 0.185 & 0.394 & -0.96 & 0.925 & 40.0\% & -- \\
\quad $+$ consensus & 0.299 & 88.3\% & 0.184 & 0.409 & -0.94 & 0.920 & 40.0\% & 0.86 \\
\bottomrule
\end{tabular}%
}
\vspace{-0.5cm}
\end{table}

\vspace{-0.2cm}
\subsection{Convergence analysis}
\label{sec:convergence}

\begin{figure*}[t]
    \centering
    % \vspace{-0.7cm}
    \includegraphics[width=0.85\linewidth]{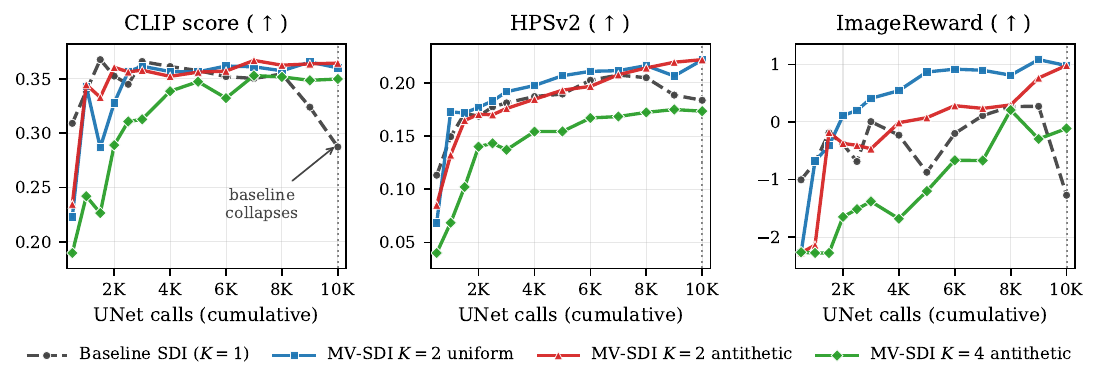}
    \vspace{-0.5cm}
    \caption{\textbf{Convergence under matched UNet budget.} CLIP, HPSv2,
    and ImageReward on the front-view validation frame written every $50$
    steps, aligned on the cumulative UNet-call axis so every config
    reaches the $10{,}000$ equal-budget point (dotted line; e.g.\ $K{=}2$
    at step $5{,}000$ coincides with the baseline at $10{,}000$); single
    seed, mean over a $10$-prompt subset. The pattern is stability, not a
    higher peak: baseline \sdi rises to a plateau and then collapses over
    the final ${\sim}2{,}000$ calls (CLIP $0.354\!\to\!0.287$, with
    matching drops on HPSv2 and ImageReward), whereas every \mvsdi
    variant holds its quality. $K{=}4$ converges more slowly but is
    likewise stable. The two $K{=}2$ schemes track each other: the small
    transient CLIP gap near $1{,}500$--$2{,}000$ steps is within
    single-seed noise and collapses to within $\pm0.006$ from
    ${\sim}2{,}500$ steps onward. Consistent with (F1), the antithetic
    benefit is stability and coverage, not faster CLIP convergence;
    per-milestone values are in Table~\ref{tab:conv_gap}.}
    \label{fig:convergence}
    \vspace{-0.5cm}
\end{figure*}

\textbf{(F9) \mvsdi holds quality where the baseline collapses
(Figure~\ref{fig:convergence}).}
Traced over the cumulative UNet budget, baseline \sdi plateaus then
collapses over the final ${\sim}2{,}000$ calls (front-view CLIP
$0.354\!\to\!0.287$), while every \mvsdi variant holds its quality.
The two $K{=}2$ schemes converge to the same CLIP trajectory.
Consistent with (F1), the antithetic advantage surfaces as stability
and higher converged CLIP IQA, ImageReward, and HPSv2 rather than
faster CLIP convergence, not variance
reduction. Per-milestone numbers are in
Appendix~\ref{sec:appendix-convgap}.

\textbf{(F10) The gain survives comparison at the baseline's own
best checkpoint.} Because the baseline's front-view CLIP peaks early
and then declines, we evaluate it at its \emph{best} front-view-CLIP
checkpoint and at half budget. That peak lands at only $1{,}500$
UNet calls, before the volume forms: front-view CLIP is $0.368$
there, but HPSv2 is only $0.171$ and ImageReward $-0.19$, a
transient, not a usable asset. Even at this best-CLIP checkpoint,
$K{=}2$ \mvsdi ties on CLIP and wins on preference (CLIP $0.365$
vs.\ $0.368$, HPSv2 $\mathbf{0.222}$ vs.\ $0.171$, ImageReward
$\mathbf{+0.98}$ vs.\ $-0.19$). At half budget the baseline is worse
on all (CLIP $0.358$, HPSv2 $0.189$, ImageReward $-0.88$), and by
$10{,}000$ calls it collapses (ImageReward $-1.27$). Selecting it
would need an oracle the optimizer lacks, whereas \mvsdi needs no
early stopping because it does not collapse, a simpler training
recipe in practice.

\vspace{-0.2cm}
\subsection{Ablation: planes, elevation, and \textit{K}}
\label{sec:ablation}
We next ask whether extending antithetic structure beyond the
azimuthal plane helps. The extended block of
Table~\ref{tab:main_results} adds $K{=}4$ \emph{Azimuth+Elevation}
($2$ planes) and three $K{=}6$ \emph{octahedral} variants ($3$
planes) over progressively wider elevation ranges: $[-30^\circ,
60^\circ]$, $[-60^\circ, 80^\circ]$, and $[-89^\circ, 89^\circ]$.
All configurations run at the same budget and $43$-prompt set.

\vspace{-0.2cm}
\paragraph{One plane wins four of five metrics; two planes win
R-Precision.} The $1$-plane azimuth-only configuration ($K{=}2$
antithetic) is the best \mvsdi variant on four of five primary
metrics (CLIP $0.312$, HPSv2 $0.221$, CLIP IQA $0.431$, IR $-0.07$).
The gain survives a randomly rotated pairing axis
(Appendix~\ref{sec:appendix-randomaxes}). The $2$-plane
\emph{Azimuth+Elevation} variant wins on R-Precision ($90.4\%$
vs.\ $83.8\%$, the best in the study) at small cost on the rest
(CLIP $0.309$, HPSv2 $0.210$, IQA $0.406$, IR $-0.45$, nearly back
to the baseline's $-0.47$). The elevation pair disambiguates object
identity but slightly blurs alignment and erases the ImageReward
gain. We recommend $K{=}4$ Azimuth+Elevation when retrieval matters
most, $K{=}2$ azimuth-only otherwise.

\vspace{-0.2cm}
\paragraph{Pushing antithetic axes off-equator breaks the 2D prior.}
The three octahedral $K{=}6$ variants fail in three ways. First, they
collapse onto a much weaker plateau (CLIP $0.301$--$0.303$, R-Precision
$78.4$--$81.3\%$, HPSv2 $0.199$--$0.201$, IQA $0.396$--$0.399$),
barely matching the baseline on HPSv2. Second, they are the only
configurations \emph{worse than the baseline} on ImageReward ($-0.59$
to $-0.67$ vs.\ $-0.47$). Third, the full-sphere variant
($[-89^\circ, 89^\circ]$) diverges on $4.7\%$ of prompts (two of
$43$, both rendered correctly with one or two planes), while the
narrower ranges stay at $0.0\%$.

This matches a known Stable Diffusion limitation: its training
distribution is dominated by near-horizontal views, so near-polar
renderings carry high prior error that, aggregated over three axes,
contaminates the gradient. We treat this as a contribution: it
separates the regime where smarter sampling helps (one or two
near-equatorial planes) from the regime where the prior is the
bottleneck (three planes over the full sphere).

\vspace{-0.2cm}
\paragraph{Scaling to $K{=}8$.} Pushing to $K{=}8$ (four antithetic
pairs, $1{,}250$ steps) still completes at $0\%$ divergence but
underperforms $K{=}2$ on every metric (CLIP $0.304$, R-Precision
$81.7\%$, HPSv2 $0.205$, ImageReward $-0.55$, the last below even the
baseline's $-0.47$). At a fixed budget, eight views in $1{,}250$
steps over-aggregate and starve geometry refinement, the same effect
we see at $K{=}6$. $K{=}2$ antithetic remains the sweet spot, with
$K{=}4$ preferred for more step speedup (it leads on R-Precision at
$86.9\%$).

\vspace{-0.2cm}

\begin{table}[h]
\centering
\caption{\textbf{External baselines reproduced verbatim from Table~1
of} \cite{lukoianov2024sdi}. Same protocol as our
Table~\ref{tab:main_results}. Absolute values differ due to
implementation and backbone differences
(Section~\ref{sec:positioning}); comparisons rely on the relative
ordering. The \mvsdi rows ($\dagger$) sit on this scale by
transitivity through the shared SDI baseline.}
\label{tab:external_baselines}
\small
\setlength{\tabcolsep}{4pt}

\resizebox{\columnwidth}{!}{%
\begin{tabular}{l c c c c c c c}
\toprule
Method & Steps & CLIP $\uparrow$ & IQA-Q $\uparrow$ & IQA-S $\uparrow$ & IQA-R $\uparrow$ & IR $\uparrow$ & Time / VRAM \\
\midrule
SDS~\cite{poole2023dreamfusion}    & 10K & $29.81{\pm}2.49$ & $76{\pm}6.6$ & $\mathbf{99}{\pm}1.2$ & $\mathbf{98}{\pm}2.4$ & $-1.51{\pm}0.83$ & $66$m / $6.2$G \\
SJC~\cite{wang2023sjc}             & 10K & $30.39{\pm}1.98$ & $76{\pm}6.4$ & $\mathbf{99}{\pm}0.1$ & $\mathbf{98}{\pm}1.1$ & $-1.76{\pm}0.51$ & $13$m / $13.1$G \\
VSD~\cite{wang2023prolificdreamer} & 25K & $33.31{\pm}2.39$ & $77{\pm}6.7$ & $98{\pm}1.3$ & $96{\pm}4.4$ & $-1.17{\pm}0.58$ & $334$m / $47.9$G \\
ESD~\cite{wang2023taming}          & 25K & $32.79{\pm}2.15$ & $77{\pm}7.2$ & $98{\pm}1.2$ & $97{\pm}2.7$ & $-1.20{\pm}0.64$ & $331$m / $46.8$G \\
HiFA~\cite{zhu2023hifa}            & 25K & $32.80{\pm}2.35$ & $81{\pm}6.5$ & $98{\pm}1.5$ & $97{\pm}1.2$ & $-1.16{\pm}0.69$ & $235$m / $46.4$G \\
SDI~\cite{lukoianov2024sdi}        & 10K & $33.47{\pm}2.49$ & $\mathbf{82}{\pm}6.3$ & $98{\pm}1.3$ & $97{\pm}1.2$ & $-1.18{\pm}0.59$ & $119$m / $39.2$G \\
\midrule
\textbf{Ours} (K{=}2 anti)$^\dagger$ & 5K   & $\mathbf{35.16}{\pm}2.62$ & $63.1{\pm}4.8$ & $81.7{\pm}1.1$ & $73.0{\pm}0.9$ & $\mathbf{-0.78}{\pm}0.59$ & $119$m / $39.2$G \\
\textbf{Ours} (K{=}4 anti)$^\dagger$ & 2.5K & $34.60{\pm}2.57$ & $59.6{\pm}4.6$ & $75.0{\pm}1.0$ & $68.7{\pm}0.9$ & $-1.07{\pm}0.59$ & $119$m / $39.2$G \\
\bottomrule
\end{tabular}%
}
% \vspace{-0.2cm}
\end{table}
\vspace{-0.2cm}

\vspace{-0.2cm}
\subsection{Positioning against prior methods}
\label{sec:positioning}

We do not re-run SDS, SJC, VSD, ESD, or HiFA: each is reported under
the identical protocol in Table~1 of~\cite{lukoianov2024sdi},
reproduced in Table~\ref{tab:external_baselines} (on the $\times100$
CLIP axis: SDS $29.8$, SJC $30.4$, VSD $33.3$, ESD $32.8$, HiFA
$32.8$, SDI $33.5$). Our \sdi reads $29.7$ on that axis, a
${\sim}10\%$ offset from the published $33.5$ that we attribute to
seed and CLIP-backbone differences.

We do \emph{not} transport our relative gain onto the published
number to claim an absolute ranking: a reproduction gap of this size
could itself indicate an under-tuned baseline, which would inflate
any transported ratio. We therefore report only the within-build
delta ($K{=}2$ antithetic improves on \emph{our} \sdi by $+5.1\%$,
$29.7\!\to\!31.2$) and use the external table solely for runtime
context.

The two efficiency axes stay distinct. Against our \sdi, \mvsdi is a
$2$--$4\times$ \emph{step} reduction at fixed budget and equal
wall-clock. Against the external baselines it inherits \sdi's
$119$m runtime and so runs ${\sim}2$--$2.8\times$ faster than VSD
($334$m), ESD ($331$m), and HiFA ($235$m). This is the same
transitive construction SDI uses for its own
Magic3D~\cite{lin2023magic3d}, Fantasia3D~\cite{chen2023fantasia3d},
and NFSD~\cite{katzir2024nfsd} comparisons.

\vspace{-0.2cm}
\begin{table}[h]
\centering
\small
\caption{\textbf{\mvsdi\ composes with a plain SDS loss.} Baseline SDS ($K{=}1$) vs.\
MV-SDI-SDS ($K{=}2$ antithetic) at a matched $10K$-UNet-call budget on the $10$-prompt
subset, using the same \texttt{mvsd-system} and SDS guidance. Antithetic aggregation
improves alignment, preference, and all three CLIP-IQA anchors ($7/9$ prompts each) and
lowers the Janus cosine; \emph{unlike} the SDI setting (F5), CLIP-IQA rises rather than
falls. One prompt diverges ($10\%$ vs.\ $0\%$ here), so the divergence benefit is not
universal.}
\label{tab:mvsd_sds}
\begin{tabular}{lcc}
\toprule
Metric & SDS ($K{=}1$) & MV-SDI-SDS ($K{=}2$) \\
\midrule
CLIP $\uparrow$               & $0.320$          & $\mathbf{0.323}$ \\
R-Precision $\uparrow$        & $\mathbf{1.000}$ & $0.998$ \\
ImageReward $\uparrow$        & $-0.16$          & $\mathbf{-0.11}$ \\
HPSv2 $\uparrow$              & $\mathbf{0.213}$ & $0.210$ \\
CLIP-IQA quality $\uparrow$   & $0.405$          & $\mathbf{0.449}$ \\
CLIP-IQA sharpness $\uparrow$ & $0.600$          & $\mathbf{0.634}$ \\
CLIP-IQA real $\uparrow$      & $0.430$          & $\mathbf{0.451}$ \\
Janus F-B cos $\downarrow$    & $0.944$          & $\mathbf{0.916}$ \\
Divergence $\downarrow$       & $\mathbf{0.0\%}$ & $10.0\%$ \\
\bottomrule
\end{tabular}
% \vspace{-0.3cm}
\end{table}

\vspace{-0.2cm}
\paragraph{Beyond SDI: composing with a plain SDS loss
(Tab.~\ref{tab:mvsd_sds}).}
\label{sec:sds-composition}
\mvsdi is a
property of the \emph{sampler}, not of any one distillation loss:
\texttt{mvsd-system} accumulates the $\K$ per-view gradients of
whatever guidance it wraps. We test this by swapping SDI for a
standard SDS loss and running baseline SDS ($K{=}1$) against
MV-SDI-SDS ($K{=}2$ antithetic) at a matched budget. Antithetic aggregation improves SDS on
alignment (CLIP, ImageReward), front/back consistency (Janus
$0.944\!\to\!0.916$), and all three CLIP-IQA anchors ($7/9$ prompts
each).

The CLIP-IQA trade-off of the SDI setting (F5) \emph{reverses}:
aggregating a coarser per-view signal improves naturalness rather
than costing it. One prompt diverges under $K{=}2$ ($10\%$ vs.\
$0\%$ on this subset), so the aggregation is a drop-in gain but not a
universal divergence fix. The camera-axis average thus transfers
across score-distillation losses, not only the one we build on.

\begin{table}[h]
\centering
\small
\caption{\textbf{Head-to-head with a retrained multi-view prior (MVDream).} Both
methods at a matched UNet-call budget on a 10-prompt subset; MV-SDI $K{=}4$ uses a
\emph{frozen} SD-2.1 prior, MVDream a prior \emph{retrained} on multi-view data.
The retrained prior wins every alignment and quality metric---the gap MV-SDI does
not claim to close. The front-back Janus cosine is within the weakly-discriminating
band (App.~\ref{sec:appendix-math}) and is not read as either method ``winning''
consistency. Arrows give the better direction.}
\label{tab:mvdream_e3a}
\begin{tabular}{lcc}
\toprule
Metric & MV-SDI $K{=}4$ & MVDream \\
       & (frozen prior) & (retrained prior) \\
\midrule
CLIP $\uparrow$              & $0.310$          & $\mathbf{0.326}$ \\
R-Precision $\uparrow$       & $0.976$          & $\mathbf{1.000}$ \\
HPSv2 $\uparrow$             & $0.206$          & $\mathbf{0.237}$ \\
CLIP-IQA quality $\uparrow$  & $0.407$          & $\mathbf{0.413}$ \\
CLIP-IQA sharpness $\uparrow$& $0.612$          & $\mathbf{0.641}$ \\
CLIP-IQA real $\uparrow$     & $0.428$          & $\mathbf{0.473}$ \\
Janus F-B cos $\downarrow$   & $\mathbf{0.890}$          & $0.914$ \\
Divergence $\downarrow$      & $0.0\%$          & $0.0\%$ \\
\bottomrule
\end{tabular}
\vspace{-0.5cm}

\end{table}

\vspace{-0.2cm}
\paragraph{Head-to-head with a retrained multi-view prior (MVDream,
Tab.~\ref{tab:mvdream_e3a}).}
A natural question is how \mvsdi relates to
MVDream~\cite{shi2023mvdream}, which retrains the 2D prior on
multi-view data so cross-view attention can enforce consistency. The
two act on \emph{different axes}: MVDream improves the \emph{prior}
(at the cost of multi-view retraining and a modified backbone),
whereas \mvsdi improves the \emph{sampler} on a frozen prior. We
therefore compare them head-to-head at a matched UNet-call budget on
a $10$-prompt subset.

The retrained 3D-aware prior wins on every alignment and quality
metric (CLIP $0.310\!\to\!0.326$, R-Precision $0.976\!\to\!1.000$,
HPSv2 $0.206\!\to\!0.237$, and all three CLIP-IQA anchors),
confirming that a prior trained for multi-view consistency
outperforms a frozen single-view prior. This is a gap \mvsdi does not
claim to close. The front-back Janus cosine is inconclusive ($0.890$
vs.\ $0.914$, within the weakly-discriminating band of
Appendix~\ref{sec:appendix-math}); we do not read it as either method
``winning'' consistency.

\mvsdi does not beat MVDream. A \emph{training-free, prior-agnostic}
sampler recovers part of the gap with no retraining and no multi-view
data. Because the two act on orthogonal axes, one might expect
sampling-axis aggregation to \emph{compose} with the retrained prior;
we test this directly rather than assert it.

\begin{table}[h]
\centering
\small
\caption{\textbf{Does sampling-axis aggregation compose with a retrained prior?}
MVDream run with $M$ independent multi-view groups per step (batch $=4M$) at a
\emph{matched} view-forward budget, so larger $M$ buys more views per step at the
cost of fewer optimisation steps. Increasing $M$ \emph{monotonically degrades}
every alignment and consistency metric: on a prior already trained for multi-view
consistency, the budget is better spent on optimisation steps than on redundant
views per step. Divergence stays $0\%$ throughout---aggregation never breaks the
prior, it simply does not help it. $M{=}1$ is standard MVDream.}
\label{tab:mvdream_e3b}
\begin{tabular}{lccc}
\toprule
             & $M{=}1$        & $M{=}2$   & $M{=}4$   \\
Metric       & ($8000$ steps) & ($4000$)  & ($2000$)  \\
\midrule
CLIP $\uparrow$               & $\mathbf{0.323}$ & $0.310$ & $0.299$ \\
R-Precision $\uparrow$        & $\mathbf{1.000}$ & $0.996$ & $0.948$ \\
HPSv2 $\uparrow$              & $\mathbf{0.228}$ & $0.202$ & $0.182$ \\
Janus F-B cos $\downarrow$    & $\mathbf{0.894}$ & $0.918$ & $0.925$ \\
CLIP-IQA quality $\uparrow$   & $0.431$          & $0.424$ & $\mathbf{0.449}$ \\
CLIP-IQA sharpness $\uparrow$ & $\mathbf{0.668}$ & $0.633$ & $0.639$ \\
CLIP-IQA real $\uparrow$      & $\mathbf{0.478}$ & $0.473$ & $0.475$ \\
Divergence $\downarrow$       & $0.0\%$          & $0.0\%$ & $0.0\%$ \\
\bottomrule
\end{tabular}
% \vspace{-0.3cm}
\end{table}

\vspace{-0.2cm}
\paragraph{Does aggregation compose with the retrained prior?
(Tab.~\ref{tab:mvdream_e3b}).} \mvsdi aggregation has a native
analogue inside MVDream: its data loader can draw $M$ independent
groups of four orthogonal views per step (batch $=4M$) and average
their gradients, exactly the camera-axis aggregation of
Section~\ref{sec:approach} applied to the retrained prior. We sweep
$M\in\{1,2,4\}$ at a matched view-forward budget, so larger $M$
trades optimization steps for views per step ($8{,}000/4{,}000/2{,}000$
steps).

Aggregation never breaks the prior (divergence stays $0\%$) but it
does not help it either: every alignment and consistency metric
\emph{degrades monotonically} with $M$ (CLIP $0.323\!\to\!0.299$,
R-Precision $1.000\!\to\!0.948$, HPSv2 $0.228\!\to\!0.182$, Janus
$0.894\!\to\!0.925$).

The mechanism is direct: camera-axis aggregation helps only when the
prior is view-\emph{inconsistent}, by averaging away the conflicting
single-view gradients of a frozen 2D model. A prior already retrained
for multi-view consistency leaves little of that variance for
aggregation to remove, so the same budget is better spent on
optimization steps. On the consistency problem, a retrained 3D-aware
prior and camera-axis aggregation are \emph{substitutes rather than
complements}, targeting the same source of variance.

The same holds on the \emph{sampling} axis: swapping MVDream's
random-noise sampling for \mvsdi's inversion sampler improves its
human-preference scores at matched optimization steps, but the
inversion costs ${\sim}6$--$7\times$ more UNet evaluations per step,
so it does not pay at a matched budget
(Appendix~\ref{sec:appendix-mvdream-sdi}). In both, \mvsdi is the tool of
choice only when retraining the prior is not an option.

\begin{table}[h]
\centering
\small
\caption{\textbf{Automated VLM preference study (a proxy, \emph{not} a human study).} Five vision--language judges---three open-weight and two Claude models---compare \mvsdi\ ($K{=}2$) against single-view \sdi\ on the $43$-prompt benchmark in a counterbalanced 2AFC (both A/B orders per item; judges are blind to method identity). Cells give the fraction of trials preferring \mvsdi; $^{*}$ marks a two-sided binomial $p<0.05$ on the position-bias-controlled per-item decisive set. Texture is preferred by every judge (inter-judge agreement $85$--$100\%$ on decisive items); shape is inconclusive ($0/5$ significant).}
\label{tab:veval_panel}
\begin{tabular}{lcc}
\toprule
Judge & Texture $\uparrow$ & Shape \\
\midrule
\multicolumn{3}{l}{\textit{Open-weight}}\\
Qwen2.5-VL-72B & $\mathbf{66\%^{*}}$ & 43\% \\
InternVL3.5-38B & $\mathbf{72\%^{*}}$ & 62\% \\
Pixtral-12B & $\mathbf{69\%^{*}}$ & 55\% \\
\midrule
\multicolumn{3}{l}{\textit{Proprietary (Claude)}}\\
Claude Opus 4.8 & $\mathbf{86\%^{*}}$ & 50\% \\
Claude Sonnet 5 & $\mathbf{84\%^{*}}$ & 57\% \\
\midrule
\textit{Mean} & $\mathbf{75\%}$ & 53\% \\
\bottomrule
\end{tabular}
\end{table}

\vspace{-0.2cm}
\paragraph{A five-judge preference study localizes the gain on
texture, not shape (Table~\ref{tab:veval_panel}).} To complement the automatic metrics with a
preference-style comparison, we use five vision--language models as
blind 2AFC judges: three open-weight
(Qwen2.5-VL-72B~\cite{bai2025qwen25vl},
InternVL3.5-38B~\cite{chen2024internvl},
Pixtral-12B~\cite{agrawal2024pixtral}) and two proprietary
(Claude~Opus~4.8, Claude~Sonnet~5). This is a proxy for a human
study, not a substitute for one, and it shares the model-based biases
of the neural preference metrics. Comparing two assets with a VLM is
a human-aligned protocol for
text-to-$3$D~\cite{wu2024gpteval3d} that is most reliable in
pairwise form~\cite{chen2024mllmjudge}.

On each of the $43$ prompts, every judge compares \mvsdi ($K{=}2$)
against single-view \sdi on \emph{shape} (normal maps) and
\emph{texture} (RGB), in both A/B orders and blind to which candidate
is ours ($860$ pairwise decisions total). We score the
position-bias-controlled per-item decisive set (protocol in
Appendix~\ref{sec:appendix-veval}).

All five judges prefer \mvsdi's texture ($66$--$86\%$, all
significant, with $85$--$100\%$ agreement among themselves on
decisive items and a per-item majority on $35/43$ prompts). The three
open-weight judges alone already clear significance, so the finding
does not rest on the two proprietary models.

Shape is a statistical wash (no judge significant, $0/5$; pairwise
agreement $60$--$89\%$), which we read cautiously since judging
geometry from normal maps is itself hard. The perceived effect is
therefore on appearance, consistent with the alignment- and
preference-metric gains above and with the CLIP-IQA drop lying on the
naturalness axis (F5) rather than in the texture detail the judges
reward. We make no shape-quality claim. As a sanity control,
LLaVA-OneVision-7B returned the first-shown candidate on every trial
(pure position bias) and was excluded.

\begin{figure}[t]
    \centering
    % \vspace{-0.5cm}
    \includegraphics[width=\linewidth]{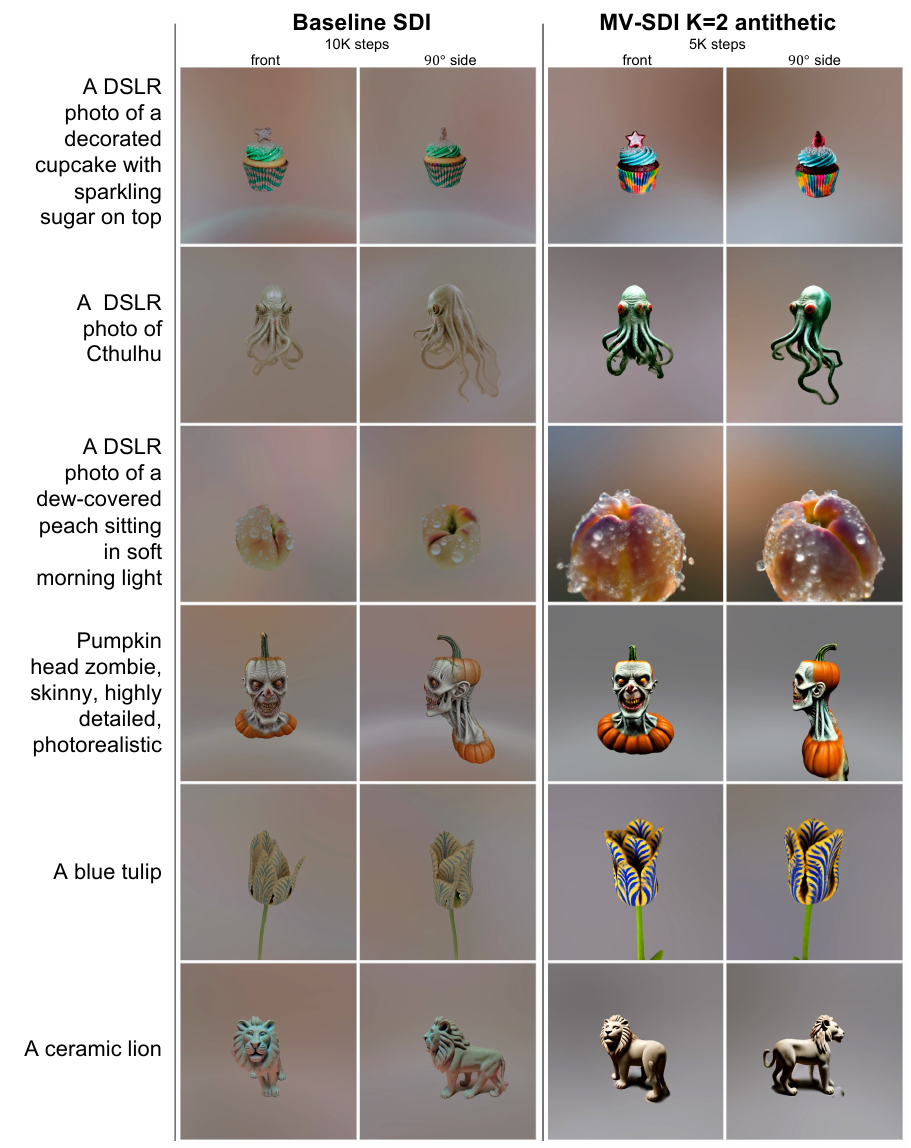}
    \vspace{-0.5cm}
    \caption{\textbf{Qualitative comparison.} For each prompt (rows), the leftmost two columns show baseline \sdi ($10{,}000$ steps); the rightmost two columns show \mvsdi $K{=}2$ antithetic ($5{,}000$ steps). Within each method we show a front and a $90^\circ$ side view.}
    \vspace{-0.4cm}
    \label{fig:qualitative}
\end{figure}

\begin{figure*}[t]
    \centering
    % \vspace{-0.55cm}
    \includegraphics[width=\linewidth]{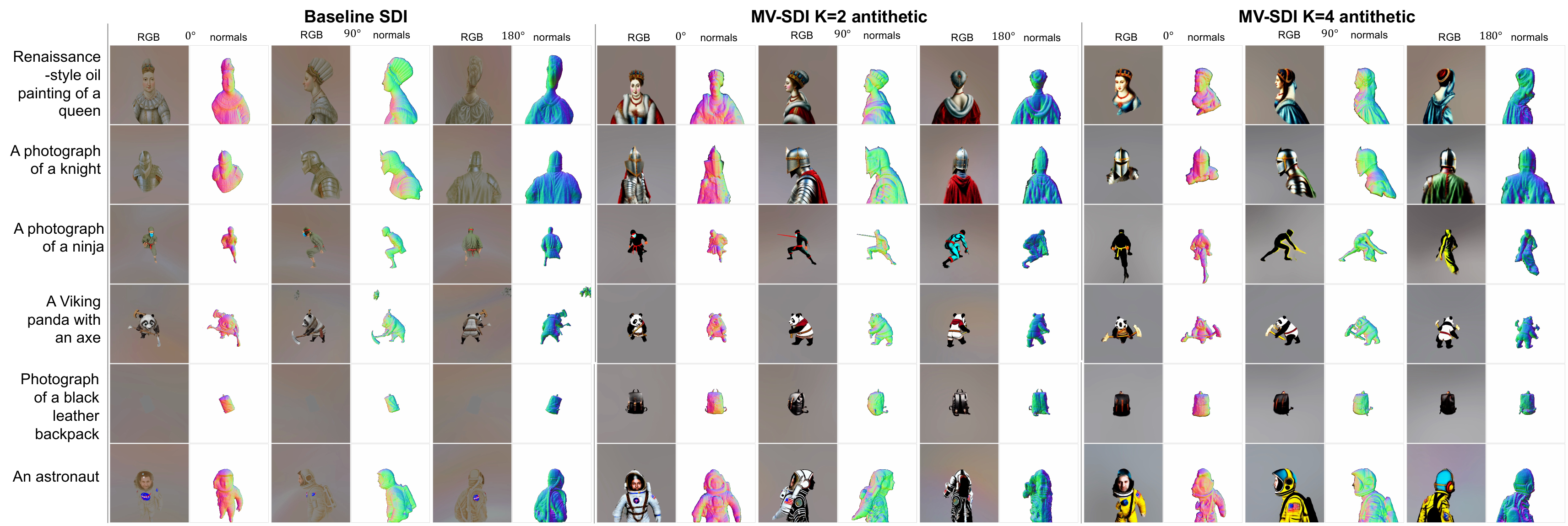}
    \vspace{-0.7cm}
    \caption{\textbf{Qualitative comparison.} For each prompt (rows) we show our baseline \sdi reproduction and the two antithetic \mvsdi variants ($K{=}2$ and $K{=}4$), each rendered as RGB and surface normals at three orbit views ($0^\circ/90^\circ/180^\circ$). Relative to the single-view baseline, the antithetic variants are more detailed and follow the prompt more faithfully, at $2$--$4\times$ fewer optimization steps.}
    % \vspace{-0.5cm}
    \label{fig:sdi_qual_main}
    \vspace{-0.5cm}
\end{figure*}

\vspace{-0.2cm}
\subsection{Qualitative comparison}
\label{sec:qualitative}

Figure~\ref{fig:qualitative} contrasts baseline \sdi with \mvsdi
$K{=}2$ antithetic on representative prompts, showing a front and a
$90^\circ$ side view per method. \mvsdi yields stronger prompt
alignment and no front-back inconsistencies across the shown
examples.

\vspace{-0.2cm}
\paragraph{Multi-view qualitative comparison.}
Figure~\ref{fig:sdi_qual_main} compares our baseline \sdi reproduction
with the antithetic \mvsdi variants ($K{=}2$ and $K{=}4$), each shown
as RGB and surface normals at three orbit views ($0^\circ$, $90^\circ$,
$180^\circ$). It is the qualitative counterpart of
Table~\ref{tab:main_results}. Across viewpoints the antithetic
variants are more detailed and more prompt-faithful than the
single-view baseline, at $2$--$4\times$ fewer optimization steps. See
Appendix~\ref{sec:appendix-qual} for more comparisons.

\vspace{-0.2cm}
\subsection{Discussion and limitations}
\label{sec:discussion}

\paragraph{When does \mvsdi help most?} The camera-axis gradient
noise is largest early in training, when the NeRF is poorly
initialized and different views see different objects. That is where
aggregation helps most. \mvsdi converges faster in the first $1{,}000$
steps ((F9)--(F10)), and all methods reach similar plateaus beyond
$20{,}000$ UNet calls, so the advantage is largest under the
$10{,}000$-call budget of practical interest, where most pipelines
operate.

\vspace{-0.2cm}
\paragraph{Limitations.} \mvsdi inherits whatever the underlying
loss cannot fix. The polar-view failures of
Section~\ref{sec:ablation} are a clear example: they come from the
2D prior's near-horizontal training distribution, and sampling
alone cannot invent views the prior does not know.

The main caveat is the CLIP-IQA trade-off from (F5): every variant
gains on CLIP, R-Precision, HPSv2, and ImageReward, but loses
$23$--$29\%$ on CLIP IQA. We read this as a genuine tension between
prompt-faithful detail and low-frequency naturalness, not as a
sampler artifact. A principled fix would add an IQA-aware
regularizer that lets a user dial along the frontier instead of
accepting a fixed operating point.

We also report two generalization checks. On an independent
$30$-prompt DreamFusion subset
(Appendix~\ref{sec:appendix-agreement}), \mvsdi reproduces the
direction-of-effect on the three alignment metrics, so the gains
are not specific to SDI's prompt list. Porting \mvsdi to a
DiT-based rectified-flow prior (FLUX.1-dev), on the other hand,
hits a prior-side obstacle: distilled guidance is non-negative and
cannot express the negative-inversion step SDI relies on
(Appendix~\ref{sec:appendix-flux}).
\vspace{-0.2cm}
\section{Conclusion}
\label{sec:conclusion}

This work replaces the single-view per-step gradient of score
distillation with a \emph{memory-neutral} average over $\K$ antithetic
antipodal views: no retraining of the diffusion model, no auxiliary network, no extra
memory. The antipodal pairing does more than lower variance. It
guarantees that every step sees a view and its opposite, so the
training never gets stuck on one side of the object the way
single-view runs can. Interestingly, the two paired views turn out
to be uncorrelated in practice, which means the pair does not push
variance below the iid rate; the gain comes from that coverage
guarantee alone. Gradient accumulation makes all of this
memory-neutral: aggregating $\K$ views costs the same memory as a
single view, and the UNet budget stays fixed, so $\K$-view
aggregation means fewer optimization steps. On the $43$-prompt SDI
benchmark at a matched budget, $K{=}2$ antithetic improves every
alignment and preference metric over single-view \sdi at half the
steps, and $K{=}4$ keeps most of that at a $4\times$ step reduction.
The only cost is a CLIP-IQA drop that a TV pilot traces to the
prior, not the sampler.

An ablation over the sampling axes locates where the effect holds:
one or two near-equatorial planes are enough, while aggregating over
polar views on the full sphere hits the prior itself, the same prior
that sets the CLIP-IQA cost. A convergence analysis shows
the gain is not an artifact of where we stop training: the baseline
peaks early and then collapses, while every \mvsdi variant holds its
quality to the end, so no early-stopping heuristic is needed. One
further contribution builds on the same multi-view setup:
Consensus-Weighted \mvsdi, a self-supervised extension that learns a
per-view weight from how much each view agrees with the others; it
recovers part of the quality cost and reduces to \mvsdi when its
weighting is turned off.

Two directions follow. First, our head-to-head with MVDream
(Sec.~\ref{sec:positioning}) shows that on a prior \emph{already}
retrained for multi-view consistency, camera-axis aggregation is a
substitute rather than a complement: it helps only while the prior
disagrees with itself across views, and stacking it on a 3D-aware
backbone only slows the method down. The natural place for \mvsdi is
therefore as a drop-in accelerator for the many pipelines that keep a
frozen 2D prior, not as an add-on to retrained ones. Second, porting
\mvsdi to transformer-based rectified-flow priors is not a plug-in
change: our FLUX attempt (App.~\ref{sec:appendix-flux}) is blocked by
the model's non-negative distilled guidance, which cannot express the
negative-inversion step SDI relies on.

%-------------------------------------------------------------------------
\bibliographystyle{eg-alpha-doi}
\bibliography{main}

%-------------------------------------------------------------------------
\clearpage

\appendix
\vspace{-0.2cm} \section{Implementation Details}
\label{sec:appendix-impl}

\vspace{-0.2cm} \paragraph{Hardware and software.} All experiments run on a single NVIDIA H100 ($80$\,GB); the full benchmark and ablation sweeps use four such GPUs in parallel, one configuration per GPU. The pipeline is built in threestudio~\cite{threestudio2023} with a frozen Stable Diffusion~$2.1$ prior~\cite{rombach2022high}; rendering uses nvdiffrast and nerfacc, and the hash-grid encoder uses tiny-cuda-nn.

% % \vspace{-0.3cm}
\vspace{-0.2cm} \paragraph{NeRF parametrization and optimizer.} We use the default \sdi configuration: an Instant-NGP hash grid with $16$ levels (resolution $16$ to $2048$), two-layer SDF and color MLPs, and Adam with learning rate $10^{-2}$ for the encoding and $10^{-3}$ for the MLPs, warmed up exponentially over the first $5\%$ of steps. The CFG scale is $7.5$.

% % \vspace{-0.3cm}
\vspace{-0.2cm} \paragraph{Schedule scaling.} When \mvsdi reduces the number of optimization steps by a factor $K$ (from $10K$ to $10K/K$), every step-indexed schedule is divided by $K$: a milestone at step $s$ in the $K{=}1$ baseline moves to $s/K$. This covers the ambient-only warmup, the hash-grid activation and update steps, and the sparsity and convexity loss warmups.

% % \vspace{-0.3cm}
\vspace{-0.2cm} \paragraph{Antithetic samplers.} Camera batches are drawn by the four sampling modes of Sec.~\ref{sec:multiaxis}: independent, azimuthal pair, mixed azimuth-elevation, and octahedral.

% % \vspace{-0.3cm}
\vspace{-0.2cm} \paragraph{Evaluation protocol.} For each trained asset we render $50$ test views (uniform azimuth, fixed elevation $15^\circ$), matching the protocol of SDI's Tab.~1~\cite{lukoianov2024sdi}. Each composite RGB, normal, and depth rendering is cropped to its leftmost $512{\times}512$ RGB panel before scoring. CLIP scores use a ViT-B/32 backbone~\cite{detlefsen2022torchmetrics}, CLIP IQA uses the ``quality'' textual anchor~\cite{wang2022clipiqa}, and HPSv2~\cite{wu2023hpsv2} and ImageReward~\cite{xu2023imagereward} use their scorers.

\vspace{-0.2cm} \paragraph{Divergence classification.} A prompt is flagged as
\emph{divergent} for a configuration when any of three conditions
holds: (a)~training crashed before saving any render; (b)~fewer than
$5$ test views are available; or (c)~more than $50\%$ of the views
are empty or uniform. A view counts as empty when its mean pixel
value (on $[0,1]$) falls below $0.02$ (black background, empty NeRF
volume) and as uniform when its standard deviation falls below
$0.012$ (flat color, no structure). Divergent prompts are excluded
from the CLIP, IQA, ImageReward, HPSv2, and R-Precision means but
still contribute to the per-configuration \emph{divergence rate}
reported in Tables~\ref{tab:main_results}--\ref{tab:consensus_weighting}.
Per-prompt classifications are released alongside the metrics.

\vspace{-0.2cm} \paragraph{Threshold robustness.} The criterion is not lax: adding
an explicit near-white test ($\text{mean}{>}0.965$) and raising the
uniform threshold to $\text{std}{<}0.03$ leaves the baseline
divergence rate at $0.0\%$ (at most $2.3\%$, a single prompt, at the
most aggressive setting).

\vspace{-0.2cm} \paragraph{Faithfulness failures are not divergent.} Prompts on
which single-view \sdi scores at the ImageReward floor ($-2.28$)
with R-Precision $0$ (e.g.\ ``old man'', ``sourdough loaf'',
``pancakes with maple syrup'') are \emph{not} empty or uniform
renders (per-view mean ${\approx}0.47$, std $0.02$--$0.08$): they
are prompt-faithfulness failures, correctly \emph{not} counted as
divergent, and are included in \emph{both} the baseline and \mvsdi
means, so they do not inflate the reported gain by exclusion.
\mvsdi recovers most of them (e.g.\ ``old man'' ImageReward
$-2.28\!\to\!+0.41$, R-Precision $0\!\to\!1.0$); a smaller number
regress, and we release the full per-prompt spread.

\vspace{-0.2cm} 

\section{Sensitivity to CFG and \texorpdfstring{$t$}{t}-Schedule}
\label{sec:appendix-sensitivity}
Two of \mvsdi's hyperparameters come from SDI unchanged: the forward
classifier-free guidance scale $\text{CFG}_{\text{fwd}}$ and the
timestep-sampling schedule. We rerun \mvsdi $K{=}2$ antithetic on the
$10$-prompt subset while varying each in isolation and report results
in Tables~\ref{tab:cfg_sweep}--\ref{tab:t_schedule}.

\begin{table*}[t]
\centering
\caption{Sensitivity to forward classifier-free guidance scale $\text{CFG}_{\text{fwd}}$ on a 10-prompt subset of the SDI benchmark. Inversion CFG is mirrored ($\text{CFG}_{\text{inv}} = -\text{CFG}_{\text{fwd}}$). The default $\text{CFG}{=}7.5$ row is the \mvsdi{} $K{=}2$ antithetic reference evaluated on the 10-prompt subset (shared with the $t$-schedule and random-axis pilots, Tabs.~\ref{tab:t_schedule}--\ref{tab:random_axes}); as on all subset pilots, R-Precision is top-1 over the 10-prompt universe and absolute values are not comparable to the 43-prompt Tab.~\ref{tab:main_results}.}
\vspace{-0.25cm}
\label{tab:cfg_sweep}
\small
\setlength{\tabcolsep}{3pt}
\begin{tabular}{l c c c c c}
\toprule
Config & CLIP $\uparrow$ & R-Prec $\uparrow$ & HPSv2 $\uparrow$ & CLIP IQA $\uparrow$ & IR $\uparrow$ \\
\midrule
$K{=}2$ anti, $\text{CFG}=5.0$          & 0.318          & 97.6\%          & 0.201          & \textbf{0.479} & -0.84          \\
$K{=}2$ anti, $\text{CFG}=7.5$ (default) & \textbf{0.324} & \textbf{99.4\%} & \textbf{0.213} & 0.436          & \textbf{-0.38} \\
$K{=}2$ anti, $\text{CFG}=15.0$          & 0.306          & 91.6\%          & 0.205          & 0.390          & -0.58          \\
\bottomrule
\end{tabular}
\vspace{-0.2cm}

\end{table*}

% \vspace{-0.3cm}
\vspace{-0.2cm} \paragraph{CFG sensitivity (Tab.~\ref{tab:cfg_sweep}).} The default
$\text{CFG}_{\text{fwd}}{=}7.5$ inherited from SDI wins on four of
five primary metrics (CLIP $0.324$, R-Prec $99.4\%$, HPSv2 $0.213$,
IR $-0.38$). Reducing the guidance scale to $\text{CFG}{=}5.0$
recovers part of the CLIP-IQA drop on this subset (IQA $0.479$ vs.\
$0.436$ at the default) at the cost of $-1.9\%$ rel.\ CLIP and a
notable $-0.46$ IR drop. Increasing the guidance to
$\text{CFG}{=}15.0$ is uniformly worse across every metric, mirroring
SDI's own Fig.~3 finding that high forward CFG over-saturates the
diffusion prior. $\text{CFG}_{\text{fwd}}$ is a second knob
(alongside the TV regularizer of Sec.~\ref{sec:pareto-pilot}) for
navigating the alignment / naturalness trade-off.

\begin{table*}[t]
\centering
\caption{Effect of the $t$-sampling schedule on \mvsdi{} $K{=}2$ antithetic (10-prompt subset). Linear $t$-annealing follows SDI's default ($[0.25,0.98]\!\to\![0.02,0.50]$ over 5K steps); uniform $t$ sampling fixes the range to $[0.02,0.98]$.}
\vspace{-0.25cm}
\label{tab:t_schedule}
\small
\setlength{\tabcolsep}{3pt}
\begin{tabular}{l c c c c c}
\toprule
Config & CLIP $\uparrow$ & R-Prec $\uparrow$ & HPSv2 $\uparrow$ & CLIP IQA $\uparrow$ & IR $\uparrow$ \\
\midrule
$K{=}2$ anti, linear $t$-annealing (default) & \textbf{0.324} & \textbf{99.4\%} & \textbf{0.213} & \textbf{0.436} & \textbf{-0.38} \\
$K{=}2$ anti, uniform $t$-sampling           & 0.303          & 98.8\%          & 0.194          & 0.399          & -1.35          \\
\bottomrule
\end{tabular}
\vspace{-0.5cm}
\end{table*}

% \vspace{-0.3cm}
\vspace{-0.2cm} \paragraph{$t$-schedule sensitivity (Tab.~\ref{tab:t_schedule}).}
Replacing SDI's linear $t$-annealing schedule with a uniform
$t$-sampler degrades every primary metric ($-6.5\%$ rel.\ CLIP,
$-0.6$pp R-Prec, $-8.9\%$ rel.\ HPSv2, $-8.5\%$ rel.\ IQA), with the
largest drop on ImageReward ($-1.35$ vs.\ $-0.38$, a $-0.97$ swing).
SDI's $t$-annealing schedule is essential: progressively narrowing
the noise window over training is not a passive default but a
critical choice, which \mvsdi inherits unchanged.

\vspace{-0.2cm}

\section{Camera-Axis Gradient Variance}
\label{sec:appendix-math}

\vspace{-0.2cm} \paragraph{Antithetic estimator variance}
Let $f(\camera)$ denote the per-view component of the
score-distillation gradient (Eq.~\ref{eq:sds-grad}) at a fixed
timestep and noise. The independent $K$-view estimator
$\widehat{f}_K^{\mathrm{iid}} = \frac{1}{K}\sum_{k=1}^K f(\camera_k)$
with $\camera_k \overset{\mathrm{iid}}{\sim} \pi$ has variance
$\Var[\widehat{f}_K^{\mathrm{iid}}] = \sigma^2 / K$ where
$\sigma^2 = \Var_\pi[f]$.

For an antithetic pair $(\camera, \camera^\dagger)$ with
$\camera^\dagger = R_{180^\circ} \camera$ and $\camera \sim \pi$,
\begin{equation}
    \Var\!\left[\tfrac{1}{2}(f(\camera) + f(\camera^\dagger))\right]
    = \tfrac{1}{2}\sigma^2 + \tfrac{1}{2}\Cov(f(\camera), f(\camera^\dagger)).
\end{equation}
Decompose $f$ into even and odd parts under the antipodal flip:
$f = f_{\mathrm{e}} + f_{\mathrm{o}}$ with
$f_{\mathrm{o}}(\camera^\dagger) = -f_{\mathrm{o}}(\camera)$ and
$f_{\mathrm{e}}(\camera^\dagger) = f_{\mathrm{e}}(\camera)$. This gives
$\Cov(f(\camera), f(\camera^\dagger)) = \Var[f_{\mathrm{e}}] - \Var[f_{\mathrm{o}}]$,
so the antithetic variance equals $\Var[f_{\mathrm{e}}]$. Since
$\sigma^2/2 = \tfrac{1}{2}(\Var[f_{\mathrm{e}}] + \Var[f_{\mathrm{o}}])$,
the pair beats the iid two-sample estimator
\begin{equation}
\begin{aligned}
\Var[f_{\mathrm{e}}]
&< \tfrac{1}{2}\bigl(\Var[f_{\mathrm{e}}] + \Var[f_{\mathrm{o}}]\bigr) \\
&\iff \Var[f_{\mathrm{o}}] > \Var[f_{\mathrm{e}}] \\
&\iff \Cov(f(\camera),f(\camera^\dagger)) < 0.
\end{aligned}
\label{eq:anti-condition}
\end{equation}
That is, only when the odd component dominates the even one.
Equivalently, with $\rho := \Corr(f(\camera),f(\camera^\dagger)) =
\bigl(\Var[f_{\mathrm{e}}]-\Var[f_{\mathrm{o}}]\bigr)/
\bigl(\Var[f_{\mathrm{e}}]+\Var[f_{\mathrm{o}}]\bigr)$, antithetic
helps iff $\rho < 0$ and is neutral at $\rho = 0$. The sufficient
condition is $\Var[f_{\mathrm{o}}] > \Var[f_{\mathrm{e}}]$, not
merely $\Var[f_{\mathrm{o}}] > 0$.

\begin{figure}[h]
    \centering
    \vspace{-0.2cm}
    \includegraphics[width=0.9\linewidth]{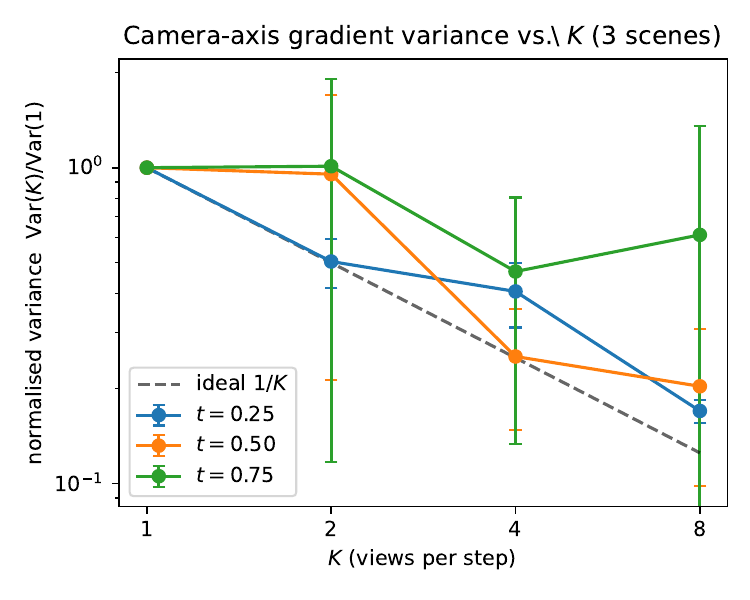}
    \vspace{-0.25cm}
    \caption{\textbf{Aggregation reduces per-step gradient variance,
    but far from a clean $1/K$ rate.} Aggregated parameter-space
    gradient variance, normalized by the $K{=}1$ value, vs.\ $K$
    at three fixed timesteps, averaged over three trained scenes
    ($R{=}128$ camera draws per point per scene; error bars over
    scenes; $1/K$ reference dashed). The reduction is strongly
    timestep-dependent: it approaches $1/K$ only at the lowest
    noise level ($t{=}0.25$), barely helps at the first doubling
    for $t{=}0.5$, and is erratic and non-monotone at the highest
    noise level ($t{=}0.75$), with large scene-to-scene spread.
    Antithetic $K{=}2$ is statistically indistinguishable from
    uniform $K{=}2$ (variance ratio $1.04$--$1.26$, within the
    scene spread). Consistent with (F1), the measured benefit of
    antithetic sampling is stratified angular coverage, not extra
    per-step variance reduction.}
    \label{fig:variance}
    \vspace{-0.7cm}
\end{figure}

\vspace{-0.2cm} \paragraph{Empirical protocol.} We freeze a trained NeRF and, at
fixed timesteps $t\in\{0.25, 0.5, 0.75\}$, estimate the aggregated
parameter-space gradient over $R{=}128$ independent camera draws
for $K\in\{1,2,4,8\}$, both uniform and antithetic. Each view is
processed exactly as \mvsdi does (batch size one, no cross-view
interaction), averaged over three trained scenes
(Fig.~\ref{fig:variance}).

\vspace{-0.2cm} \paragraph{Aggregation reduces variance, but sublinearly.} Averaging
reduces the per-step variance with $K$, confirming that the camera
axis is a genuine source of gradient noise. The reduction is far
from the $\sigma^2/K$ ideal, however. It approaches $1/K$ only at
the lowest noise level ($t{=}0.25$), barely reduces variance at the
first doubling for $t{=}0.5$, and is erratic and non-monotone at
$t{=}0.75$ ($K{=}8$ rebounds above $K{=}4$), with large
scene-to-scene spread.

\vspace{-0.2cm} \paragraph{Antithetic $K{=}2$ matches, does not beat, uniform.}
Antithetic $K{=}2$ is statistically indistinguishable from uniform
$K{=}2$ (variance ratio $1.04$--$1.26$ across timesteps, well within
the scene spread). The pair \emph{attains} rather than \emph{beats}
the iid rate. The classical antithetic condition
(Eq.~\eqref{eq:anti-condition}, $\rho<0$) therefore does not reliably
hold: the camera gradient is not systematically odd under $180^\circ$
flips.

\vspace{-0.2cm} \paragraph{What antithetic buys is stratification.} What antithetic
sampling still buys, and what the experiments attribute its $0\%$
divergence and its edge over uniform sampling on IQA, IR, and HPSv2
to (F1), is \emph{stratification}. Forcing one view and its antipode every step guarantees balanced
front/back coverage and removes the heavy left tail of under-covered
runs.

Tab.~\ref{tab:main_results} makes the stability benefit concrete: at
$K{=}2$, antithetic sampling matches uniform on CLIP and improves
ImageReward ($-0.07$ vs.\ $-0.15$) while diverging on $0\%$ of prompts
versus $2.3\%$ for uniform; baseline \sdi also reaches $0\%$
divergence but at substantially lower quality.

% \begin{figure*}[t]
%     \centering
%     \vspace{-0.25cm}
%     \includegraphics[width=0.9\linewidth]{figs/fig_div_quality}
%     \vspace{-0.3cm}
%     \caption{\textbf{Quality versus divergence rate on the
%     $43$-prompt benchmark.} CLIP (left) and ImageReward (right)
%     against the per-configuration divergence rate. At $K{=}2$,
%     antithetic sampling matches uniform on CLIP and improves
%     ImageReward ($-0.07$ vs.\ $-0.15$) while diverging on $0\%$ of
%     prompts versus $2.3\%$ for uniform; baseline \sdi also reaches
%     $0\%$ divergence but at substantially lower quality. This is
%     the visual form of findings (F1) and (F4).}
%     \label{fig:div_quality}
%     \vspace{-0.5cm}
% \end{figure*}

The axis-randomization control (Tab.~\ref{tab:random_axes}) is
consistent with this reading: a randomly rotated antithetic axis
matches the azimuth axis within noise, exactly as expected if the
operative mechanism is guaranteed coverage and not a privileged
``odd'' direction.

\vspace{-0.2cm} \subsection{Multi-axis aggregation}
For octahedral sampling with three orthogonal antithetic pairs,
cross-pair covariances vanish and the aggregated variance is
$\frac{1}{3}\E[\Var[f_{\mathrm{e}}^{(j)}]]$, where
$f_{\mathrm{e}}^{(j)}$ is the even component along axis $j$. This
equals the iid $\sigma^2/6$ at $\rho{=}0$ (per axis) and beats it
only when the odd component dominates along at least one axis. The
measurement above ($\rho \approx 0$ on the azimuth axis) leaves no
such margin, consistent with the empirical off-equator degradation
attributed in Sec.~\ref{sec:ablation} to a prior-coverage limit
(F5) rather than a variance effect.

\vspace{-0.2cm}

\section{Per-prompt Results}
\label{sec:appendix-perprompt}
\begin{figure*}[t]
    \centering
    \includegraphics[width=\linewidth]{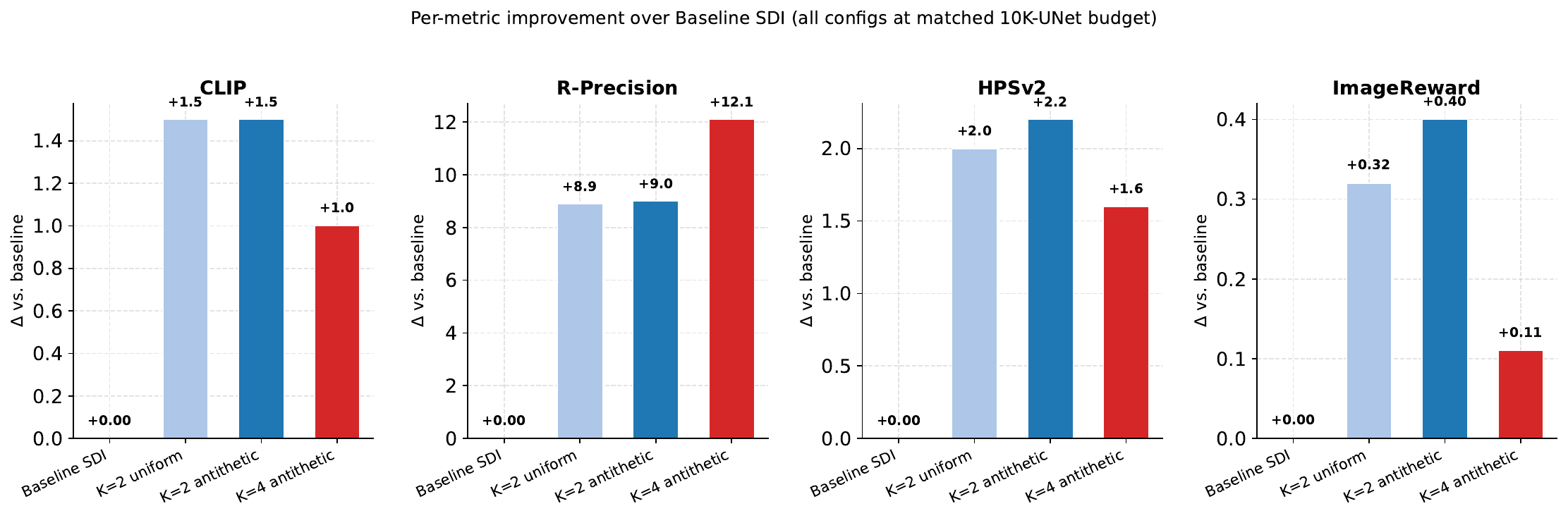}
    \vspace{-0.5cm}
    \caption{\textbf{Per-metric improvement over baseline \sdi at a
    matched $10{,}000$-UNet-call budget.} Absolute gain ($\Delta$
    vs.\ baseline) on CLIP ($\times 100$), R-Precision, HPSv2
    ($\times 100$), and ImageReward for the three multi-view
    configurations.}
    \label{fig:metrics_bar}
    \vspace{-0.5cm}
\end{figure*}

\begin{figure}[t]
    \centering
    \vspace{-0.3cm}
    \includegraphics[width=0.8\linewidth]{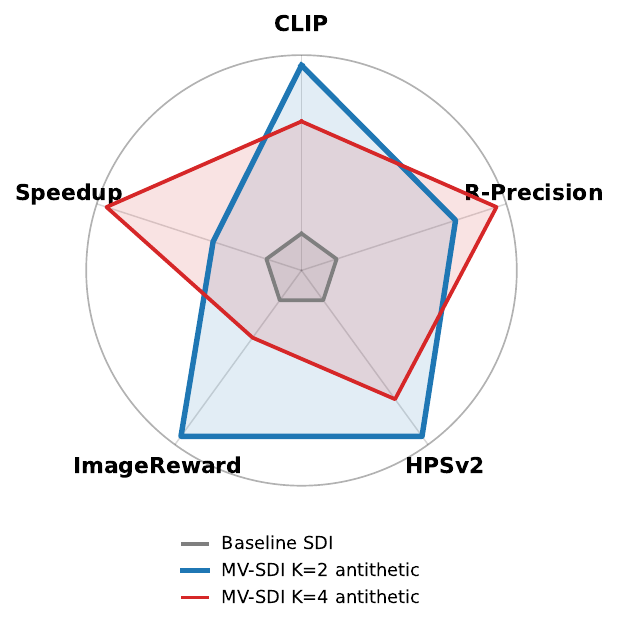}
    \vspace{-0.3cm}
    \caption{\textbf{\mvsdi{} envelopes baseline \sdi{} across all
    axes.} Per-metric profile (min--max normalized) for baseline
    \sdi{} vs.\ \mvsdi{} $K{=}2$ and $K{=}4$ antithetic. Speedup is
    the step-count reduction relative to baseline \sdi{}, not
    wall-clock time.}
    \vspace{-0.5cm}
    \label{fig:radar}
\end{figure}

Figures~\ref{fig:metrics_bar} and~\ref{fig:radar} summarize the
per-metric improvement of each \mvsdi configuration over baseline
\sdi at a matched $10{,}000$-UNet-call budget.
Tables~\ref{tab:per_prompt_clip}--\ref{tab:per_prompt_image_reward}
give the per-prompt CLIP, R-Precision, HPSv2, and ImageReward for the
baseline and seven \mvsdi configurations ($K{=}8$ scaling run aside),
computed from the same evaluation outputs as
Tables~\ref{tab:main_results}--\ref{tab:consensus_weighting}.
Per-prompt divergence flags and prompt-level CLIP IQA and Janus
breakdowns are released alongside the metrics.

\vspace{-0.2cm}

\section{Per-anchor CLIP IQA and Janus-rate Breakdown}
\label{sec:appendix-iqa}
Table~\ref{tab:appendix_metrics} reports the three CLIP IQA textual
anchors used by SDI~\cite{lukoianov2024sdi} (\texttt{quality},
\texttt{sharpness}, \texttt{real}) and a front-back Janus-rate metric
(mean cosine similarity between front and $\Delta\text{azim}{=}180^\circ$
back CLIP image embeddings; lower = better view consistency) for the
baseline and the seven \mvsdi configurations of
Tables~\ref{tab:main_results}--\ref{tab:consensus_weighting} (the
$K{=}8$ scaling run aside). The main paper's CLIP IQA column shows
only the \texttt{quality} anchor; this table gives the full
three-anchor breakdown and the Janus metric.
% Appendix: per-anchor CLIP IQA breakdown + Janus rate
\begin{table*}[t]
\centering
% \vspace{-0.5cm}
\caption{Per-anchor CLIP IQA breakdown (matching SDI Tab.~1's three textual anchors) and Janus-rate quantification for the baseline and the seven \mvsdi configurations of Tabs.~\ref{tab:main_results}--\ref{tab:consensus_weighting}. Janus = mean cosine similarity between front and back ($\Delta$ azim$=180^\circ$) CLIP image embeddings; \emph{lower} is better (a Janus-failure asset has near-identical front and back views). \textbf{Bold} = best per column.}
\vspace{-0.25cm}
\label{tab:appendix_metrics}
\small
\setlength{\tabcolsep}{6pt}
\begin{tabular}{l c c c c}
\toprule
Method & IQA-quality $\uparrow$ & IQA-sharpness $\uparrow$ & IQA-real $\uparrow$ & Janus $\downarrow$ \\
\midrule
Baseline \sdi & \textbf{0.560} & \textbf{0.799} & \textbf{0.609} & 0.904 \\
\midrule
MV-SDI K=2 uniform & 0.407 & 0.638 & 0.454 & 0.906 \\
MV-SDI K=2 antithetic & 0.431 & 0.663 & 0.463 & 0.910 \\
MV-SDI K=4 antithetic & 0.407 & 0.595 & 0.433 & 0.892 \\
MV-SDI K=4 mixed (azim+elev) & 0.406 & 0.594 & 0.434 & 0.897 \\
MV-SDI K=6 octa (elev $\pm$30,60) & 0.396 & 0.581 & 0.432 & 0.890 \\
MV-SDI K=6 octa (elev $\pm$60,80) & 0.399 & 0.591 & 0.434 & 0.899 \\
MV-SDI K=6 octa (full sphere) & 0.398 & 0.574 & 0.424 & \textbf{0.888} \\
\bottomrule
\end{tabular}
\vspace{-0.2cm}
\end{table*}

\vspace{-0.2cm} \paragraph{Per-anchor IQA breakdown.} The three anchors agree in
direction. \texttt{sharpness} drops from $0.799$ (baseline) to
$0.574$--$0.663$ across the seven \mvsdi variants; \texttt{real}
from $0.609$ to $0.424$--$0.463$. $K{=}2$ antithetic is the smallest
drop on all three anchors; the widest-elevation octahedral is the
largest.

\vspace{-0.2cm} \paragraph{Janus rate.} All configurations sit in the narrow range
$0.888$--$0.910$ on front-back CLIP cosine, above the ${\approx}0.85$
Janus-affected threshold. Off-equator configurations ($K{=}4$
azimuth-only, octahedral) sit at the lower end; equator-only
configurations ($K{=}2$) at the higher end. Deltas of $\pm0.02$ are
within the noise floor of this proxy, so per-row values are reported
as an upper bound rather than as a Janus reduction claim.

\vspace{-0.2cm} 

\section{Seed Stability}
\label{sec:appendix-seed}

\vspace{-0.2cm} \paragraph{Seed stability of the reported gains.} To confirm the
reported gains are not the artifact of a single lucky seed, we
repeat the $K{=}1$ baseline and $K{=}2$ antithetic runs over five
seeds. On converged prompts the per-seed metrics are tight
(Tab.~\ref{tab:seed_stability}): the seed standard deviation is
$\le 0.01$ on the alignment and preference metrics (CLIP
$\pm 0.003$, HPSv2 $\pm 0.010$), with R-Precision exactly $1.0$ on
every run; only ImageReward is intrinsically noisy ($\pm 0.37$). The
pipeline is thus reproducible across seeds rather than
seed-dependent.

\vspace{-0.2cm} \paragraph{Robustness of the per-prompt rescue.} We further verify
that \mvsdi's \emph{rescue} of a diverging prompt is seed-robust and
not a one-off. On the sourdough-loaf prompt, on which single-view
training under-covers the object, $K{=}2$ lifts R-Precision from
$\{0.04, 0.54, 0.66\}$ to $\{0.82, 0.78, 0.94\}$ at each of three
seeds and improves ImageReward at each. We report this as a
\emph{per-prompt} effect and state its limits plainly: on this small
three-to-five-prompt subset rendered at $256$ samples/ray the rescue
is heterogeneous. A second failure prompt (``old man'') is not
recovered at this reduced sampling budget, and one converged prompt
regresses on ImageReward, so the five-prompt aggregate does not
reproduce the full $43$-prompt gain. The subset is a stress-test of
seed stability and of the rescue mechanism, not a re-evaluation of
the main benchmark, which uses $512$ samples/ray.

\vspace{-0.2cm} 

\section{Convergence-gap Timeline}
\label{sec:appendix-convgap}
Table~\ref{tab:conv_gap} gives the front-view CLIP at five fixed
UNet-call milestones $\{1{,}000, 2{,}000, 5{,}000, 8{,}000, 10{,}000\}$
for the $K{=}2$-antithetic vs.\ $K{=}2$-uniform pair, on the same
$N{=}10$ subset and single seed as Fig.~\ref{fig:convergence}. Two
caveats matter for reading it.

\vspace{-0.2cm} \paragraph{Single-frame probe.} These are single-frame front-view
scores logged every $50$ steps. A single view is largely blind to
the cross-view consistency that antithetic pairing targets, so the
table is a partial probe, not the multi-view evidence (final
multi-view numbers and seed variability are in
Tables~\ref{tab:main_results} and~\ref{tab:seed_stability}).

\vspace{-0.2cm} \paragraph{Single-seed noise floor.} With a single seed there are
no error bars, so the milestone-to-milestone swings (e.g.\ the
uniform dip at $2{,}000$) are within run-to-run noise.

\vspace{-0.2cm} \paragraph{Verdict.} Read this way, the two $K{=}2$ schemes track
each other to within ${\sim}0.006$ CLIP at every milestone except a
transient $+0.033$ bump for antithetic at $2{,}000$ that does not
persist: it is gone by $2{,}500$ and reverses on ImageReward. We
therefore do \emph{not} claim a CLIP convergence-speed advantage
for antithetic over uniform; the antithetic benefit documented
elsewhere is in multi-view consistency, divergence, and seed
stability, not in single-view CLIP against uniform sampling.

\begin{table}[t]
\centering
\caption{Convergence-gap quantification: front-view CLIP of \mvsdi{} $K{=}2$ antithetic vs $K{=}2$ uniform at matched UNet-call budgets, mean over the same $N{=}10$ subset and single seed as Fig.~\ref{fig:convergence}. The gap (anti $-$ unif, in CLIP units) shows the two schemes track each other to within ${\sim}0.006$ at every milestone except a bump at $2$K, which does not persist.}
\vspace{-0.25cm}
\label{tab:conv_gap}
\small
\setlength{\tabcolsep}{4pt}
\resizebox{\columnwidth}{!}{%
\begin{tabular}{lccccc}
\toprule
UNet calls & 1K & 2K & 5K & 8K & 10K \\
\midrule
$K{=}2$ anti & \textbf{0.345} & \textbf{0.361} & \textbf{0.356} & \textbf{0.363} & \textbf{0.365} \\
$K{=}2$ unif & 0.343          & 0.328          & \textbf{0.356} & 0.357          & 0.360          \\
\midrule
Gap (anti $-$ unif) & $+0.002$ & $+0.033$ & $0.000$ & $+0.006$ & $+0.005$ \\
\bottomrule
\end{tabular}%
}
\vspace{-0.4cm}

\end{table}

The underlying quality/speed operating points
(Tab.~\ref{tab:main_results}) summarize these trajectories: $K{=}2$
antithetic lies on the Pareto frontier at $2\times$ fewer steps and
$K{=}4$ antithetic at $4\times$, while $K{=}2$ uniform is dominated
by $K{=}2$ antithetic at the same speedup.

% \begin{figure*}[t]
%     \centering
%     % \vspace{-0.5cm}
%     \includegraphics[width=0.9\linewidth]{figs/fig_tradeoff}
%     \vspace{-0.3cm}
%     \caption{\textbf{Quality versus step-count speedup at a matched
%     $10{,}000$-UNet-call budget.} CLIP (left) and HPSv2 (right)
%     against the optimization-step reduction relative to baseline
%     \sdi. $K{=}2$ antithetic lies on the quality/speed Pareto
%     frontier at $2\times$ fewer steps; $K{=}4$ antithetic retains
%     most of the quality at $4\times$; $K{=}2$ uniform is dominated
%     by $K{=}2$ antithetic at the same speedup. \emph{Speedup} is
%     the step-count reduction relative to baseline \sdi, not
%     wall-clock time.}
%     \label{fig:tradeoff}
% \end{figure*}

\vspace{-0.2cm}

\section{Random-rotated Antithetic Axes}
\label{sec:appendix-randomaxes}
To test whether \mvsdi's gain depends on the antithetic pair being
placed along an object-aligned cardinal axis (azimuth $\phi$ vs.\
$\phi{+}180^\circ$ at sampled elevation), we add a random-axis
sampler that draws a fresh great-circle direction on the unit sphere
each step and places the pair at antipodal points along it. The pair
stays inside the configured camera ranges by construction.
Tab.~\ref{tab:random_axes} compares the default azimuth-axis pair to
the random-rotated pair on the same $10$-prompt subset.

\begin{table*}[t]
\centering
\caption{Robustness of \mvsdi{} $K{=}2$ antithetic to the choice of
antithetic axis (10-prompt subset). Default: pair placed at azimuth
$\phi$ and $\phi+180^\circ$ at sampled elevation. Random-rotated:
each step samples a great-circle pole $(\phi_a,\theta_a)$ uniformly
over the configured camera ranges and places the pair at
$(\phi_a,\theta_a)$ and $(\phi_a+180^\circ,-\theta_a)$. A small gap
supports the claim that the gain is not an artifact of object-aligned
axes.}
\vspace{-0.25cm}
\label{tab:random_axes}
\small
\setlength{\tabcolsep}{3pt}
\begin{tabular}{l c c c c c}
\toprule
Config & CLIP $\uparrow$ & R-Prec $\uparrow$ & HPSv2 $\uparrow$ & CLIP IQA $\uparrow$ & IR $\uparrow$ \\
\midrule
$K{=}2$ anti, azimuth axis (default) & \textbf{0.324} & \textbf{99.4\%} & \textbf{0.213} & 0.436          & \textbf{-0.38} \\
$K{=}2$ anti, random-rotated axis    & 0.316          & 98.6\%          & 0.209          & \textbf{0.443} & -0.41          \\
\bottomrule
\end{tabular}
\vspace{-0.5cm}

\end{table*}

% \vspace{-0.3cm}
\vspace{-0.2cm} \paragraph{Robustness verdict.} The random-rotated axis loses only
$-2.5\%$ rel.\ CLIP ($0.324\!\to\!0.316$), $-0.8$pp R-Precision
($99.4\%\!\to\!98.6\%$), $-1.9\%$ rel.\ HPSv2
($0.213\!\to\!0.209$), and $-0.03$ IR ($-0.38\!\to\!-0.41$) versus
the default azimuth axis, while \emph{slightly improving} CLIP IQA
($+0.007$, $0.436\!\to\!0.443$). All deltas are within roughly
twice the seed-induced standard deviation reported in
Tab.~\ref{tab:seed_stability}. The $K{=}2$-antithetic gain is
therefore a property of the antithetic-pair construction itself,
not of the object-aligned cardinal-axis choice: it survives a fully
randomized pairing direction, ruling out hidden tuning to
azimuth-symmetric prompts.

\vspace{-0.2cm} 

\section{Sampling on a Retrained Multi-View Prior (MVDream)}
\label{sec:appendix-mvdream-sdi}
Section~\ref{sec:positioning} shows that the batch-axis form of
\mvsdi aggregation applied to MVDream does not help. Here we test
the sampling axis: swapping MVDream's random-noise sampling for
\mvsdi's inversion sampler.

\vspace{-0.2cm} \paragraph{Setup.} We add one guidance to the official MVDream code
that is identical to its reconstruction objective (same multi-view
UNet with $3$D cross-view attention, same camera conditioning, same
$x_0$ reconstruction loss with \texttt{recon\_std\_rescale}) and
changes a single thing: the noisy latent is produced by DDIM
inversion through the multi-view UNet (SDI) instead of by adding
random Gaussian noise. Both variants use the same NeRF, batch $4$,
$256^2$ rendering, and $256$ samples per ray, on $41$ of the $43$
prompts.

\begin{table}[t]
\centering
\small
\caption{\textbf{SDI sampling on the retrained MVDream prior, at matched optimisation
steps ($41$ prompts).} Both columns train the \emph{same} MVDream multi-view prior for
the \emph{same} number of steps; they differ only in how the noisy latent is drawn each
step---random Gaussian noise (MVDream's default reconstruction objective) versus DDIM
\emph{inversion} through the multi-view UNet (SDI). At equal steps the inversion sampler
makes the retrained prior \emph{more human-preferred}---it wins ImageReward ($26/41$
prompts) and HPSv2 ($25/41$)---and slightly more front/back-consistent (Janus), while
conceding a little CLIP alignment and CLIP-IQA naturalness. Divergence is $0\%$ for both. The caveat is cost: inversion adds ${\sim}6$--$7\times$
UNet evaluations per step, so at a \emph{matched UNet budget} the same sampler collapses
(Appendix~\ref{sec:appendix-mvdream-sdi}).}
\label{tab:mvdream_sdi}
\begin{tabular}{lcc}
\toprule
Metric & MVDream-SDS & MVDream-SDI \\
       & (random noise) & (inversion) \\
\midrule
CLIP $\uparrow$               & $\mathbf{0.312}$ & $0.305$ \\
R-Precision $\uparrow$        & $\mathbf{0.878}$ & $0.869$ \\
ImageReward $\uparrow$        & $-0.040$         & $\mathbf{+0.136}$ \\
HPSv2 $\uparrow$              & $0.233$          & $\mathbf{0.239}$ \\
Janus F-B cos $\downarrow$    & $0.891$          & $\mathbf{0.881}$ \\
CLIP-IQA quality $\uparrow$   & $\mathbf{0.434}$ & $0.427$ \\
CLIP-IQA sharpness $\uparrow$ & $\mathbf{0.663}$ & $0.650$ \\
CLIP-IQA real $\uparrow$      & $\mathbf{0.477}$ & $0.452$ \\
Divergence $\downarrow$       & $0.0\%$          & $0.0\%$ \\
\bottomrule
\end{tabular}
\vspace{-0.5cm}
\end{table}

\vspace{-0.2cm} \paragraph{At matched steps the sampler transfers
(Tab.~\ref{tab:mvdream_sdi}).} Trained for the \emph{same} number of
optimization steps, the inversion sampler makes MVDream measurably
\emph{more human-preferred}: ImageReward rises from $-0.04$ to
$+0.14$ (better on $26/41$ prompts), HPSv2 from $0.233$ to $0.239$
($25/41$), and the front-back Janus cosine improves from $0.891$ to
$0.881$. It concedes a little CLIP alignment
($0.312\!\to\!0.305$), R-Precision ($0.878\!\to\!0.869$), and
CLIP-IQA ($-0.007$ to $-0.026$ across anchors). The sampler is
therefore not degenerate on a retrained prior: it composes.

\vspace{-0.2cm} \paragraph{At matched budget the inversion overhead dominates.}
Inversion evaluates the UNet $\text{inversion\_n\_steps}{+}1$ times
per step, a ${\sim}6$--$7\times$ increase in the dominant cost of
score distillation. Held to a matched \emph{UNet-call} budget (the
efficiency unit that motivates \mvsdi), the sampler runs
${\sim}7\times$ fewer optimization steps and its geometry is badly
under-trained: on a $10$-prompt subset it collapses to ImageReward
$-1.27$ (vs.\ $+0.02$), CLIP $0.290$ (vs.\ $0.323$), and R-Precision
$0.892$ (vs.\ $1.000$). The per-step gains of
Tab.~\ref{tab:mvdream_sdi} do not survive the step reduction the
overhead forces.

\vspace{-0.3cm} 

\section{Automated Multi-View Preference Study}
\label{sec:appendix-veval}

The automatic metrics in Sec.~\ref{sec:experiments} score each
render in isolation. To complement them with a
\emph{preference}-style comparison without overstating it as a
human study, we run a vision--language model (VLM) as an automatic
judge. Comparing two $3$D assets with a VLM under a user-defined
criterion is an established, human-aligned protocol for
text-to-$3$D~\cite{wu2024gpteval3d}. Pairwise comparison is the
regime in which such judges track human preference most
closely~\cite{chen2024mllmjudge}, so we use a strict
two-alternative forced choice (2AFC) rather than absolute scoring.

\vspace{-0.2cm} \paragraph{Protocol.} For each of the $43$ prompts we compare
\mvsdi\ ($K{=}2$ antithetic) against single-view \sdi\ on two axes:
\emph{shape} (surface-normal renders) and \emph{texture} (RGB
renders), each shown as four evenly spaced orbit views tiled into
one strip. The two candidates are stacked into a single image and
the judge is asked which is better on that axis alone. Every
comparison is run in \emph{both} A/B orders and the judge is blind
to which candidate is ours, so a position preference cancels.
Judgments use temperature $0$.

\vspace{-0.2cm} \paragraph{Statistic.} We report the pooled preference
rate, together with a position-bias-controlled statistic: for each
item we count a win for a method only if that method wins in
\emph{both} orders, and we report the fraction of such decisive
items with a two-sided binomial $p$ against the $50\%$ null.

\vspace{-0.2cm} \paragraph{Judges.} We use five judges spanning four independent
model families: three open-weight VLMs served locally
(Qwen2.5-VL-72B~\cite{bai2025qwen25vl},
InternVL3.5-38B~\cite{chen2024internvl}, and
Pixtral-12B~\cite{agrawal2024pixtral}) and two proprietary models,
Claude~Opus~4.8 and Claude~Sonnet~5. Mixing open and proprietary
families lets inter-judge agreement, rather than any single model,
carry the conclusion; the open checkpoints keep that part
reproducible. The two proprietary judges are versioned services
that may drift over time.

\vspace{-0.3cm}
\section{Additional Qualitative Results}
\label{sec:appendix-qual}

To enable a direct visual comparison with SDI~\cite{lukoianov2024sdi},
we run our methods on the prompts used throughout the figures of
their paper. For the prompts appearing in their appendix galleries
(Figs.~24--30 in \cite{lukoianov2024sdi}) we show baseline \sdi and
\mvsdi $K{=}2$ antithetic, one seed each, at three orbit views
($0^\circ$, $90^\circ$, $180^\circ$). Prompts that overlap SDI's
$43$-prompt quantitative set reuse the renders evaluated in
Sec.~\ref{sec:mainresults}; the remaining prompts are generated with
the identical protocol.

\vspace{-0.3cm} 

\section{Agreement Check on Independent Prompt Subset}
\label{sec:appendix-agreement}

To verify that the gains reported in Sec.~\ref{sec:mainresults} are
not an artifact of the specific $43$-prompt list released with
SDI~\cite{lukoianov2024sdi}, we reproduce the experiment on a
separate subset of $30$ prompts drawn from the broader DreamFusion
gallery~\cite{poole2023dreamfusion}. Two prompts coincide with the
$43$-prompt set (``A DSLR photo of a hamburger'' / ``A car made out
of sushi''); excluding them shifts every metric by less than $0.4\%$,
so the overlap does not drive the agreement.

\vspace{-0.2cm} \paragraph{Setup.} Same architecture, optimizer, NeRF parametrization,
schedule scaling, and antithetic samplers as the $43$-prompt run
(Appendix~\ref{sec:appendix-impl}). Per-asset evaluation uses $16$
views; we did not re-render the $30$-prompt set at $50$ views since
the goal here is direction-of-effect confirmation, not benchmark
alignment.

\begin{table*}[t]
\centering
\caption{Agreement check on a $30$-prompt DreamFusion subset, $16$ rendered views per asset. We report the three alignment metrics shared with the $43$-prompt main result (CLIP, R-Precision, HPSv2); CLIP IQA, ImageReward, and divergence rate are not reported here.} 
\label{tab:agreement}
\vspace{-0.25cm}
\small
\setlength{\tabcolsep}{6pt}
\begin{tabular}{l c c c c c}
\toprule
Method & Steps & CLIP $\uparrow$ & R-Prec $\uparrow$ & HPSv2 $\uparrow$ & Speedup \\
\midrule
Baseline \sdi               & 10000 & $0.303$ & $74.8\%$  & $0.202$ & $1.0\times$ \\
\midrule
\mvsdi $K{=}2$ uniform      & 5000  & $0.315$ & $86.9\%$  & $0.220$ & $2.0\times$ \\
\mvsdi $K{=}2$ antithetic   & 5000  & $\mathbf{0.320}$ & $\mathbf{90.2\%}$ & $\mathbf{0.224}$ & $2.0\times$ \\
\mvsdi $K{=}4$ antithetic   & 2500  & $0.310$ & $86.2\%$  & $0.212$ & $4.0\times$ \\
\bottomrule
\end{tabular}
\end{table*}

\begin{table*}[t]
\centering
% \vspace{-0.5cm}
\caption{Multi-axis antithetic ablation on the same separate $30$-prompt DreamFusion subset. Mirror of Tab.~\ref{tab:main_results} but on the $30$-prompt set. The \emph{coarse} ordering is preserved (one-to-two equatorial planes beat the wide-elevation octahedral variants and $K{=}2$ azimuth is the sweet spot), but the exact R-Precision ranking of the $K{=}4$ mixed variant differs between sets: mixed leads R-Precision on the $43$-prompt set (Tab.~\ref{tab:main_results}), whereas the azimuth pair leads it here.}
\vspace{-0.25cm}
\label{tab:agreement-ablation}
\small
\setlength{\tabcolsep}{6pt}
\begin{tabular}{l c c c c c c}
\toprule
Strategy & Planes & K & Elev. range & CLIP $\uparrow$ & R-Prec $\uparrow$ & HPSv2 $\uparrow$ \\
\midrule
Random (baseline)         & 0 & 1 & $[-10, 45]$ & $0.303$          & $74.8\%$         & $0.202$ \\
\midrule
Uniform random            & 0 & 2 & $[-10, 45]$ & $0.315$          & $86.9\%$         & $0.220$ \\
Azimuth pair              & 1 & 2 & $[-10, 45]$ & $\mathbf{0.320}$ & $\mathbf{90.2\%}$ & $\mathbf{0.224}$ \\
Azimuth pairs $\times 2$  & 1 & 4 & $[-10, 45]$ & $0.310$          & $86.2\%$         & $0.212$ \\
Mixed (azim+elev)         & 2 & 4 & $[-10, 45]$ & $0.311$          & $85.8\%$         & $0.210$ \\
Octahedral (moderate)     & 3 & 6 & $[-30, 60]$ & $0.308$          & $82.1\%$         & $0.199$ \\
Octahedral (aggressive)   & 3 & 6 & $[-60, 80]$ & $0.302$          & $83.5\%$         & $0.192$ \\
Octahedral (full sphere)  & 3 & 6 & $[-89, 89]$ & $0.301$          & $79.8\%$         & $0.195$ \\
\bottomrule
\end{tabular}
\vspace{-0.3cm}
\end{table*}

% \vspace{-0.3cm}
\vspace{-0.2cm} \paragraph{Direction of effect transfers.} The central findings
transfer to the separate prompt set. First, antithetic sampling beats
uniform at $K{=}2$ (F1). Second, every non-octahedral \mvsdi
configuration beats the baseline on all three shared metrics (F2),
with the $K{=}2$ azimuth pair the sweet spot and equatorial
one-to-two-plane configurations dominating the wide-elevation
octahedral ones. Third, the octahedral variants remain the weakest
family here as on the $43$-prompt set: they trail the baseline on
HPSv2, and the two wide-elevation ones sit just below it on CLIP
within prompt-level noise.

\vspace{-0.2cm} \paragraph{Set-specific fine orderings.} Two R-Precision orderings
are set-specific: the $K{=}4$ mixed variant's lead and, more broadly,
the higher-$K$ retrieval gain (F3), both of which hold on the
$43$-prompt set but not here, where the $K{=}2$ azimuth pair leads
R-Precision.

\vspace{-0.2cm}
\section{Limitations: Extension to DiT-Based Rectified-Flow Priors}
\label{sec:appendix-flux}

A natural question is whether the \mvsdi recipe transfers from
UNet/$\epsilon$-prediction priors to transformer-based
\emph{rectified-flow} priors. We port the full pipeline to
FLUX.1-dev (Black Forest Labs), a
$12$B-parameter DiT trained with flow matching and distilled
classifier-free guidance, and report a clean negative result with
attribution. We release the full implementation (math derivation,
guidance module, prompt processor, four-stage diagnostic) so future work
on DiT-based score distillation can assess whether the same
obstacle applies to other distilled priors.

\vspace{-0.3cm} \subsection{Flow-matching reformulation of \sdi}
\label{sec:appendix-flux-math}
The \sdi surrogate, $\tfrac{1}{2}\|\Render(\thetaparam) - x_0^{\text{denoised}}\|^2$,
was originally derived for variance-preserving DDPMs with
$\epsilon$-prediction. For FLUX's rectified-flow forward process
$z_\sigma = (1-\sigma)\,x_0 + \sigma\,\epsilon$ and velocity prediction
$v = \epsilon - x_0$, the corresponding algebraic identities are
\begin{equation}
\label{eq:flux-x0}
x_0 = z_\sigma - \sigma\,v_\text{pred}, \quad
\epsilon = z_\sigma + (1-\sigma)\,v_\text{pred}.
\end{equation}
The DDIM-inversion loop is replaced by a forward-Euler integration of the
velocity field, $z_{\sigma + d\sigma} = z_\sigma + d\sigma\,v_\text{pred}$,
discretized on the model's own shifted-sigma grid. The surrogate loss is
\emph{unchanged}; only the inversion and the $x_0$ recovery are reformulated,
so the rest of \mvsdi (gradient accumulation, antithetic camera sampling,
surrogate loss) carries over without modification. The full derivation is
given in the released code.

\vspace{-0.3cm} \subsection{Implementation}
\label{sec:appendix-flux-impl}
Our port covers the full FLUX-dev stack: the $16$-channel VAE, the dual
text encoders (T5-XXL and CLIP-L), the DiT's $2{\times}2$ latent packing
with separate rotary position encodings for image and text tokens, and
bf16 precision throughout. To fit the model alongside the NeRF, we offload
the T5 encoder after caching its embeddings, freeing roughly $5$\,GB before
the distillation loop. The guidance module exposes both forms of guidance
the model admits: \emph{distilled} guidance, a single scalar restricted by
the model's training distribution to $[1, 10]$, and \emph{classical} CFG, a
two-pass forward on conditional and unconditional embeddings combined as
$v = v_{\text{uncond}} + s\,(v_{\text{cond}} - v_{\text{uncond}})$. The
latter is required because, as shown below, distilled guidance alone is
insufficient.

\vspace{-0.3cm} \subsection{Four-stage diagnostic ladder}
\label{sec:appendix-flux-diag}
Initial runs ($1500$ NeRF steps on a single \emph{red apple} prompt)
produced unrecognizable, mostly grey or coloured-blob renders. To isolate
the failure we ran four checks, each removing one dependency.

\begin{enumerate}
\item \textbf{Stand-alone guidance.} Running the guidance module alone on a
flat-grey input and sweeping $\sigma \in \{0.3, 0.5, 0.7, 0.9\}$, the decoded
$x_0$ targets at high $\sigma$ are grey noise with no prompt content,
suggesting the denoising prediction ignores the conditioning.

\item \textbf{Encoder and forward-call parity.} Both the text encoder
and the transformer wrapper are exact: cached vs.\ fresh-encoded
embeddings give byte-identical output (both produce a recognizable
apple), and our wrapper's velocity prediction matches the reference
implementation to $\cos$-sim $=1.000000$. The decoded $x_0$
estimates from this verified prediction nonetheless look like grey
noise.

\item \textbf{Classical CFG sweep.} With the verified forward call, sweeping
the classical CFG scale $s \in \{1, 3.5, 7.5, 15, 30, 50\}$ at fixed distilled
guidance moves the target from grey ($s \leq 15$) to uniformly pink
($s = 50$); structural prompt content never materializes.
\end{enumerate}

Together, the second and third checks rule out both the encoder and the
transformer call; the fourth narrows the failure to the guidance-amplification
regime.

\begin{figure*}[!t]
    \centering
    \vspace{-0.2cm}
    \includegraphics[width=\linewidth]{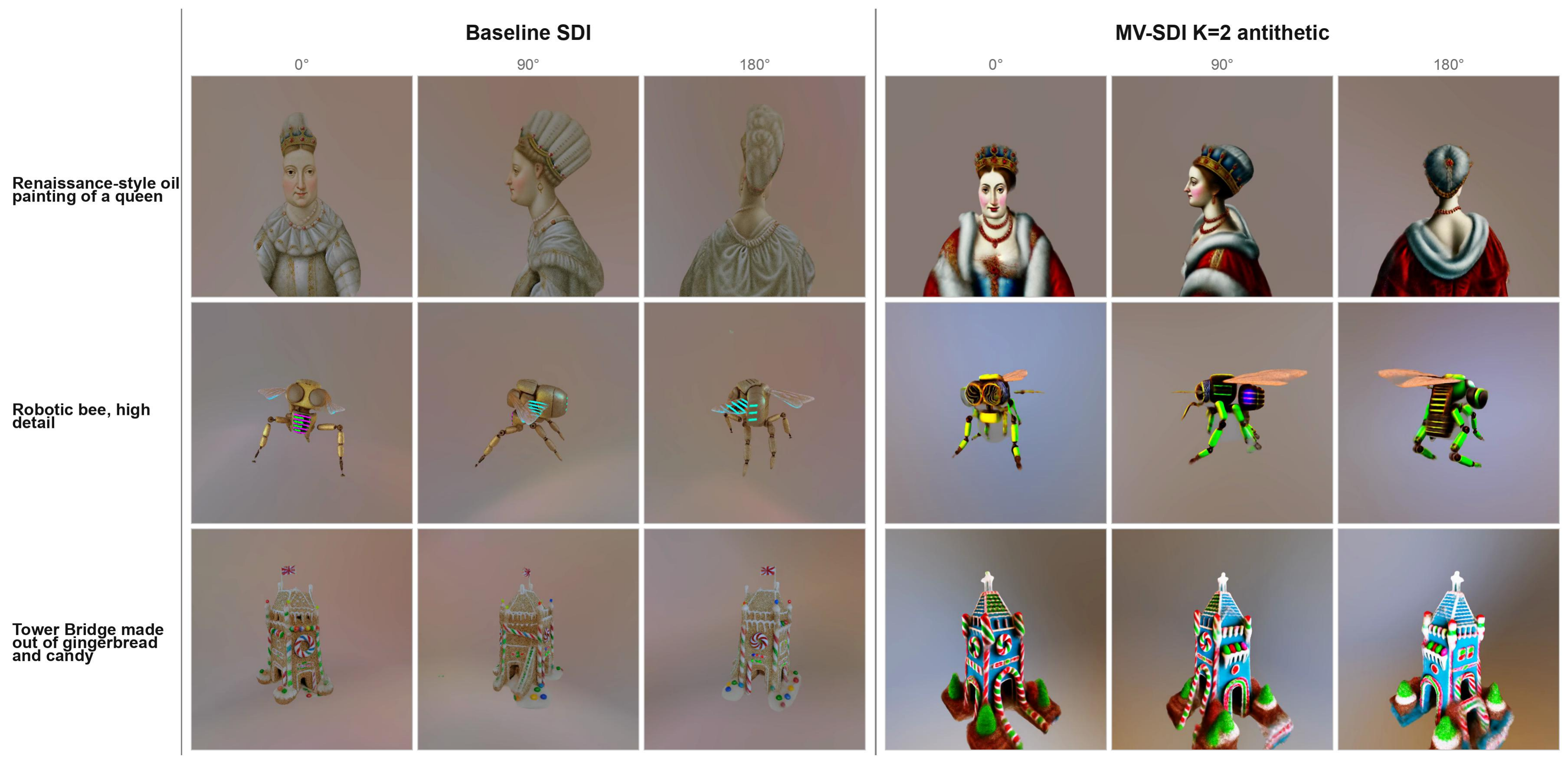}
    \vspace{-0.6cm}
    \captionof{figure}{\textbf{Qualitative gallery on SDI's prompt suite (part 1/6).}
    For each prompt (rows) we show baseline \sdi{} (left) and \mvsdi{} $K{=}2$ antithetic (right),
    each at three orbit views ($0^\circ/90^\circ/180^\circ$). Best viewed zoomed-in.}
    \label{fig:sdi_qual_appendix_p1}
    \vspace{-0.6cm}
\end{figure*}

\vspace{-0.2cm} \subsection{Empirical findings on the full training loop}
\label{sec:appendix-flux-findings}
A three-way sweep at $1{,}500$ NeRF steps and forward/inversion
scale pairs $s \in \{15, 30, 60\}$, $s_\text{inv} \in \{-3.5,
-7.5, -15\}$ converges in every case to the same failure mode: a
well-formed $3$D blob geometry (visible in normal and opacity
renders) fixed at a uniform pink-magenta color in RGB. The loss
saturates by ${\sim}500$ steps and no run drifts toward the
prompt's red apple.

\vspace{-0.4cm} \subsection{Attribution: distilled guidance is the bottleneck}
\label{sec:appendix-flux-attribution}
Two architectural properties of FLUX-dev explain the observed mode
collapse:
\begin{description}
% \vspace{-0.3cm}
\item[(P1) Non-negative, bounded distilled guidance.] FLUX-dev was trained
with distilled CFG values in approximately $[1, 10]$ and is
non-negative by construction (it enters the model as a non-negative
embedding). Our SD~2.1 \sdi recipe instead uses forward CFG $+7.5$ during
prediction and a \emph{negative} CFG $-7.5$ during the DDIM-inversion
step, the anti-prompt move at the heart of \sdi. The forward $+7.5$
lies inside FLUX-dev's distilled range, but the negative inversion
guidance cannot be expressed through a non-negative $[1, 10]$
embedding; reproducing it requires the classical two-pass CFG path,
which (P2) is too weak to drive structure on this model.

\vspace{0.1cm}
\item[(P2) High-dim. text conditioning dilutes external CFG.]
FLUX-dev's text branch comprises a $256$-token T5 sequence
($\mathbb{R}^{256 \times 4096}$) plus a single $768$-d pooled CLIP
embedding. Classical CFG amplifies the difference
$(v_{\text{cond}} - v_{\text{uncond}})$ uniformly. Empirically the
pooled CLIP signal (low-dimensional, dominant in colour cues)
\emph{does} amplify with CFG; this is what produces the pink colour
trajectory at higher $s$. The T5 structural signal (high-dimensional,
distributed across $256$ tokens) is averaged across heads in the DiT's
joint attention and is dominated by the input's structure once the
NeRF has reached a stable blob shape; further CFG amplification
saturates without re-allocating mass toward the prompt's structural
information.
\end{description}

We measured the conditional-to-unconditional separation at
$\sigma = 0.9$:
$\|v_\text{cond} - v_\text{uncond}\| \,/\, \|v_\text{cond}\| \approx 3\%$,
versus typical values of $20$--$40\%$ for SD~2.1 at the same noise
level. The prompt signal is simply too weak at the velocity level to
overcome the input's attractor.

\vspace{-0.2cm} \subsection{Pathways for future work}
\label{sec:appendix-flux-future}
Two concrete paths could circumvent \textbf{(P1)} and \textbf{(P2)}:
\begin{itemize}
\item \textbf{Multi-step $x_0$ estimation.} Replace the single-step
$x_0 = z_\sigma - \sigma\,v_\text{pred}$ with $K$ Euler denoising
steps from $\sigma$ down to $0$. This trades a $K\!\times$ compute
slowdown per training iteration for a much sharper $x_0$ target
(closer to the $20$-step generation that produces prompt-aligned
outputs from pure noise with the same model). Recent work adapting
score distillation to rectified-flow priors~\cite{lupascu2026ot,
yan2025cfd} attributes the residual over-smoothing to the
distillation target rather than the prior; a multi-step $x_0$
estimate is a direct way to sharpen that target.

\item \textbf{VSD-style online critic.} Train a small LoRA on the
DiT online, conditioned on the current NeRF render, to provide a
prompt-aware gradient that does not rely on aggressive CFG
amplification. This is the
ProlificDreamer~\cite{wang2023prolificdreamer} recipe adapted to
flow matching, and trivially compatible with our multi-view
antithetic sampling.
\end{itemize}
A third path would sidestep (P1) entirely: a future FLUX release
without the distilled-guidance bottleneck (e.g.\ a
non-distilled variant) would directly unlock the SD-style
aggressive-CFG regime.

None of the above invalidates our main \mvsdi findings on
SD 2.1: the multi-view aggregation logic is unchanged in all three
paths. Our finding is that the \emph{prior-side prerequisites} for
single-step SDS (broad-range, possibly-negative CFG) are not met by
the current crop of distilled rectified-flow models, not that the
aggregation principle itself fails.

% \vspace{-0.3cm}
\paragraph{Released artifacts.}
To support further work, we release the flow-matching SDI derivation, the
guidance and prompt-processor modules, three reference configurations
(baseline and $K{=}2$/$K{=}4$ antithetic), and the four diagnostic scripts,
together with implementation notes.

\vspace{-0.2cm}

\begin{figure*}[h]
    \centering
    \includegraphics[width=\linewidth]{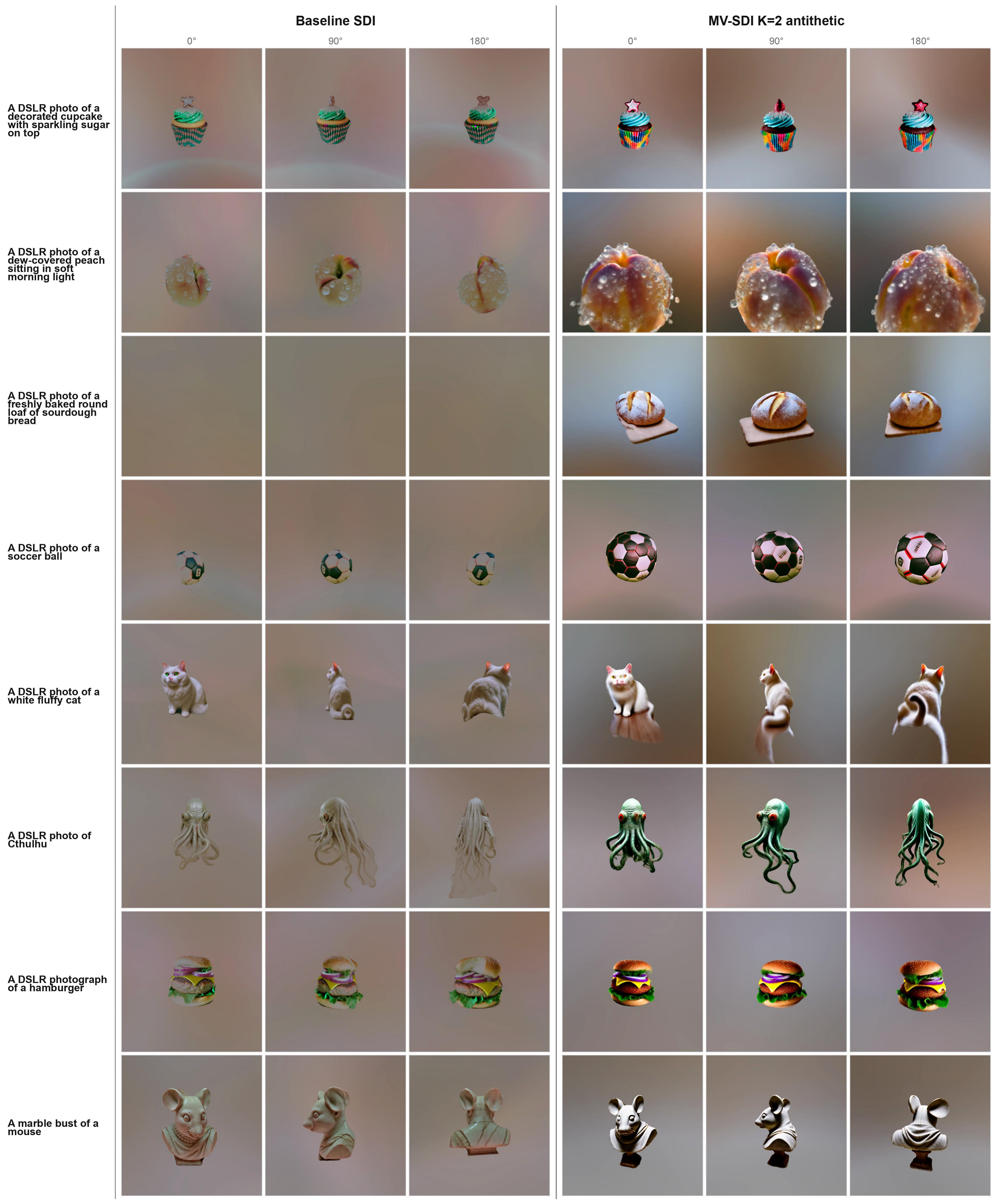}
    \caption{\textbf{Qualitative gallery on SDI's prompt suite (part 2/6).}
    Layout as in Fig.~\ref{fig:sdi_qual_appendix_p1}.}
    \label{fig:sdi_qual_appendix_p2}
\end{figure*}

\begin{figure*}[h]
    \centering
    \includegraphics[width=\linewidth]{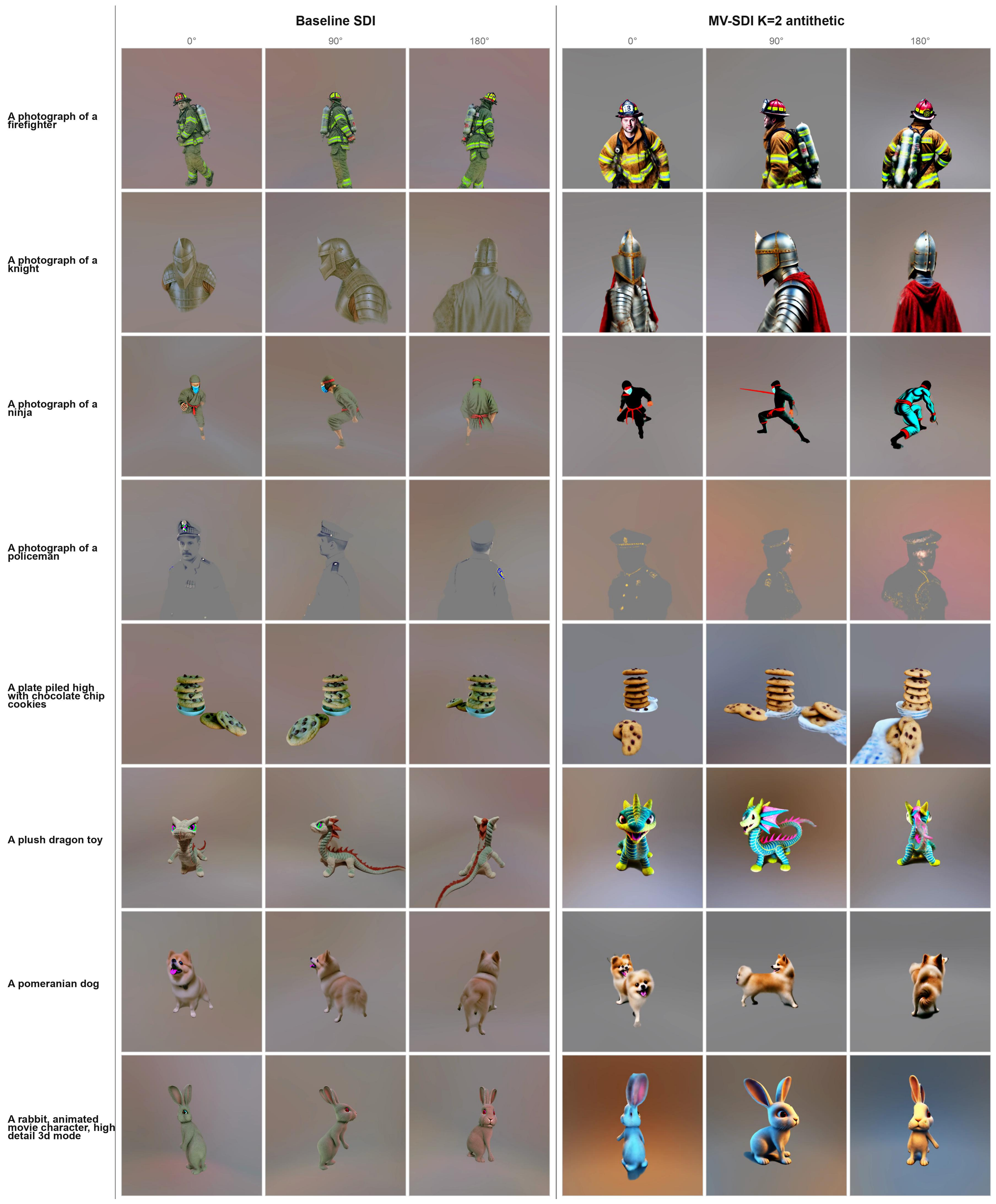}
    \caption{\textbf{Qualitative gallery on SDI's prompt suite (part 3/6).}
    Layout as in Fig.~\ref{fig:sdi_qual_appendix_p1}.}
    \label{fig:sdi_qual_appendix_p3}
\end{figure*}

\begin{figure*}[h]
    \centering
    \includegraphics[width=\linewidth]{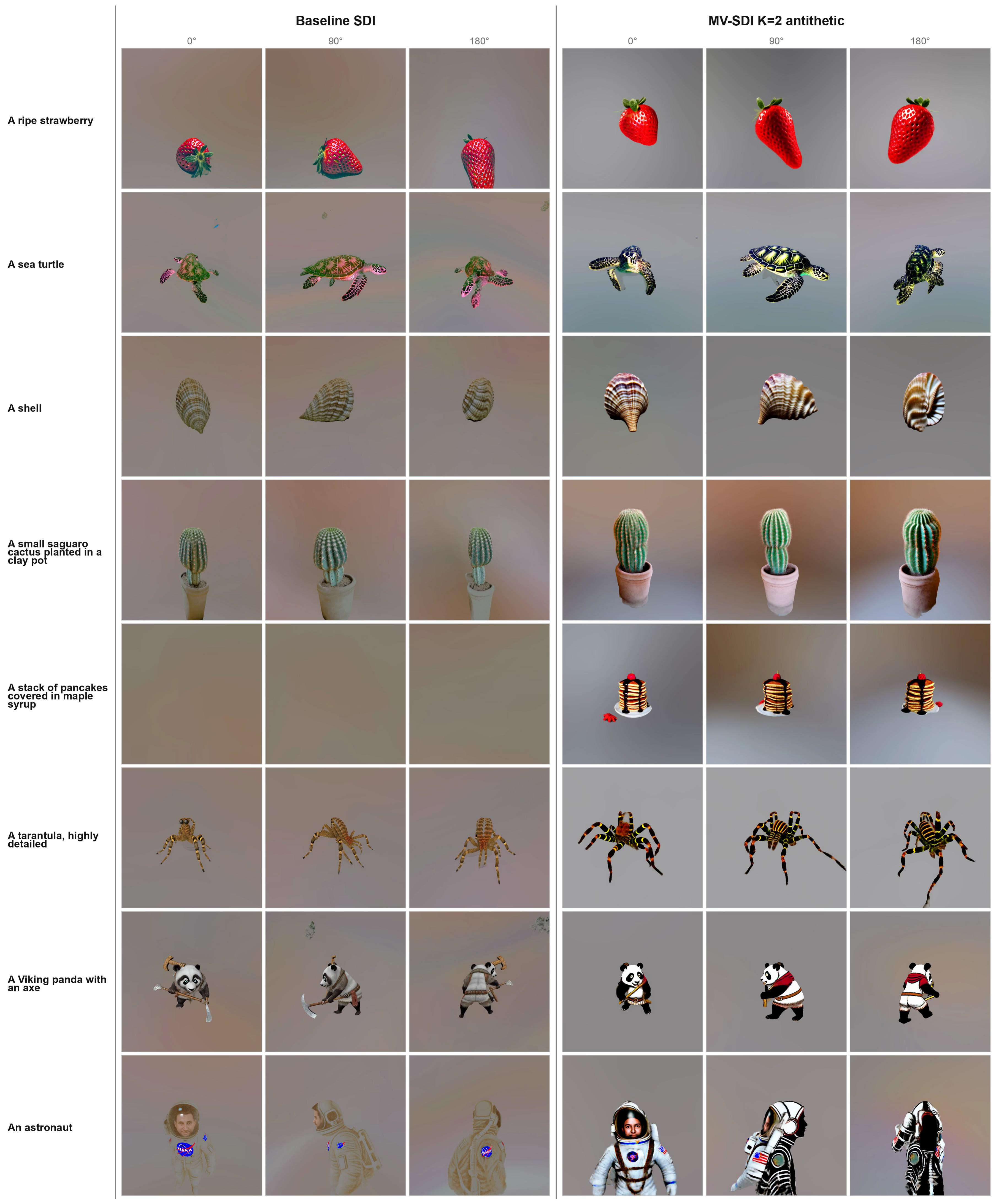}
    \caption{\textbf{Qualitative gallery on SDI's prompt suite (part 4/6).}
    Layout as in Fig.~\ref{fig:sdi_qual_appendix_p1}.}
    \label{fig:sdi_qual_appendix_p4}
\end{figure*}

\begin{figure*}[h]
    \centering
    \includegraphics[width=\linewidth]{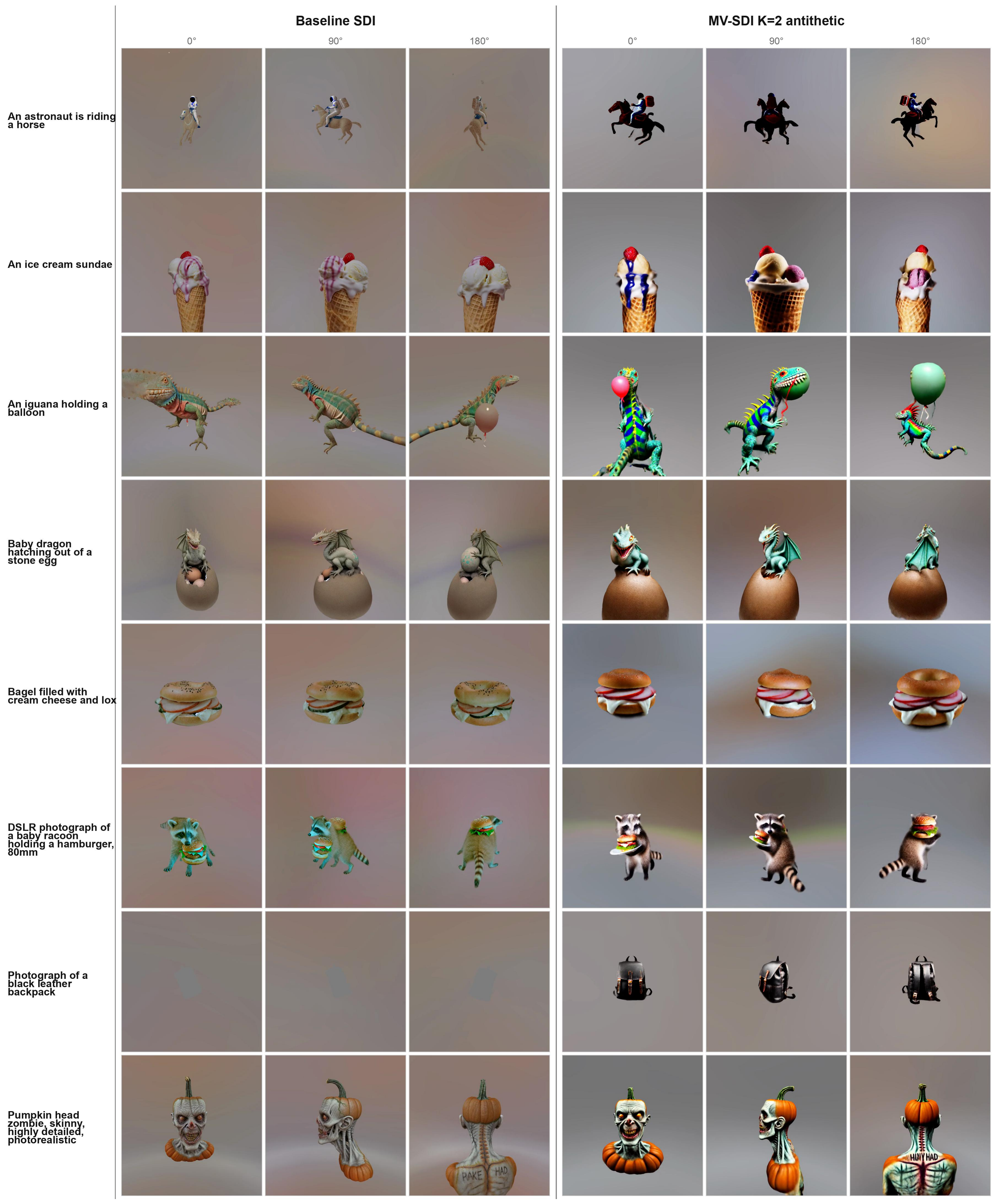}
    \caption{\textbf{Qualitative gallery on SDI's prompt suite (part 5/6).}
    Layout as in Fig.~\ref{fig:sdi_qual_appendix_p1}.}
    \label{fig:sdi_qual_appendix_p5}
\end{figure*}

\begin{figure*}[h]
    \centering
    \includegraphics[width=\linewidth]{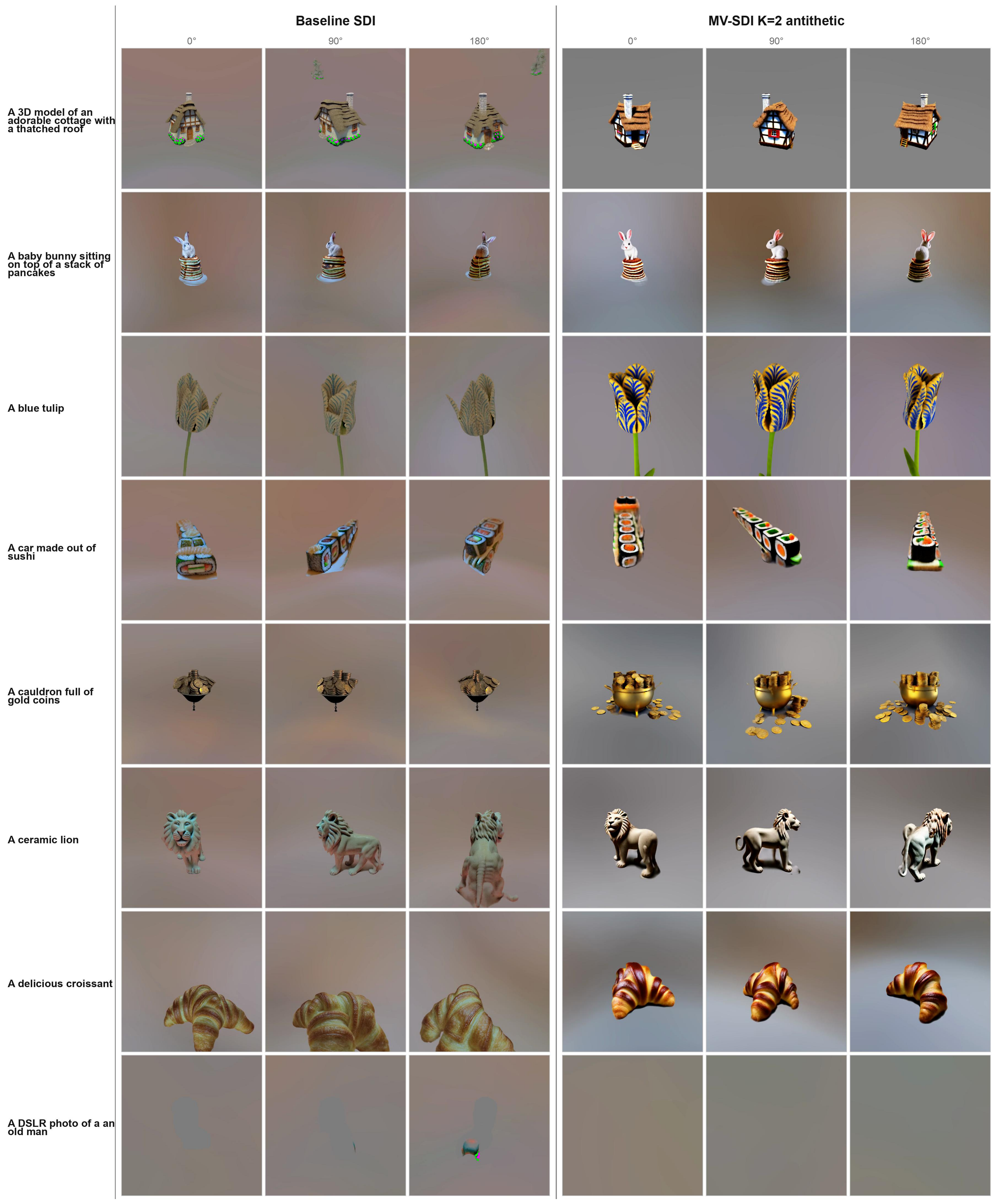}
    \caption{\textbf{Qualitative gallery on SDI's prompt suite (part 6/6).}
    Layout as in Fig.~\ref{fig:sdi_qual_appendix_p1}.}
    \label{fig:sdi_qual_appendix_p6}
\end{figure*}

\begin{figure*}[h]
    \centering
    \includegraphics[width=\linewidth]{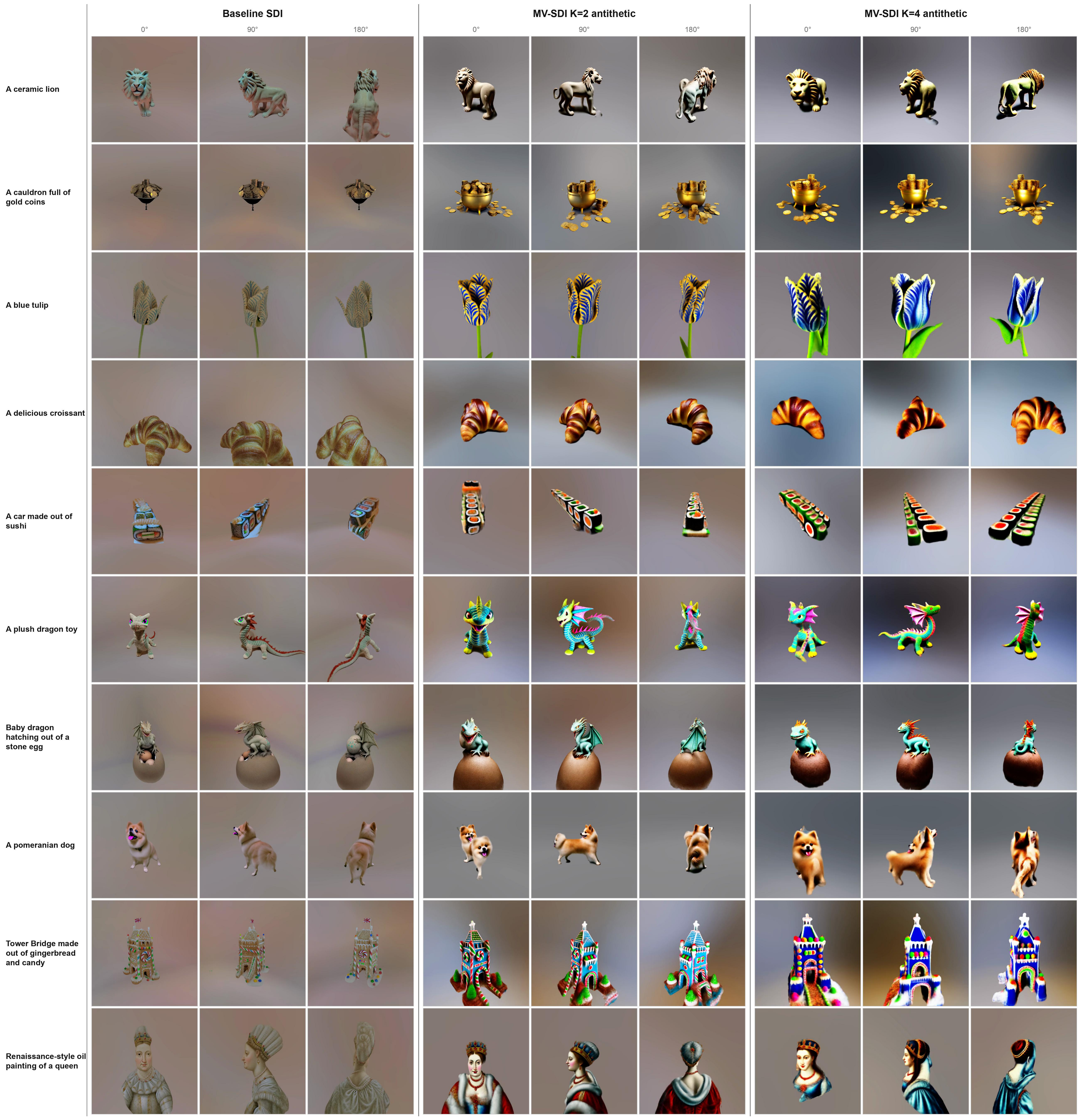}
    \caption{\textbf{Full variant comparison on selected prompts (part 1/2).}
    For each prompt (rows) we show baseline \sdi{} (left), \mvsdi{} $K{=}2$ antithetic (center),
    and \mvsdi{} $K{=}4$ antithetic (right), each at three orbit views ($0^\circ/90^\circ/180^\circ$).
    Best viewed zoomed-in.}
    \label{fig:ablation_gallery2_p1}
\end{figure*}

\begin{figure*}[h]
    \centering
    \includegraphics[width=\linewidth]{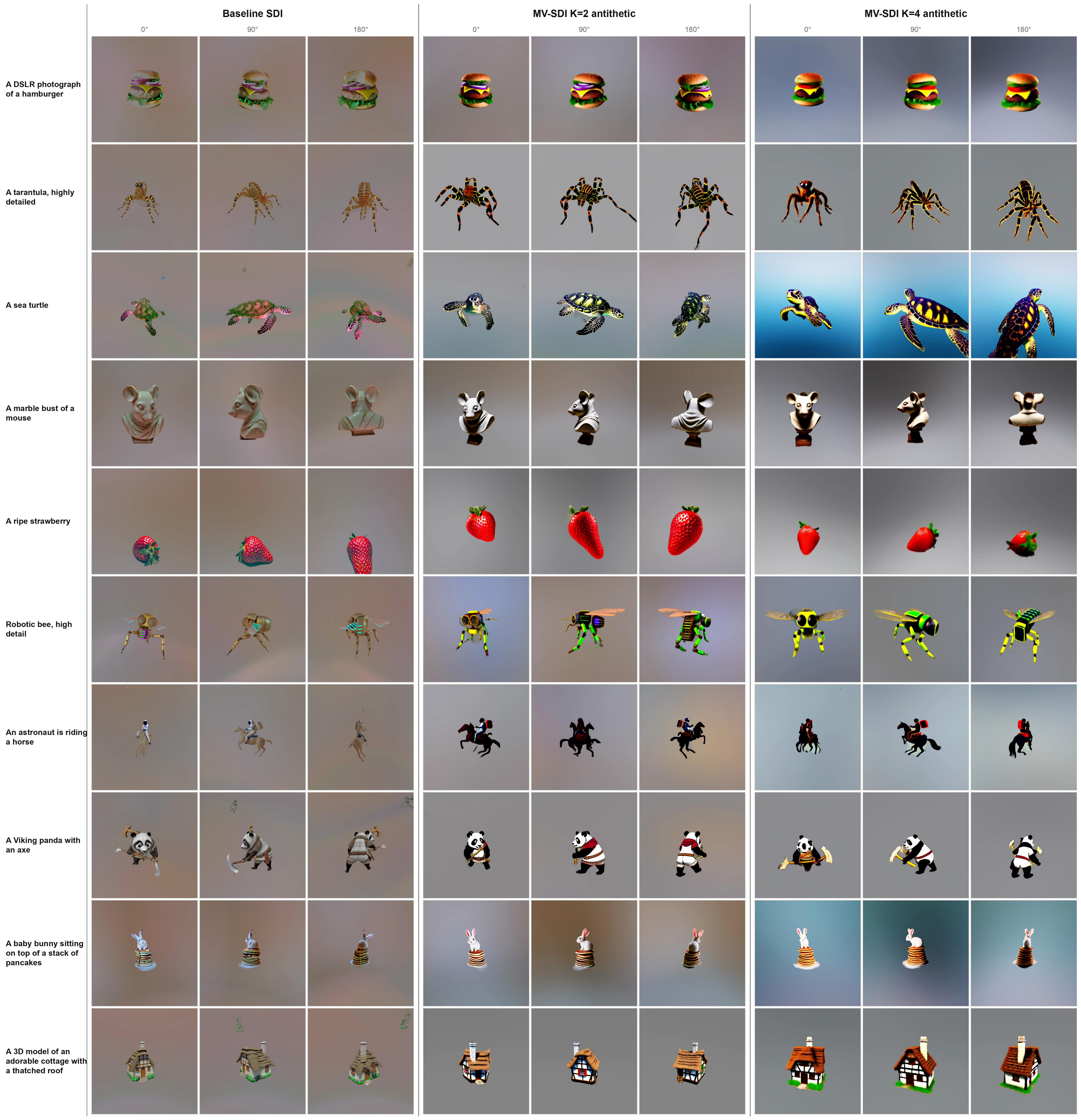}
    \caption{\textbf{Full variant comparison on selected prompts (part 2/2).}
    Layout as in Fig.~\ref{fig:ablation_gallery2_p1}.}
    \label{fig:ablation_gallery2_p2}
\end{figure*}

\clearpage

% Per-prompt scores (one full-width table per metric). Auto-generated.
% Each table: rows = prompts (43 alphabetical), columns = baseline + the seven MV-SDI configs.

\begin{table*}[p]
\centering
\footnotesize
\setlength{\tabcolsep}{4pt}
\caption{Per-prompt CLIP score on the 43 SDI prompts. \textit{base} = our reproduction of baseline \sdi; column headings are short tags for MV-SDI variants (Tab.~\ref{tab:main_results} K2u = MV-SDI K=2 uniform, K2a = MV-SDI K=2 antithetic, K4a = MV-SDI K=4 antithetic, Mix4 = MV-SDI K=4 mixed (azim+elev), Oct.m = MV-SDI K=6 octa (elev $\pm$30,60), Oct.a = MV-SDI K=6 octa (elev $\pm$60,80), Oct.f = MV-SDI K=6 octa (full sphere)). \textit{---} indicates the prompt diverged for that config (see Div\% in main tables).}
\label{tab:per_prompt_clip}
\begin{tabular}{p{0.34\linewidth} c c c c c c c c}
\toprule
Prompt & base & K2u & K2a & K4a & Mix4 & Oct.m & Oct.a & Oct.f \\
\midrule
A 3D model of an adorable cottage with ... & 0.315 & 0.359 & 0.366 & 0.347 & 0.334 & 0.341 & 0.332 & 0.321 \\
A baby bunny sitting on top of a stack ... & 0.299 & 0.321 & 0.301 & 0.343 & 0.327 & 0.325 & 0.328 & 0.325 \\
A blue tulip & 0.288 & 0.335 & 0.327 & 0.343 & 0.334 & 0.344 & 0.341 & 0.304 \\
A car made out of sushi & 0.290 & 0.274 & 0.281 & 0.286 & 0.279 & 0.279 & 0.322 & 0.263 \\
A cauldron full of gold coins & 0.271 & 0.286 & 0.345 & 0.342 & 0.319 & 0.340 & 0.342 & 0.343 \\
A ceramic lion & 0.308 & 0.327 & 0.316 & 0.303 & 0.304 & 0.296 & 0.295 & 0.305 \\
A delicious croissant & 0.325 & 0.317 & 0.319 & 0.301 & 0.289 & 0.304 & 0.273 & 0.311 \\
A DSLR photo of a an old man & 0.217 & --- & 0.294 & 0.283 & 0.288 & 0.226 & 0.240 & --- \\
A DSLR photo of a decorated cupcake wit... & 0.314 & 0.329 & 0.319 & 0.333 & 0.321 & 0.340 & 0.321 & 0.321 \\
A DSLR photo of a dew-covered peach sit... & 0.298 & 0.323 & 0.335 & 0.270 & 0.328 & 0.310 & 0.314 & 0.301 \\
A DSLR photo of a freshly baked round l... & 0.194 & 0.275 & 0.288 & 0.245 & 0.266 & 0.257 & 0.284 & 0.286 \\
A DSLR photo of a soccer ball & 0.304 & 0.316 & 0.318 & 0.308 & 0.309 & 0.309 & 0.304 & 0.298 \\
A DSLR photo of a white fluffy cat & 0.302 & 0.305 & 0.292 & 0.300 & 0.300 & 0.295 & 0.294 & 0.292 \\
A DSLR photo of Cthulhu & 0.280 & 0.285 & 0.292 & 0.285 & 0.289 & 0.285 & 0.290 & 0.288 \\
A DSLR photograph of a hamburger & 0.318 & 0.319 & 0.314 & 0.318 & 0.316 & 0.324 & 0.318 & 0.319 \\
A marble bust of a mouse & 0.331 & 0.314 & 0.313 & 0.279 & 0.305 & 0.313 & 0.293 & 0.302 \\
A photograph of a firefighter & 0.333 & 0.338 & 0.335 & 0.333 & 0.337 & 0.325 & 0.326 & 0.313 \\
A photograph of a knight & 0.299 & 0.308 & 0.310 & 0.304 & 0.311 & 0.276 & 0.284 & 0.303 \\
A photograph of a ninja & 0.279 & 0.301 & 0.292 & 0.280 & 0.299 & 0.285 & 0.294 & 0.281 \\
A photograph of a policeman & 0.295 & 0.265 & 0.288 & 0.293 & 0.291 & 0.293 & 0.290 & 0.295 \\
A plate piled high with chocolate chip ... & 0.306 & 0.319 & 0.315 & 0.301 & 0.283 & 0.285 & 0.280 & 0.287 \\
A plush dragon toy & 0.268 & 0.286 & 0.277 & 0.285 & 0.282 & 0.296 & 0.278 & 0.287 \\
A pomeranian dog & 0.283 & 0.296 & 0.297 & 0.287 & 0.290 & 0.284 & 0.286 & 0.283 \\
A rabbit, animated movie character, hig... & 0.375 & 0.376 & 0.382 & 0.368 & 0.373 & 0.378 & 0.363 & 0.366 \\
A ripe strawberry & 0.325 & 0.337 & 0.338 & 0.325 & 0.330 & 0.323 & 0.333 & 0.335 \\
A sea turtle & 0.306 & 0.309 & 0.313 & 0.310 & 0.315 & 0.307 & 0.311 & 0.309 \\
A shell & 0.322 & 0.334 & 0.334 & 0.332 & 0.323 & 0.328 & 0.311 & 0.323 \\
A small saguaro cactus planted in a cla... & 0.341 & 0.321 & 0.329 & 0.320 & 0.313 & 0.316 & 0.295 & 0.302 \\
A stack of pancakes covered in maple sy... & 0.198 & 0.315 & 0.323 & 0.311 & 0.308 & 0.302 & 0.297 & 0.309 \\
A tarantula, highly detailed & 0.298 & 0.283 & 0.295 & 0.281 & 0.300 & 0.287 & 0.285 & 0.281 \\
A Viking panda with an axe & 0.326 & 0.337 & 0.316 & 0.322 & 0.314 & 0.292 & 0.319 & 0.301 \\
An astronaut & 0.303 & 0.303 & 0.306 & 0.303 & 0.301 & 0.310 & 0.303 & 0.299 \\
An astronaut is riding a horse & 0.315 & 0.304 & 0.310 & 0.308 & 0.309 & 0.298 & 0.300 & 0.301 \\
An ice cream sundae & 0.292 & 0.292 & 0.291 & 0.295 & 0.291 & 0.305 & 0.301 & 0.298 \\
An iguana holding a balloon & 0.348 & 0.330 & 0.324 & 0.337 & 0.313 & 0.245 & 0.279 & 0.270 \\
Baby dragon hatching out of a stone egg & 0.344 & 0.340 & 0.318 & 0.311 & 0.302 & 0.289 & 0.298 & 0.283 \\
Bagel filled with cream cheese and lox & 0.312 & 0.295 & 0.298 & 0.286 & 0.288 & 0.275 & 0.277 & 0.275 \\
DSLR photograph of a baby racoon holdin... & 0.277 & 0.351 & 0.354 & 0.336 & 0.352 & 0.330 & 0.328 & 0.311 \\
Photograph of a black leather backpack & 0.230 & 0.285 & 0.232 & 0.295 & 0.308 & 0.287 & 0.294 & --- \\
Pumpkin head zombie, skinny, highly det... & 0.346 & 0.336 & 0.353 & 0.320 & 0.298 & 0.318 & 0.318 & 0.301 \\
Renaissance-style oil painting of a que... & 0.299 & 0.210 & 0.319 & 0.291 & 0.298 & 0.305 & 0.260 & 0.249 \\
Robotic bee, high detail & 0.323 & 0.334 & 0.325 & 0.328 & 0.323 & 0.316 & 0.308 & 0.320 \\
Tower Bridge made out of gingerbread an... & 0.297 & 0.305 & 0.294 & 0.281 & 0.295 & 0.291 & 0.257 & 0.279 \\
\bottomrule
\end{tabular}
\end{table*}

\begin{table*}[p]
\centering
\footnotesize
\setlength{\tabcolsep}{4pt}
\caption{Per-prompt R-Precision on the 43 SDI prompts. \textit{base} = our reproduction of baseline \sdi; column headings are short tags for MV-SDI variants (Tab.~\ref{tab:main_results}: K2u = MV-SDI K=2 uniform, K2a = MV-SDI K=2 antithetic, K4a = MV-SDI K=4 antithetic, Mix4 = MV-SDI K=4 mixed (azim+elev), Oct.m = MV-SDI K=6 octa (elev $\pm$30,60), Oct.a = MV-SDI K=6 octa (elev $\pm$60,80), Oct.f = MV-SDI K=6 octa (full sphere)). \textit{---} indicates the prompt diverged for that config (see Div\% in main tables).}
\label{tab:per_prompt_rprecision}
\begin{tabular}{p{0.34\linewidth} c c c c c c c c}
\toprule
Prompt & base & K2u & K2a & K4a & Mix4 & Oct.m & Oct.a & Oct.f \\
\midrule
A 3D model of an adorable cottage with ... & 96.00 & 100.00 & 100.00 & 100.00 & 100.00 & 100.00 & 98.00 & 96.00 \\
A baby bunny sitting on top of a stack ... & 40.00 & 6.00 & 8.00 & 76.00 & 48.00 & 34.00 & 16.00 & 14.00 \\
A blue tulip & 86.00 & 100.00 & 100.00 & 100.00 & 100.00 & 100.00 & 100.00 & 100.00 \\
A car made out of sushi & 68.00 & 74.00 & 80.00 & 80.00 & 72.00 & 76.00 & 100.00 & 62.00 \\
A cauldron full of gold coins & 28.00 & 64.00 & 100.00 & 100.00 & 100.00 & 100.00 & 100.00 & 100.00 \\
A ceramic lion & 88.00 & 100.00 & 100.00 & 100.00 & 98.00 & 92.00 & 96.00 & 100.00 \\
A delicious croissant & 96.00 & 80.00 & 58.00 & 18.00 & 6.00 & 60.00 & 62.00 & 64.00 \\
A DSLR photo of a an old man & 0.00 & --- & 100.00 & 96.00 & 100.00 & 0.00 & 16.00 & --- \\
A DSLR photo of a decorated cupcake wit... & 80.00 & 100.00 & 100.00 & 100.00 & 100.00 & 100.00 & 100.00 & 100.00 \\
A DSLR photo of a dew-covered peach sit... & 100.00 & 100.00 & 100.00 & 34.00 & 100.00 & 94.00 & 96.00 & 90.00 \\
A DSLR photo of a freshly baked round l... & 0.00 & 14.00 & 28.00 & 0.00 & 0.00 & 0.00 & 4.00 & 6.00 \\
A DSLR photo of a soccer ball & 90.00 & 100.00 & 100.00 & 100.00 & 100.00 & 100.00 & 100.00 & 100.00 \\
A DSLR photo of a white fluffy cat & 74.00 & 74.00 & 46.00 & 54.00 & 62.00 & 70.00 & 46.00 & 38.00 \\
A DSLR photo of Cthulhu & 62.00 & 100.00 & 100.00 & 100.00 & 100.00 & 100.00 & 100.00 & 100.00 \\
A DSLR photograph of a hamburger & 100.00 & 100.00 & 100.00 & 100.00 & 100.00 & 100.00 & 100.00 & 100.00 \\
A marble bust of a mouse & 74.00 & 52.00 & 86.00 & 72.00 & 96.00 & 96.00 & 70.00 & 92.00 \\
A photograph of a firefighter & 100.00 & 100.00 & 100.00 & 100.00 & 100.00 & 100.00 & 100.00 & 100.00 \\
A photograph of a knight & 94.00 & 100.00 & 100.00 & 100.00 & 100.00 & 82.00 & 68.00 & 100.00 \\
A photograph of a ninja & 78.00 & 100.00 & 98.00 & 96.00 & 100.00 & 88.00 & 76.00 & 82.00 \\
A photograph of a policeman & 88.00 & 52.00 & 32.00 & 64.00 & 94.00 & 96.00 & 80.00 & 94.00 \\
A plate piled high with chocolate chip ... & 90.00 & 100.00 & 100.00 & 86.00 & 52.00 & 72.00 & 66.00 & 76.00 \\
A plush dragon toy & 30.00 & 86.00 & 54.00 & 90.00 & 100.00 & 92.00 & 50.00 & 72.00 \\
A pomeranian dog & 52.00 & 98.00 & 76.00 & 56.00 & 78.00 & 68.00 & 54.00 & 62.00 \\
A rabbit, animated movie character, hig... & 100.00 & 100.00 & 100.00 & 100.00 & 100.00 & 100.00 & 100.00 & 100.00 \\
A ripe strawberry & 100.00 & 100.00 & 100.00 & 100.00 & 100.00 & 96.00 & 100.00 & 100.00 \\
A sea turtle & 100.00 & 92.00 & 100.00 & 100.00 & 100.00 & 100.00 & 100.00 & 100.00 \\
A shell & 100.00 & 100.00 & 100.00 & 100.00 & 100.00 & 100.00 & 100.00 & 100.00 \\
A small saguaro cactus planted in a cla... & 100.00 & 100.00 & 100.00 & 100.00 & 100.00 & 100.00 & 96.00 & 100.00 \\
A stack of pancakes covered in maple sy... & 0.00 & 100.00 & 100.00 & 82.00 & 100.00 & 88.00 & 98.00 & 86.00 \\
A tarantula, highly detailed & 100.00 & 36.00 & 100.00 & 92.00 & 100.00 & 80.00 & 70.00 & 62.00 \\
A Viking panda with an axe & 98.00 & 100.00 & 100.00 & 100.00 & 100.00 & 100.00 & 98.00 & 94.00 \\
An astronaut & 100.00 & 100.00 & 100.00 & 94.00 & 100.00 & 100.00 & 98.00 & 84.00 \\
An astronaut is riding a horse & 82.00 & 100.00 & 100.00 & 94.00 & 98.00 & 84.00 & 84.00 & 94.00 \\
An ice cream sundae & 100.00 & 100.00 & 100.00 & 100.00 & 100.00 & 100.00 & 100.00 & 98.00 \\
An iguana holding a balloon & 100.00 & 100.00 & 100.00 & 100.00 & 98.00 & 8.00 & 72.00 & 36.00 \\
Baby dragon hatching out of a stone egg & 100.00 & 100.00 & 90.00 & 100.00 & 100.00 & 80.00 & 90.00 & 56.00 \\
Bagel filled with cream cheese and lox & 82.00 & 30.00 & 56.00 & 50.00 & 40.00 & 6.00 & 46.00 & 16.00 \\
DSLR photograph of a baby racoon holdin... & 36.00 & 100.00 & 100.00 & 94.00 & 100.00 & 94.00 & 96.00 & 98.00 \\
Photograph of a black leather backpack & 2.00 & 68.00 & 2.00 & 100.00 & 98.00 & 80.00 & 94.00 & --- \\
Pumpkin head zombie, skinny, highly det... & 100.00 & 100.00 & 100.00 & 100.00 & 100.00 & 98.00 & 86.00 & 80.00 \\
Renaissance-style oil painting of a que... & 96.00 & 0.00 & 100.00 & 82.00 & 100.00 & 98.00 & 44.00 & 24.00 \\
Robotic bee, high detail & 100.00 & 98.00 & 96.00 & 100.00 & 100.00 & 100.00 & 92.00 & 100.00 \\
Tower Bridge made out of gingerbread an... & 60.00 & 100.00 & 94.00 & 90.00 & 100.00 & 96.00 & 44.00 & 82.00 \\
\bottomrule
\end{tabular}
\end{table*}

\begin{table*}[p]
\centering
\footnotesize
\setlength{\tabcolsep}{4pt}
\caption{Per-prompt HPSv2 on the 43 SDI prompts. \textit{base} = our reproduction of baseline \sdi; column headings are short tags for MV-SDI variants (Tab.~\ref{tab:main_results}: K2u = MV-SDI K=2 uniform, K2a = MV-SDI K=2 antithetic, K4a = MV-SDI K=4 antithetic, Mix4 = MV-SDI K=4 mixed (azim+elev), Oct.m = MV-SDI K=6 octa (elev $\pm$30,60), Oct.a = MV-SDI K=6 octa (elev $\pm$60,80), Oct.f = MV-SDI K=6 octa (full sphere)). \textit{---} indicates the prompt diverged for that config (see Div\% in main tables).}
\label{tab:per_prompt_hpsv2}
\begin{tabular}{p{0.34\linewidth} c c c c c c c c}
\toprule
Prompt & base & K2u & K2a & K4a & Mix4 & Oct.m & Oct.a & Oct.f \\
\midrule
A 3D model of an adorable cottage with ... & 0.191 & 0.219 & 0.219 & 0.195 & 0.193 & 0.182 & 0.177 & 0.177 \\
A baby bunny sitting on top of a stack ... & 0.225 & 0.234 & 0.229 & 0.235 & 0.224 & 0.225 & 0.221 & 0.232 \\
A blue tulip & 0.215 & 0.262 & 0.262 & 0.278 & 0.263 & 0.270 & 0.265 & 0.262 \\
A car made out of sushi & 0.155 & 0.164 & 0.167 & 0.176 & 0.168 & 0.170 & 0.179 & 0.171 \\
A cauldron full of gold coins & 0.165 & 0.220 & 0.240 & 0.250 & 0.209 & 0.227 & 0.227 & 0.232 \\
A ceramic lion & 0.233 & 0.247 & 0.240 & 0.229 & 0.232 & 0.233 & 0.220 & 0.219 \\
A delicious croissant & 0.209 & 0.198 & 0.207 & 0.183 & 0.175 & 0.177 & 0.154 & 0.178 \\
A DSLR photo of a an old man & 0.122 & --- & 0.206 & 0.199 & 0.211 & 0.116 & 0.109 & --- \\
A DSLR photo of a decorated cupcake wit... & 0.185 & 0.217 & 0.217 & 0.213 & 0.221 & 0.210 & 0.206 & 0.194 \\
A DSLR photo of a dew-covered peach sit... & 0.165 & 0.196 & 0.198 & 0.175 & 0.179 & 0.166 & 0.175 & 0.173 \\
A DSLR photo of a freshly baked round l... & 0.099 & 0.186 & 0.198 & 0.172 & 0.154 & 0.144 & 0.170 & 0.162 \\
A DSLR photo of a soccer ball & 0.198 & 0.242 & 0.234 & 0.239 & 0.238 & 0.230 & 0.229 & 0.233 \\
A DSLR photo of a white fluffy cat & 0.194 & 0.199 & 0.206 & 0.197 & 0.189 & 0.190 & 0.186 & 0.185 \\
A DSLR photo of Cthulhu & 0.194 & 0.230 & 0.220 & 0.227 & 0.223 & 0.217 & 0.212 & 0.218 \\
A DSLR photograph of a hamburger & 0.225 & 0.233 & 0.228 & 0.233 & 0.232 & 0.234 & 0.232 & 0.239 \\
A marble bust of a mouse & 0.240 & 0.240 & 0.235 & 0.223 & 0.226 & 0.207 & 0.186 & 0.189 \\
A photograph of a firefighter & 0.232 & 0.275 & 0.260 & 0.250 & 0.251 & 0.244 & 0.231 & 0.215 \\
A photograph of a knight & 0.177 & 0.203 & 0.230 & 0.195 & 0.210 & 0.158 & 0.155 & 0.196 \\
A photograph of a ninja & 0.191 & 0.209 & 0.204 & 0.179 & 0.222 & 0.173 & 0.197 & 0.194 \\
A photograph of a policeman & 0.189 & 0.164 & 0.167 & 0.213 & 0.206 & 0.219 & 0.195 & 0.209 \\
A plate piled high with chocolate chip ... & 0.183 & 0.205 & 0.188 & 0.184 & 0.174 & 0.159 & 0.161 & 0.164 \\
A plush dragon toy & 0.208 & 0.223 & 0.238 & 0.232 & 0.198 & 0.216 & 0.201 & 0.207 \\
A pomeranian dog & 0.194 & 0.202 & 0.194 & 0.186 & 0.189 & 0.176 & 0.178 & 0.175 \\
A rabbit, animated movie character, hig... & 0.234 & 0.259 & 0.266 & 0.259 & 0.244 & 0.238 & 0.227 & 0.234 \\
A ripe strawberry & 0.215 & 0.232 & 0.246 & 0.214 & 0.216 & 0.214 & 0.231 & 0.225 \\
A sea turtle & 0.214 & 0.231 & 0.231 & 0.247 & 0.242 & 0.245 & 0.224 & 0.221 \\
A shell & 0.243 & 0.259 & 0.252 & 0.257 & 0.247 & 0.251 & 0.245 & 0.246 \\
A small saguaro cactus planted in a cla... & 0.222 & 0.260 & 0.221 & 0.233 & 0.238 & 0.233 & 0.208 & 0.236 \\
A stack of pancakes covered in maple sy... & 0.114 & 0.200 & 0.184 & 0.183 & 0.191 & 0.177 & 0.183 & 0.180 \\
A tarantula, highly detailed & 0.206 & 0.226 & 0.215 & 0.210 & 0.216 & 0.222 & 0.207 & 0.194 \\
A Viking panda with an axe & 0.233 & 0.236 & 0.222 & 0.222 & 0.227 & 0.201 & 0.209 & 0.210 \\
An astronaut & 0.214 & 0.216 & 0.236 & 0.235 & 0.228 & 0.225 & 0.198 & 0.183 \\
An astronaut is riding a horse & 0.195 & 0.182 & 0.206 & 0.193 & 0.210 & 0.178 & 0.201 & 0.177 \\
An ice cream sundae & 0.214 & 0.236 & 0.232 & 0.223 & 0.220 & 0.222 & 0.222 & 0.208 \\
An iguana holding a balloon & 0.275 & 0.269 & 0.272 & 0.261 & 0.248 & 0.197 & 0.228 & 0.214 \\
Baby dragon hatching out of a stone egg & 0.267 & 0.255 & 0.267 & 0.249 & 0.231 & 0.215 & 0.218 & 0.216 \\
Bagel filled with cream cheese and lox & 0.199 & 0.208 & 0.214 & 0.207 & 0.179 & 0.181 & 0.186 & 0.181 \\
DSLR photograph of a baby racoon holdin... & 0.206 & 0.239 & 0.250 & 0.225 & 0.211 & 0.200 & 0.185 & 0.190 \\
Photograph of a black leather backpack & 0.203 & 0.210 & 0.118 & 0.192 & 0.154 & 0.185 & 0.214 & --- \\
Pumpkin head zombie, skinny, highly det... & 0.210 & 0.218 & 0.237 & 0.221 & 0.204 & 0.201 & 0.206 & 0.190 \\
Renaissance-style oil painting of a que... & 0.181 & 0.066 & 0.207 & 0.167 & 0.159 & 0.153 & 0.141 & 0.111 \\
Robotic bee, high detail & 0.191 & 0.227 & 0.221 & 0.225 & 0.209 & 0.205 & 0.177 & 0.207 \\
Tower Bridge made out of gingerbread an... & 0.184 & 0.208 & 0.197 & 0.189 & 0.184 & 0.179 & 0.152 & 0.166 \\
\bottomrule
\end{tabular}
\end{table*}

\begin{table*}[p]
\centering
\footnotesize
\setlength{\tabcolsep}{4pt}
\caption{Per-prompt ImageReward on the 43 SDI prompts. \textit{base} = our reproduction of baseline \sdi; column headings are short tags for MV-SDI variants (Tab.~\ref{tab:main_results}: K2u = MV-SDI K=2 uniform, K2a = MV-SDI K=2 antithetic, K4a = MV-SDI K=4 antithetic, Mix4 = MV-SDI K=4 mixed (azim+elev), Oct.m = MV-SDI K=6 octa (elev $\pm$30,60), Oct.a = MV-SDI K=6 octa (elev $\pm$60,80), Oct.f = MV-SDI K=6 octa (full sphere)). \textit{---} indicates the prompt diverged for that config (see Div\% in main tables).}
\label{tab:per_prompt_image_reward}
\begin{tabular}{p{0.34\linewidth} c c c c c c c c}
\toprule
Prompt & base & K2u & K2a & K4a & Mix4 & Oct.m & Oct.a & Oct.f \\
\midrule
A 3D model of an adorable cottage with ... & -0.13 & +0.80 & +0.75 & +0.34 & +0.43 & +0.14 & +0.01 & -0.36 \\
A baby bunny sitting on top of a stack ... & +0.16 & -0.21 & -0.00 & -0.08 & -0.87 & +0.52 & -0.15 & +0.99 \\
A blue tulip & -1.05 & +0.31 & +0.05 & +0.85 & +0.33 & +1.03 & +0.85 & +0.38 \\
A car made out of sushi & -1.96 & -1.65 & -1.50 & -1.24 & -2.00 & -0.50 & -0.55 & -1.39 \\
A cauldron full of gold coins & -2.28 & -0.39 & +1.22 & +1.34 & -0.08 & +0.75 & +0.66 & +1.18 \\
A ceramic lion & -0.26 & -0.08 & +0.01 & -0.86 & -0.53 & -0.63 & -0.83 & -1.38 \\
A delicious croissant & -1.01 & -1.84 & -1.96 & -1.87 & -2.01 & -1.73 & -2.17 & -1.82 \\
A DSLR photo of a an old man & -2.28 & --- & +0.42 & +0.30 & +0.33 & -2.28 & -2.15 & --- \\
A DSLR photo of a decorated cupcake wit... & -0.78 & -0.29 & -0.59 & -0.71 & -0.22 & -0.12 & -1.27 & -0.42 \\
A DSLR photo of a dew-covered peach sit... & -0.13 & +0.22 & +0.57 & -1.87 & -1.34 & -1.32 & -1.77 & -1.73 \\
A DSLR photo of a freshly baked round l... & -2.28 & -0.66 & -0.34 & -1.43 & -1.89 & -2.11 & -1.47 & -1.22 \\
A DSLR photo of a soccer ball & -0.03 & +0.33 & +0.28 & +0.25 & +0.15 & -0.15 & -0.13 & +0.09 \\
A DSLR photo of a white fluffy cat & -1.44 & -1.65 & -1.47 & -1.69 & -1.63 & -1.66 & -1.58 & -1.86 \\
A DSLR photo of Cthulhu & -0.96 & -0.52 & -0.61 & -0.64 & -0.61 & -0.68 & -0.90 & -0.87 \\
A DSLR photograph of a hamburger & -0.02 & +0.10 & -0.07 & +0.02 & +0.20 & +0.09 & +0.17 & +0.19 \\
A marble bust of a mouse & +1.06 & +1.04 & +0.20 & -0.77 & -0.02 & -1.05 & -1.66 & -1.20 \\
A photograph of a firefighter & -0.22 & +0.80 & +0.56 & +0.45 & +0.51 & +0.33 & +0.17 & -0.44 \\
A photograph of a knight & -0.41 & -0.29 & -0.17 & -0.83 & -0.38 & -2.16 & -1.47 & -0.30 \\
A photograph of a ninja & -0.81 & -0.17 & -0.95 & -1.01 & -0.15 & -0.92 & -0.63 & -0.98 \\
A photograph of a policeman & -0.13 & -0.85 & -0.52 & +0.23 & -0.06 & +0.00 & -0.56 & -0.17 \\
A plate piled high with chocolate chip ... & -1.03 & -0.39 & -0.52 & -1.03 & -1.92 & -1.70 & -2.14 & -1.84 \\
A plush dragon toy & -1.09 & -0.54 & -0.43 & -0.85 & -1.45 & -0.91 & -1.43 & -1.23 \\
A pomeranian dog & -1.53 & -1.21 & -1.25 & -1.41 & -1.32 & -1.45 & -1.69 & -1.89 \\
A rabbit, animated movie character, hig... & -0.38 & -0.13 & -0.05 & -0.30 & -0.38 & -0.52 & -0.70 & -0.51 \\
A ripe strawberry & -0.56 & -0.31 & +0.06 & -0.71 & -0.68 & -0.41 & +0.27 & -0.13 \\
A sea turtle & -0.30 & -0.06 & +0.10 & +0.69 & +0.62 & +0.72 & -0.10 & -0.58 \\
A shell & +0.84 & +1.35 & +1.29 & +1.27 & +1.02 & +1.19 & +0.66 & +0.90 \\
A small saguaro cactus planted in a cla... & +1.19 & +0.96 & -0.90 & +0.31 & +0.52 & +0.19 & -0.47 & +0.19 \\
A stack of pancakes covered in maple sy... & -2.27 & -0.32 & -0.94 & -0.86 & -0.91 & -1.10 & -1.04 & -0.52 \\
A tarantula, highly detailed & +0.13 & +0.66 & +0.52 & +0.43 & +0.48 & +0.43 & +0.35 & +0.24 \\
A Viking panda with an axe & +0.33 & +0.06 & -1.05 & -0.32 & -0.48 & -1.90 & -0.79 & -1.52 \\
An astronaut & +0.01 & -0.06 & -0.08 & -0.01 & -0.05 & -0.12 & -0.69 & -1.39 \\
An astronaut is riding a horse & +0.19 & -1.15 & -0.23 & -0.96 & -0.75 & -0.97 & -0.54 & -0.81 \\
An ice cream sundae & +0.19 & +0.26 & +0.18 & +0.27 & -0.16 & +0.32 & +0.24 & -0.38 \\
An iguana holding a balloon & +1.93 & +1.83 & +1.89 & +1.77 & +1.39 & -1.53 & +1.42 & +0.89 \\
Baby dragon hatching out of a stone egg & +0.95 & +0.12 & +0.06 & -0.51 & -0.73 & -1.25 & -0.77 & -1.18 \\
Bagel filled with cream cheese and lox & +0.75 & +0.78 & +1.28 & +0.71 & -0.22 & +0.07 & -0.33 & -0.99 \\
DSLR photograph of a baby racoon holdin... & +0.35 & +1.21 & +1.32 & +0.70 & +0.31 & +0.16 & -0.57 & -0.25 \\
Photograph of a black leather backpack & -2.21 & -1.57 & -2.28 & -2.10 & -1.91 & -0.98 & -0.77 & --- \\
Pumpkin head zombie, skinny, highly det... & +0.37 & +0.73 & +1.56 & +0.79 & +0.43 & +0.09 & -0.65 & -1.41 \\
Renaissance-style oil painting of a que... & +0.15 & -2.25 & +0.47 & -0.00 & -0.30 & -0.46 & -0.58 & -1.28 \\
Robotic bee, high detail & -0.46 & +0.45 & +0.15 & +0.29 & -0.24 & -0.25 & -0.61 & -0.09 \\
Tower Bridge made out of gingerbread an... & -0.96 & -1.00 & -1.49 & -1.82 & -1.57 & -1.68 & -2.11 & -2.04 \\
\bottomrule
\end{tabular}
\end{table*}

\end{document}